\documentclass[10pt,twocolumn,letterpaper]{article}

\makeatletter%atrocious fix
\@namedef{ver@everyshi.sty}{}
\makeatother%please people fix your templates

\usepackage{iccv}
\usepackage{times}
\usepackage{epsfig}
\usepackage{graphicx}
\usepackage{amsmath}
\usepackage{amssymb}
\usepackage{booktabs}
\usepackage{multirow}
\usepackage{styles}
\usepackage{xcolor}
\usepackage{placeins}
\usepackage{balance}
\usepackage[normalem]{ulem}
\usepackage{lipsum}
\usepackage{subcaption}
\usepackage{enumitem}
\usepackage{tikz}
\usepackage[accsupp]{axessibility}
\usepackage[breaklinks=true,bookmarks=false,colorlinks = true, linkcolor = blue]{hyperref}

\newcommand{\ARXIV}[2]{#1} %arxiv

\iccvfinalcopy % *** Uncomment this line for the final submission

\ificcvfinal\fi
\begin{document}
\setlist[itemize]{leftmargin=5mm}

%%%%%%%%% TITLE
\title{Panoptic Segmentation of Satellite Image Time Series \\
with Convolutional Temporal Attention Networks  }

\author{Vivien Sainte Fare Garnot \qquad\qquad Loic Landrieu\\
LASTIG, Univ. Gustave Eiffel, ENSG,
IGN, F-94160 Saint-Mande, France\\
{\tt\small \{vivien.sainte-fare-garnot,\quad loic.landrieu\}@ign.fr}
}

\maketitle
\ificcvfinal\thispagestyle{empty}\fi

%%%%%%%%% ABSTRACT
\begin{abstract}
Unprecedented access to multi-temporal satellite imagery has opened new perspectives for a variety of Earth observation tasks. Among them, pixel-precise panoptic segmentation of agricultural parcels has major economic and environmental implications.
While researchers have explored this problem for single images, we argue that the complex temporal patterns of crop phenology are better addressed with temporal sequences of images.
In this paper, we present the first end-to-end, single-stage method for panoptic segmentation of Satellite Image Time Series (SITS). 
This module can be combined with our novel image sequence encoding network which relies on temporal self-attention to extract rich and adaptive multi-scale spatio-temporal features.
We also introduce PASTIS, the first open-access SITS dataset with panoptic annotations. We demonstrate the superiority of our encoder for semantic segmentation against multiple competing %network
architectures, and set up the first state-of-the-art of panoptic segmentation of SITS. {Our \href{https://github.com/VSainteuf/utae-paps}{implementation} and \href{https://github.com/VSainteuf/pastis-benchmark}{PASTIS} are publicly available.} %\url{link-upon-publication}.

%Open access multi-temporal satellite imagery recently opened new perspectives for a wide range of Earth Observation tasks. Among these, crop type mapping is a challenging one as the downstream public and private users require a pixel precise instance and semantic segmentation of the agricultural parcels.
%To the best of our knowledge, instance segmentation of agricultural parcels has yet to addressed on satellite image time series, as previous work focuses on the mono-temporal case. We argue that the temporal dimension can bring crucial information to uphold ambiguities in the problem, and propose to fill this gap.

%In this aim, we introduce an instance segmentation architecture inspired by the single-stage CenterMask method and adapt it to our earth observation setting. We also release the first benchmark dataset of satellite image time series for crop type mapping with instance annotations. Lastly, we introduce a novel spatio-temporal encoder based on temporal self-attention, and provide experimental evidence of its superiority to  convolutional-recurrent encoders for both semantic and instance segmentation.
%We make our Pytorch code publicly available at \url{link-upon-publication}.

%\lipsum[1]
\end{abstract}

%%%%%%%%% BODY TEXT
%===================================================
\section{Introduction}
%===================================================

%auto-ignore
The precision and availability of Earth observations have continuously improved thanks to sustained advances in space-based remote sensing, such as the launch of the Planet \cite{boshuizen2014results} and the open-access Sentinel constellations \cite{drusch2012sentinel}. 
In particular, satellites with high revisit frequency contribute to a better understanding of phenomena 
with complex temporal dynamics.
Crop mapping---the driving application of this paper---relies on exploiting such temporal patterns \cite{garnot2019time} and entails major financial and environmental stakes. % for private and public actors alike.
Indeed, remote monitoring of the surface and nature of agricultural parcels is necessary for a fair allocation of agricultural subsidies ($50$ and $22$ billion euros per year in Europe and in the US, respectively) and for ensuring that best crop rotation practices are respected. More generally, the automated analysis of SITS represents a significant interest for a wide range of applications, such as surveying urban development and deforestation.
\begin{figure}[t]
    \centering
    \includegraphics[width=\columnwidth, trim=0cm 1.3cm 2cm 0cm, clip]{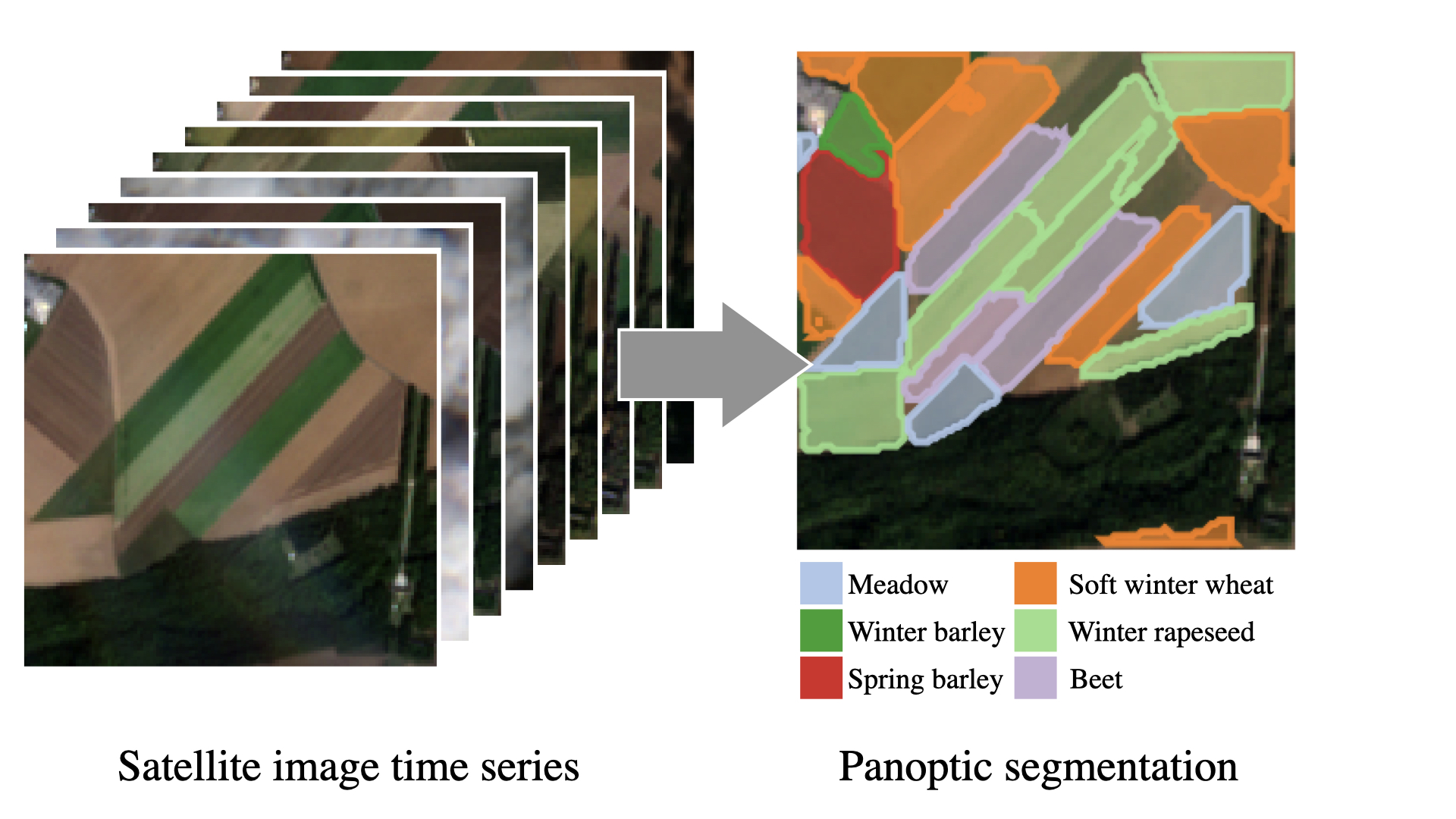}
    \caption{\textbf{Overview.} We propose an end-to-end, single-stage model for panoptic segmentation of agricultural parcels from time series of satellite images. Note the difficulty of resolving the parcels' borders from a single image, highlighting the need for modeling temporal dynamics.}
    \label{fig:my_label}
\end{figure}

The task of monitoring both the content and extent of agricultural parcels can be framed as the panoptic segmentation of an image sequence.
Panoptic segmentation consists of assigning to each pixel a class and a unique instance label, and has become a standard visual perception task in computer vision \cite{kirillov2019panoptic, mohan2021efficientps}.
%This task answers the need to monitor both the content and extent of agricultural parcels.
%However, performing panoptic segmentation  of SITS is fundamentally different than for sequences of natural images or videos.
However, panoptic segmentation is a fundamentally different task for SITS versus sequences of natural images or videos.
Indeed, understanding videos requires tracking objects through time and space \cite{tokmakov2019learning}. In yearly SITS, the targets are static in a geo-referenced frame, which removes the need for spatial tracking. 
Additionally, SITS share a common temporal frame of reference, which means that the time of acquisition itself contains information useful for modeling the underlying temporal dynamics. In contrast, the frame number in videos is often arbitrary.
Finally, while objects on the Earth surface generally do not occlude one another, as is commonly the case for objects in natural images, varying cloud cover can make the analysis of SITS arduous. 
For the specific problem addressed in this paper, individualizing agricultural parcels requires learning complex and specific temporal, spatial, and spectral patterns not commonly encountered in video processing, such as differences in plant phenological profiles, subpixel border information, and swift human interventions such as harvests.

While deep networks have proven efficient for learning such complex patterns for pixel classification \cite{kamilaris2018deep, ienco2017land, audebert2016semantic}, there is no dedicated approach for detecting individual objects in SITS. Existing work {on instance segmentation} has been restricted to analysing a single satellite image \cite{rieke2017deep}.
In summary, specialized remote sensing methods are limited to semantic segmentation or single-image instance segmentation, while computer vision's panoptic-ready networks require significant adaptation to be applied to SITS.

In this paper, we introduce U-TAE (U-net with Temporal Attention Encoder), a novel spatio-temporal encoder combining multi-scale spatial convolutions \cite{ronneberger2015u} and a temporal self-attention mechanism \cite{garnot2019time} which learns to focus on the most salient acquisitions.
While convolutional-recurrent methods are limited to extracting temporal features at the highest \cite{russwurm2018convolutional} or lowest \cite{m2019semantic} spatial resolutions, our proposed method can use the predicted temporal masks to extract specialized and adaptive spatio-temporal features at different resolutions simultaneously. 
We also propose Parcels-as-Points (PaPs), the first end-to-end deep learning method for panoptic segmentation of SITS. Our approach is built upon the efficient CenterMask network \cite{wang2020centerMASK}, which we modify to fit our problem.
Lastly, we present Panoptic Agricultural Satellite TIme-Series (PASTIS), the first open-access dataset for training and evaluating panoptic segmentation models on SITS, with over $2$ billion annotated pixels covering over $4000$km$^2$.
Evaluated on this dataset, our approach outperforms all reimplemented competing methods for semantic segmentation, and defines the first state-of-the-art of SITS panoptic segmentation.
%We evaluate our method as well as competing algorithms on this large dataset covering over $4000$km$^2$.
%showing encouraging results for panoptic segmentation, and defining a new state-of-the-art for semantic segmentation. 

%The key contributions of this paper are as follows:
%\begin{itemize}
%    \item We introduce U-TAE, a novel spatio-temporal %encoder leveraging temporal attention masks at %different spatial scales and outperforming multiple %state-of-the-art methods for semantic segmentation of %agricultural parcels.
%    \item We propose PaPs, a single-stage, end-to-end %panoptic segmentation module, specially designed to %exploit the characteristics of SITS, and defining the %first state-of-the-art for panoptic segmentation of %SCIITS.
%    \item We present PASTIS, the first public SITS %dataset with panoptic annotations, comprised of over %$2400$ image time series with $120\,000$ individual %agricultural parcels and over $2$ billion annotated %pixels.
%\end{itemize}

%===================================================
\section{Related Work}
%===================================================

%auto-ignore
To the best of our knowledge, no instance or panoptic segmentation method operating on SITS has been proposed to date. However, there is a large body of work 
on both the encoding of satellite sequences, and the panoptic segmentation of videos and single satellite images.
%on encoding satellite sequences and panoptic segmentation of videos.
%
\paragraph{Encoding Satellite Image Sequences.} 
While the first automated tools for SITS analysis relied on traditional machine learning \cite{inglada2015assessment, vuolo2018much}, deep convolutional networks allow for the extraction of richer spatial descriptors \cite{kussul2017deep, ienco2017land, audebert2016semantic, kamilaris2018deep}.
The temporal dimension was initially dealt via handcrafted temporal descriptors \cite{bailly2015dense,tavenard2017efficient,ye2009time} or probabilistic models \cite{bailly2018crop}, which have been advantageously replaced by recurrent \cite{russwurm2018convolutional, garnot2019time, ozgur2021crop}, convolutional \cite{pelletier2019temporal, m2019semantic, ji20183d}, or differential \cite{metzger2020crop} architectures.
\begin{figure*}[ht!]
    \centering
    \includegraphics[width=1\textwidth, trim=0cm 18cm 19cm 0cm, clip]{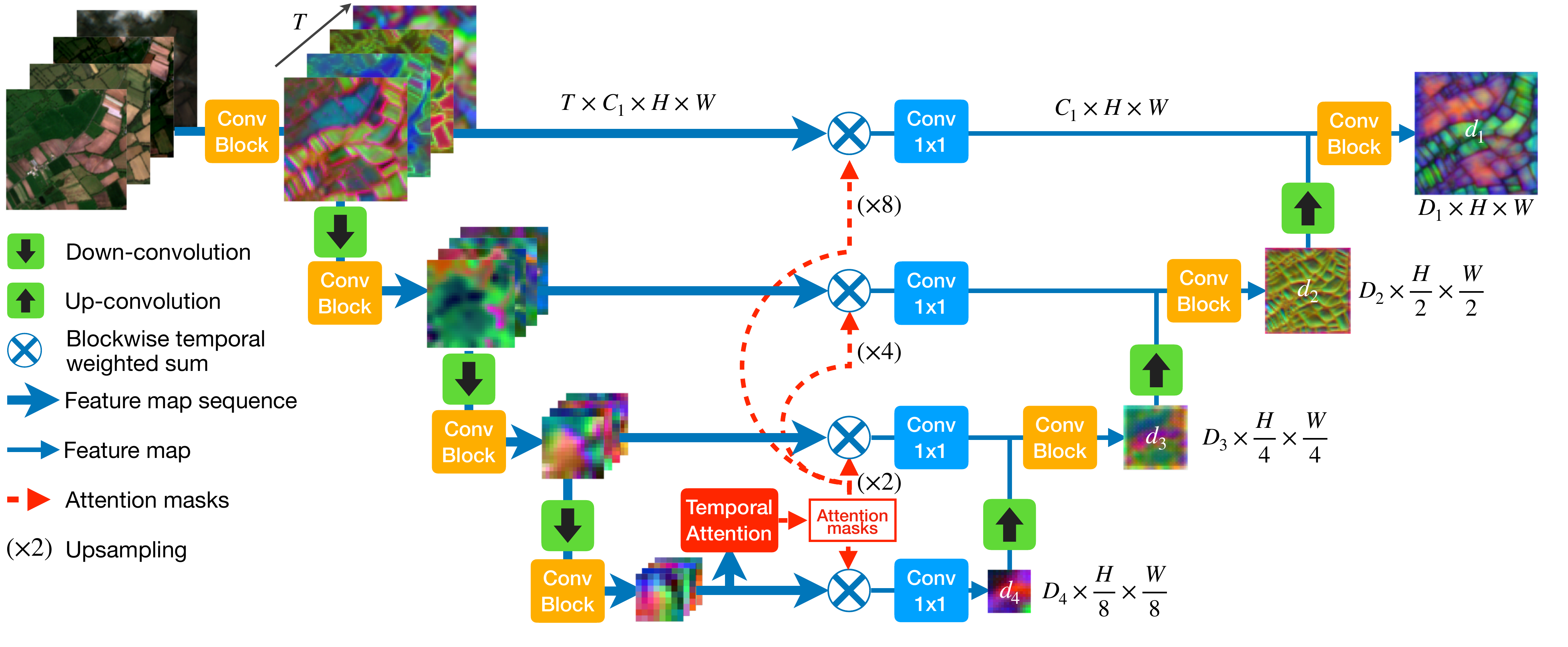}
    \caption{\textbf{Spatio-temporal Encoding.} A sequence of images is processed in parallel by a shared convolutional encoder. At the lowest resolution, an attention-based temporal encoder produces a set of temporal attention masks for each pixel, which are then spatially interpolated at all resolutions. These masks are used to collapse the temporal dimension of the feature map sequences into a single map per resolution. A convolutional decoder then computes features at all resolution levels. All convolutions operate purely on the spatial and channel dimensions, and we use strided convolutions for both spatial up and down-sampling. The feature maps are projected in RGB space to help visual interpretation. %In comparison to recurrent-based approaches, the attention masks allow for supervising the temporal feature extraction at all resolutions.
    }
    \label{fig:pipeline}
\end{figure*}
Recently, attention-based approaches have been adapted to encode sequences of remote sensing images and have led to significant progress for pixel-wise and parcel-wise classification \cite{garnot2020satellite, russwurm2020self, yuan2020self}.
In parallel, hybrid architectures \cite{stoian2019land, m2019semantic, papadomanolaki2021deep} relying on U-Net-type architectures \cite{ronneberger2015u} for encoding the spatial dimension and recurrent networks for the temporal dimension have shown to be well suited for the semantic segmentation of SITS. In this paper, we propose to combine this hybrid architecture with the promising temporal attention mechanism.
\paragraph{Instance Segmentation of Satellite Images.}
The first step of panoptic segmentation is to delineate all individual instances, { \ie instance segmentation}.
Most remote sensing instanciation approaches operate on a single acquisition. For example, %multiple instance segmentation
several methods have been proposed to detect individual instances of trees \cite{qin2014individual, zhao2018tree}, buildings \cite{wagner2020u}, or fields \cite{rieke2017deep}.  
Several algorithms start with a delineation step (border detection) \cite{garcia2017machine, masoud2020delineation, waldner2020deep}, and require postprocessing to obtain individual instances. Other methods use segmentation as a preprocessing step and compute cluster-based features \cite{censi2021spatial, derksen2018spatially}, but do not produce explicit cluster-to-object mappings.
Petitjean \etal \cite{petitjean2012spatio} propose a segmentation-aided classification method operating on image time series. However, their approach partitions each image separately and does not attempt to retrieve individual objects consistently across the entire sequence. In this paper, we propose the first end-to-end framework for directly performing joint semantic and instance segmentation on SITS.
\paragraph{Panoptic Segmentation of Videos.}
%Panoptic segmentation initially requires to individually delineate all relevant instances. 
Among the vast literature on instance segmentation, Mask-RCNN \cite{he2017mask} is the leading method for natural images. 
Recently, Wang \etal proposed CenterMask \cite{wang2020centerMASK}, a lighter and more efficient single-stage method which we use as a starting point in this paper.
Several approaches propose extending instance or panoptic segmentation methods from image to video \cite{yang2019video, tokmakov2019learning, kim2020video}. However, as explained in the introduction, SITS differs from natural video in several key ways which require specific algorithmic and architectural adaptations.

\section{Method}
%===================================================
%auto-ignore
We consider an image time sequence $X$, organized into a four-dimensional tensor of shape $T \times C \times H \times W$, with $T$ the length of the sequence, $C$ the number of channels, and $H \times W$ the spatial extent. 

\subsection{Spatio-Temporal Encoding}
Our model, dubbed U-TAE (U-Net with Temporal Attention Encoder), encodes a sequence $X$ in three steps:
\textbf{(a)} each image in the sequence is embedded simultaneously and independently by a shared multi-level spatial convolutional encoder, 
\textbf{(b)} a temporal attention encoder collapses the temporal dimension of the resulting sequence of feature maps into a single map for each level, 
\textbf{(c)} a spatial convolutional decoder produces a single feature map with the same resolution as the input images, see \figref{fig:pipeline}.
%\begin{itemize}
%    \item[\textbf{(a)}] each image in %the sequence is embedded %simultaneously and independently %by a shared multi-level spatial %convolutional encoder, 
%    \item[\textbf{(b)}]  a temporal %attention encoder collapses the %temporal dimension of the %resulting sequence of feature %maps into a single map for each %level, 
%    \item[\textbf{(c)}] a spatial %convolutional decoder produces a %single feature map with the same %resolution than the input images, %see \figref{fig:pipeline}.
%\end{itemize}

\paragraph{a) Spatial Encoding.}
We consider a convolutional encoder $\Enc$ with $L$ levels $1, \cdots, L$. Each level is composed of a sequence of convolutions, Rectified Linear Unit (ReLu) activations, and normalizations.
Except for the first level, each block starts with a strided convolution, dividing the resolution of the feature maps by a factor $2$.

For each time stamp $t$ simultaneously, the encoder $\Enc_l$ at level $l$ takes as input the feature map of the previous level $e^{l-1}_t$, and outputs a feature map $e^l_t$ of size $C_l \times {H_l} \times {W_l}$ with $H_l=H / 2^{l-1}$ and  $W_l=W /2^{l-1}$. The resulting feature maps are then temporally stacked into a feature map sequence $e^l$ of size $T \times C_l \times {H_l} \times {W_l}$:
\begin{align}
    e^{l} &= [\Enc_l(e^{l-1}_t)]_{t=0}^{T}\;\;\text{for}\;\; l \in [1, L]~,
\end{align}
%\noindent
with $e^0=X$ and $[\,\cdot\,]$ the concatenation operator along the temporal dimension.
When constituting batches, we flatten the temporal and batch dimensions. Since each sequence comprises images acquired at different times, the batches' samples are not identically distributed. 
To address this issue, we use Group Normalization \cite{wu2018group} with 4 groups instead of Batch Normalization \cite{ioffe2015batch} in the encoder.

\paragraph{b) Temporal Encoding.}
In order to obtain a single representation per sequence, we need to collapse the temporal dimension of each feature map sequence $e^l$  before using them as \emph{skip connections}.
Convolutional-recurrent U-Net networks \cite{stoian2019land, m2019semantic, papadomanolaki2021deep} only process the temporal dimension of the lowest resolution feature map with a temporal encoder. The rest of the skip connections are collapsed with a simple temporal average. This prevents the extraction of spatially adaptive and parcel-specific temporal patterns at higher resolutions.
Conversely, processing the highest resolution would result in small spatial receptive fields for the temporal encoder, and an increased memory requirement.
Instead, we propose an attention-based scheme which only processes the temporal dimension at the lowest feature map resolution, but is able to utilize the predicted temporal attention masks at all resolutions simultaneously.

Based on its performance and computational efficiency, we choose the Lightweight-Temporal Attention Encoder (L-TAE) \cite{garnot2020lightweight} to handle the temporal dimension.
The L-TAE is a simplified multi-head self-attention network \cite{vaswani2017attention} in which the attention masks are directly applied to the input sequence of vectors instead of predicted \emph{values}. Additionally, the L-TAE implements a channel grouping strategy similar to Group Normalization \cite{wu2018group}.
%Each head attends to a group of contiguous channels and predicts an attention mask highlighting the most relevant elements of the sequence.
%: the input vectors' channels are grouped into contiguous blocks of equal size, each processed by a different \emph{head}. Each head generates its own temporal attention mask highlighting the most relevant elements of the sequence for the channel group. These masks are then used to compute a weighted temporal average of the sequence for each feature block.
%The resulting vectors are concatenated channelwise into a single vector with as many channels as the input vectors.
%This grouping strategy allows for each attention head to encode specialized temporal patterns as it operates on distinct subsets of channels.
%\LOIC{The first step consists in grouping the input vectors' channels into contiguous blocks of equal size, or \emph{heads}.}Using a self-attention mechanism \cite{vaswani2017attention}, each head generates its own temporal attention mask expressing which elements of the sequence are the most relevant \LOIC{to its channels}. \LOIC{Each channel block is then averaged along the time dimension using the mask predicted by their corresponding head as weights. The resulting vectors are concatenated channelwise into a single vector with as many channels as the input vectors. This grouping strategy allows for each channel group to encode specialized temporal patterns.%, and results in improved computational efficiency and parameter utilization. }

We apply a shared L-TAE with $G$ heads independently at each pixel of $e^L$, the feature map sequence at the lowest level resolution $L$.
This generates $G$ temporal attention masks for each pixel, which can be arranged into $G$ tensors $a^{L,g}$ with values in $[0,1]$ and of shape $T \times {H}_L \times {W}_L$:
\begin{align}
    \label{eq:TAE_att}
    &a^{L,1}, \cdots, a^{L,G} = \LTAE(e^L)\;\text{, applied pixelwise.}
\end{align}
In order to use these attention masks at all scale levels $l$ of the encoder, we compute spatially-interpolated masks $a^{l,g}$ of shape ${T \times {H}_{l} \times {W}_{l}}$ for all $l$ in $[1, L-1]$ and $g$ in $[1, G]$ with bilinear interpolation:
\begin{align}
    \label{eq:TAE_att_resize}
    &a^{l,g} = \text{resize}\; a^{L,g} \;\text{to} \; H_l \times W_l~.
\end{align}
The interpolated masks $a^{l,g}$ at level $l$ of the encoder are then used as if they were generated by a temporal attention module operating at this resolution. We apply the L-TAE channel-grouping strategy at all resolution levels: 
the channels of each feature map sequence $e^l$ are split into $G$ contiguous groups $e^{l,1},\cdots,e^{l,G}$ of identical shape ${T \times} C_l / G \times W_l \times H_l$.
For each group $g$, the feature map sequence $e^{l,g}$ is averaged on the {temporal} dimension using $a^{l,g}$ as weights.
The resulting maps are concatenated along the channel dimension, and processed by a shared $1\times1$ convolution layer $\text{Conv}_{1 \times 1}^l$ of width $C_l$.
We denote by $f^l$ the resulting map of size ${C_l \times W_l \times H_l}$ by :
\begin{align}
    \label{eq:TAE_fea}
    &f^{l} = \text{Conv}_{1 \times 1}^l\left(\left[\sum_{t=1}^{T}a^{l,g}_t \odot e^{l,g}_t\right]_{g=1}^G\right)~,
\end{align}
with $[\,\cdot\,]$ the concatenation along the channel dimension and $\odot$ the term-wise multiplication with channel broadcasting.

\paragraph{c) Spatial Decoding.} 
We combine the feature maps $f^l$ learned at the previous step with a convolutional decoder to obtain spatio-temporal features at all resolutions. 
The decoder is composed of $L-1$ blocks $\Dec_l$ for $1 \leq l < L$, with  convolutions, ReLu activations, and BatchNorms \cite{ioffe2015batch}.
Each decoder block uses a strided transposed convolution $\Dec^{\text{up}}_l$  to up-sample the previous feature map.
%The decoder is composed of $L$ blocks $\Dec_l$, with  convolutions, ReLu activations, and BatchNorms \cite{ioffe2015batch}.
%For levels $l>1$ except the first one,
%we define as $\Dec^{\text{up}}_l$ a strided transpose convolution for up-sampling the previous feature map.
The decoder at level $l$ produces a feature map $d^l$ of size $D_l \times {H}_{l} \times {W}_l$. In a U-Net fashion, the encoder's map at level $l$ is concatenated with the output of the decoder block at level $l-1$:
\begin{align}
    % \\ 
    d^{l}  &= \Dec_l([ \Dec^{\text{up}}_l(d^{l+1}), f^{l}]) \;\text{for}\; l \in [1, L-1]~,
\end{align}
with $d^L=f^L$ and $[\,\cdot\,]$ is the channelwise concatenation.
\begin{figure}
\begin{tabular}{cc}
    \begin{subfigure}[t]{0.475\columnwidth}  
        \centering 
        \includegraphics[width=\textwidth, trim=1.5cm 1.5cm 0cm 0cm, clip]{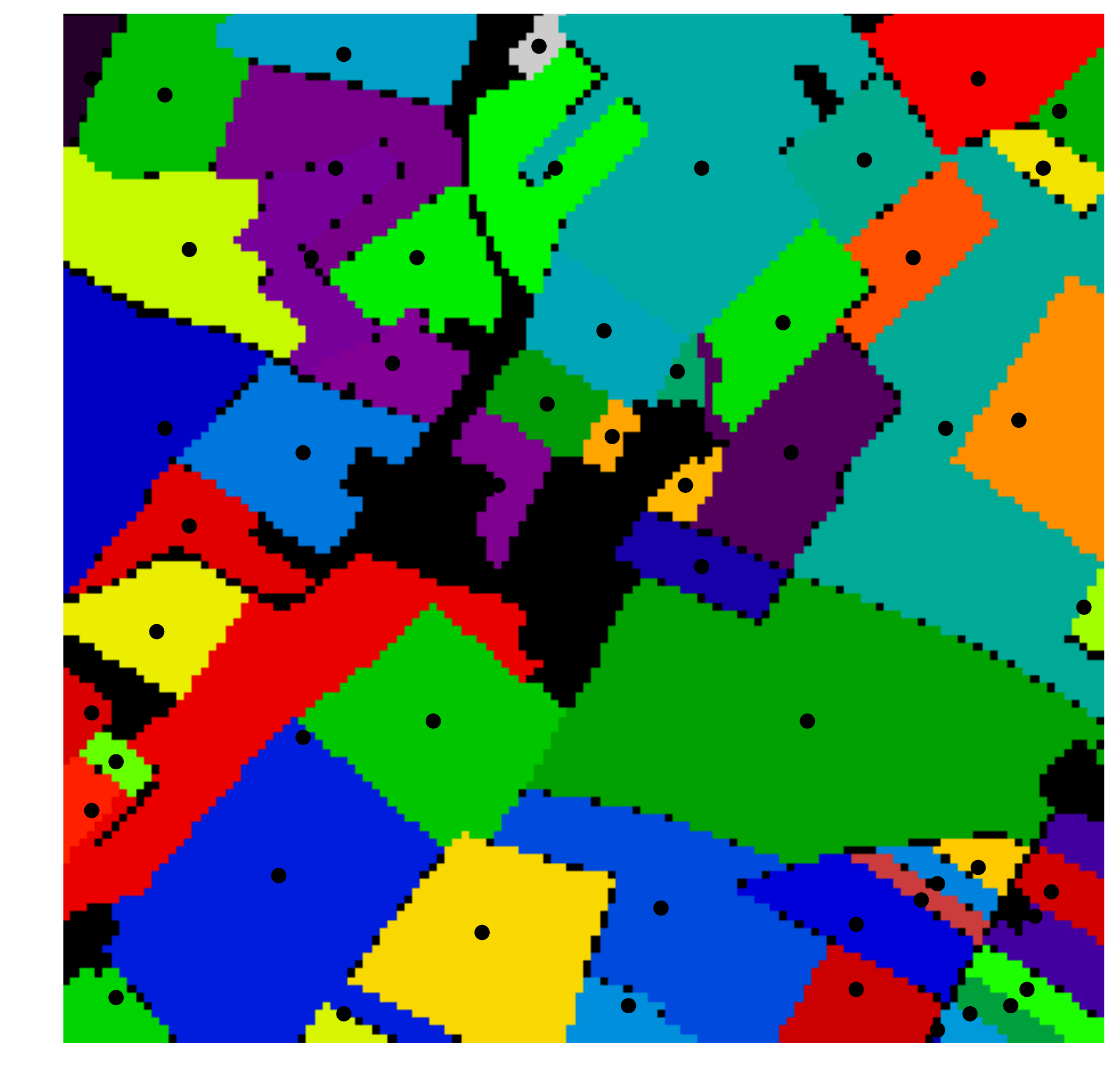}
        \caption{Instance masks}
        \label{fig:heatmap:instance}
    \end{subfigure}
    %\vfill
    &
    \begin{subfigure}[t]{0.475\columnwidth}  
        \centering 
        \includegraphics[width=\textwidth, trim=1.5cm 1.5cm 0cm 0cm, clip]{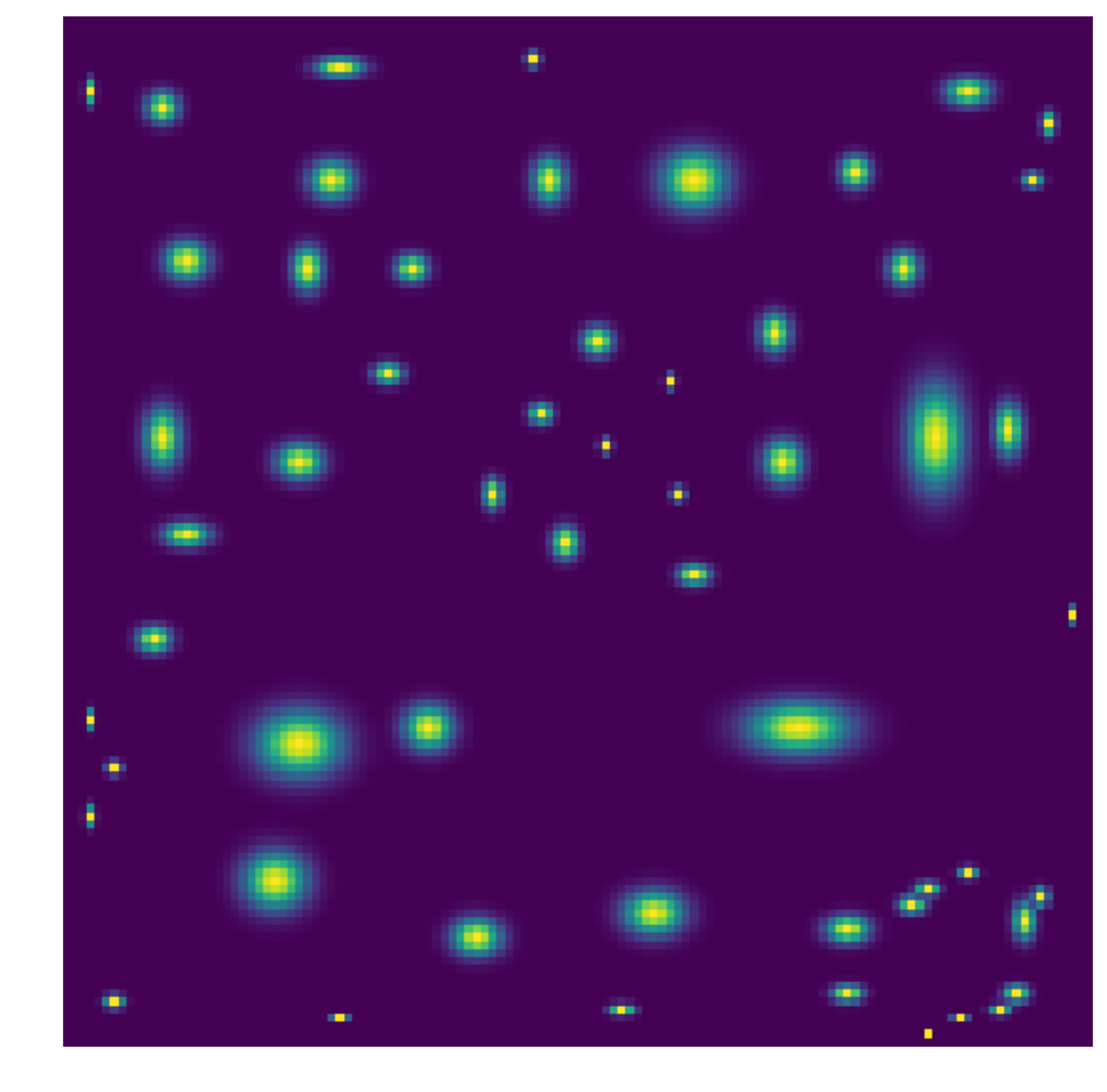}
        \caption{Target heatmap}
         \label{fig:heatmap:target}
    \end{subfigure}
    \\
        \begin{subfigure}[t]{0.475\columnwidth}  
        \centering 
        \includegraphics[width=\textwidth, trim=1.5cm 1.5cm 0cm 0cm, clip]{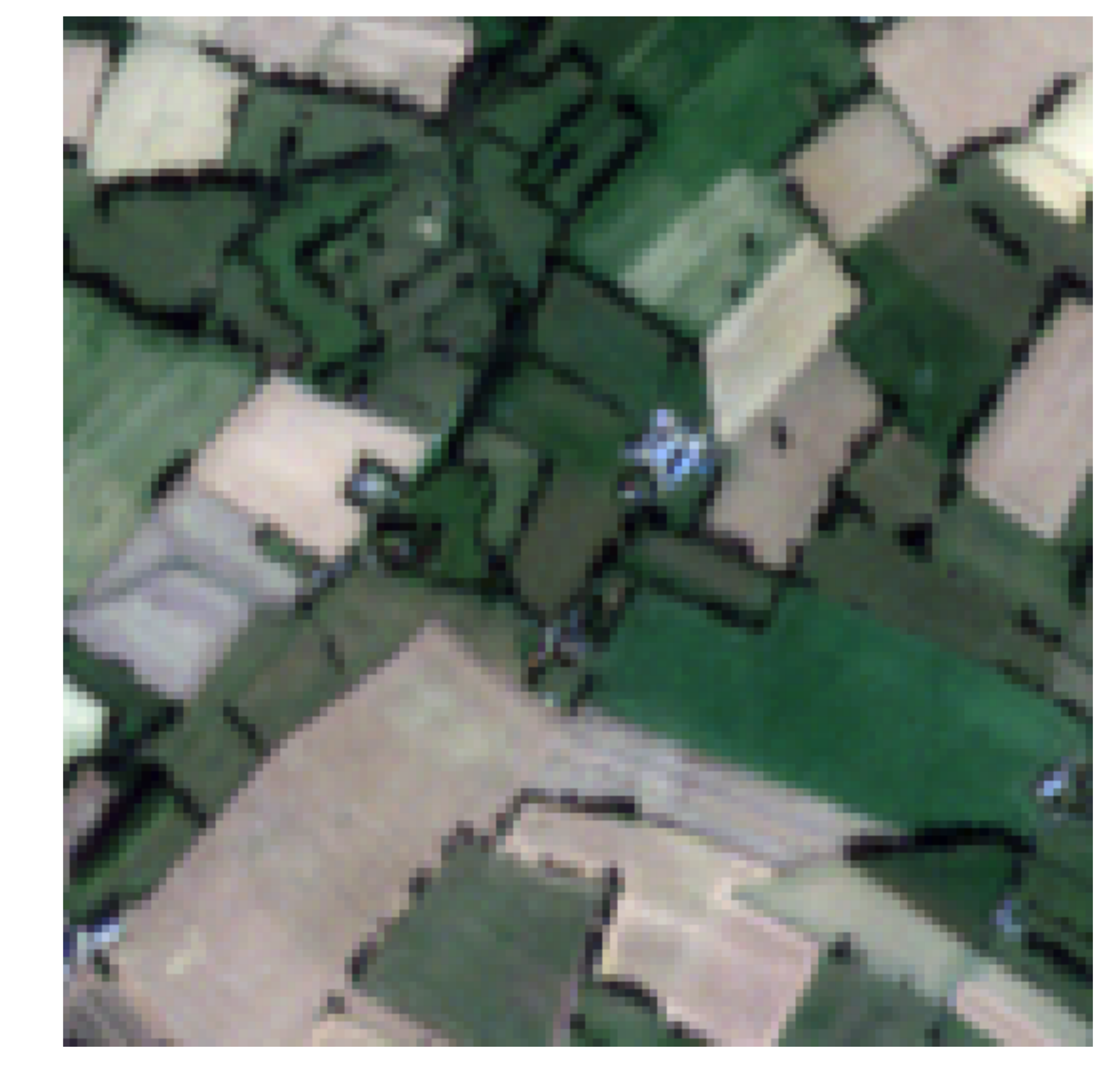}
        \caption{Observation from sequence.}
        \label{fig:heatmap:rgb}
    \end{subfigure}
    &
    \begin{subfigure}[t]{0.475\columnwidth}  
        \centering 
        \includegraphics[width=\textwidth, trim=1.5cm 1.5cm 0cm 0cm, clip]{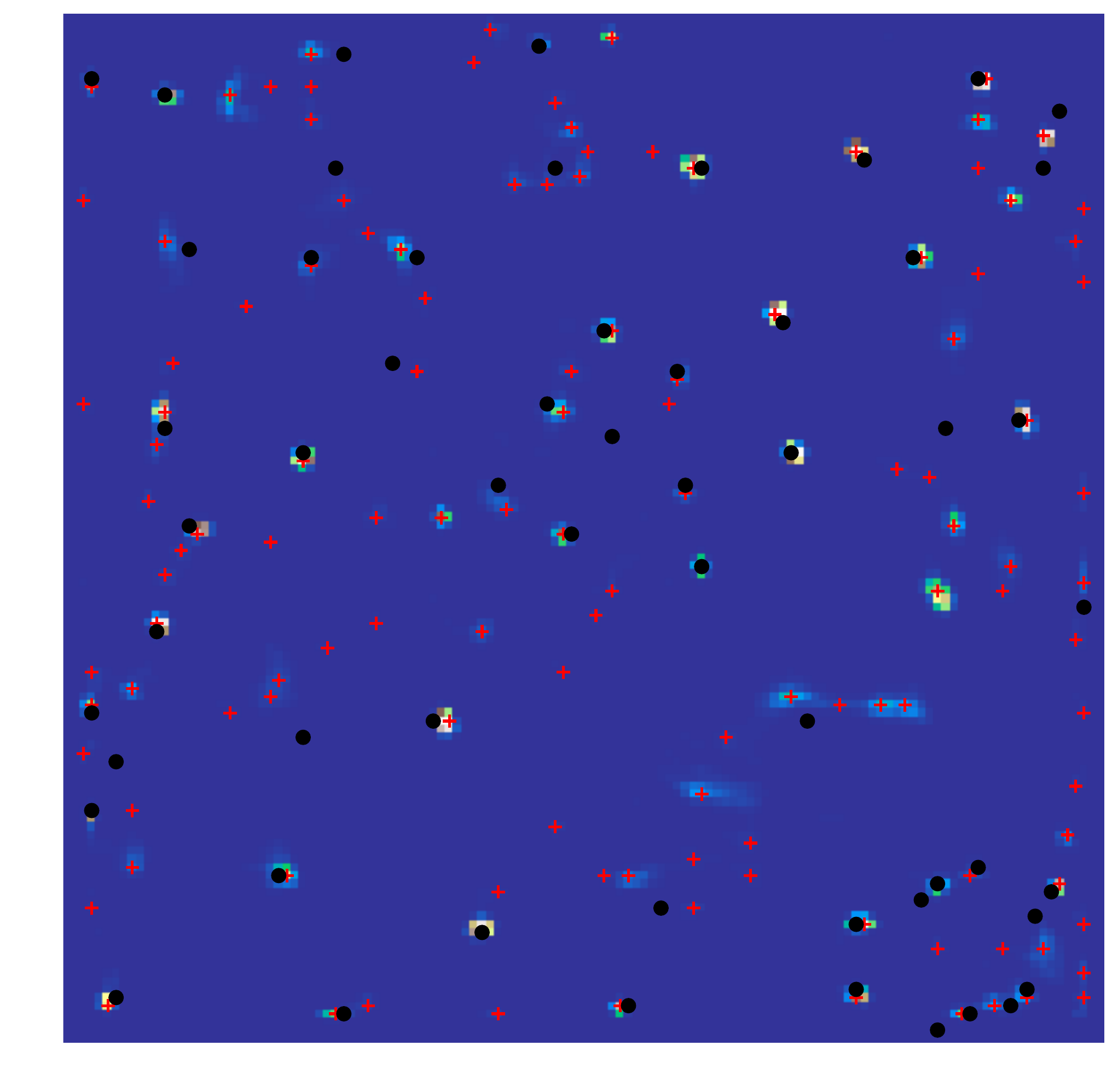}
        \caption{Predicted centerpoints}
        \label{fig:heatmap:prediction}
    \end{subfigure}
    \end{tabular}
    \caption{\textbf{Centerpoint Detection.} The ground truth instance masks \Subref{fig:heatmap:instance} is used to construct a target heatmap \Subref{fig:heatmap:target}. Our parcel detection module maps the raw sequence of observation \Subref{fig:heatmap:rgb} to a predicted heatmap \Subref{fig:heatmap:prediction}. The predicted centerpoints (red crosses) are the local maxima of the predicted heatmap \Subref{fig:heatmap:prediction}. The black dots are the true parcels centers. %\LOIC{Note the difficulty of identifying the violet instance (top middle in the ground truth mask \Subref{fig:heatmap:instance}) as a single parcel from a unique observation.}
    }
    \label{fig:heatmaps}
\end{figure}
\begin{figure*}[ht!]
    \centering
    \includegraphics[width=\linewidth, trim=0cm 23.5cm 28cm 0cm, clip]{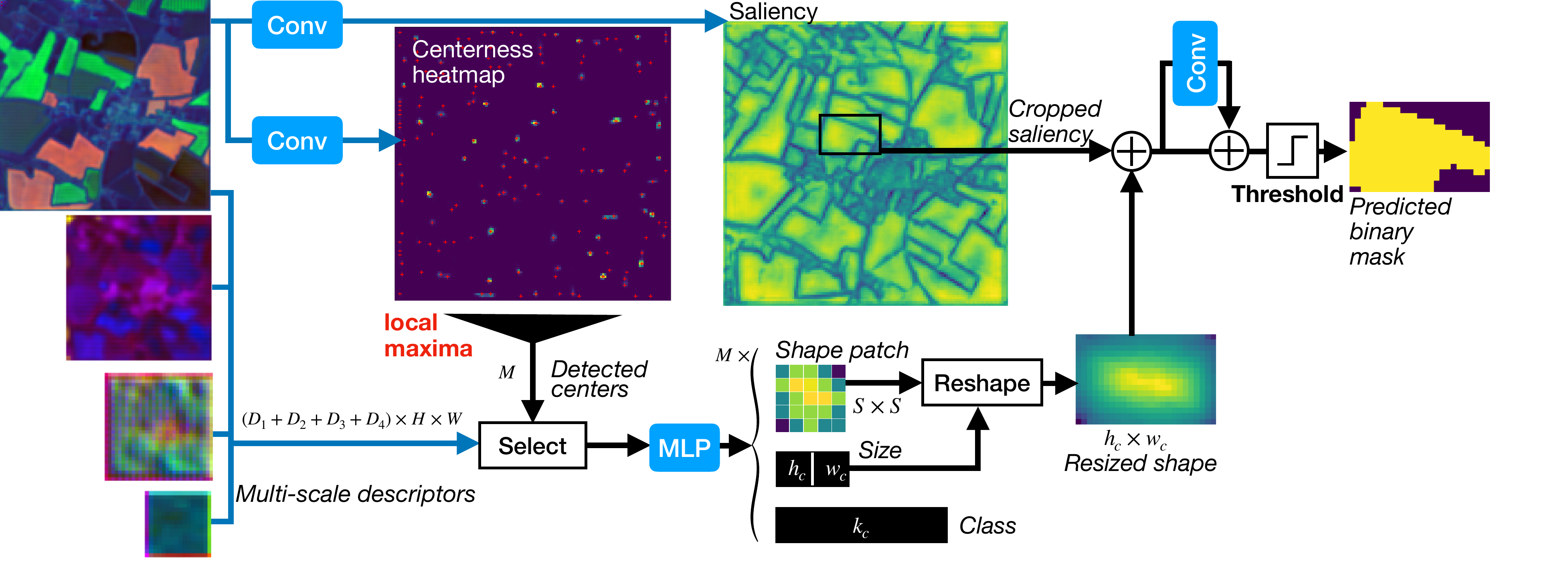}
    \caption{\textbf{Panoptic Segmentation.}
    The local maxima of the predicted centerness heatmap defines $M$ tentative parcels.
    For each one, the pixel features at all levels are concatenated and used to predict a bounding box size, a semantic class, and an $S\times S$ shape patch. The latter is combined with a global saliency map for predicting pixel-precise masks. The instance predictions are combined into a panoptic segmentation using the centerness as quality.}
    \label{fig:cropmask}
\end{figure*}
%=================================
\subsection{Panoptic Segmentation}
%=================================
%\vspace{-3cm}
%
Our goal is to use the multi-scale feature maps  $\{d^l\}_{l=1}^L$ learned by the spatio-temporal encoder to perform panoptic segmentation of a sequence of satellite images over an area of interest. 
The first stage of panoptic segmentation is to produce instance proposals, which are then combined into a single panoptic instance map.
Since an entire sequence of images (often over $50$) must be encoded to compute $\{d^l\}_{l=1}^L$, we favor an efficient design for our panoptic segmentation module. Furthermore,  given the relative simplicity of parcels' borders, we avoid complex region proposal networks such as Mask-RCNN. Instead, we adapt the single-stage CenterMask instance segmentation network \cite{wang2020centerMASK}, and detail our modifications in the following paragraphs. We name our approach \emph{Parcels-as-Points} (PaPs) to highlight our inspiration from CenterNet/Mask \cite{zhou2019centerNET, wang2020centerMASK}.

We denote by $P$ the set of {ground truth} parcels in the image sequence $X$. Note that the position of these parcels is time-invariant and hence
only defined by their spatial extent. Each parcel $p$ is associated with (i) a centerpoint $\hat{\imath}_p, \hat{\jmath}_p$ with integer coordinates, (ii) a bounding box of size $\hat{h}_p, \hat{w}_p$, 
(iii) a binary instance mask $\hat{\shape}_p \in \{0,1\}^{H \times W}$, 
(iv) a class $\hat{k}_p \in [1, K]$ with $K$ the total number of classes.
%
%
%
%\vspace{-0.8em}
\paragraph{Centerpoint Detection.} 
{Following} CenterMask, we perform parcel detection by predicting \emph{centerness heatmaps} supervized by the ground truth parcels' bounding boxes. In the original approach~\cite{zhou2019centerNET}, each class has its own heatmap:
detection doubles as classification. This is a sensible choice for natural images, since the tasks of detecting an object's nature, location, and shape are intrinsically related. In our setting however, the parcels' shapes and border characteristics are mostly independent of the cultivated crop. For this reason, we use a single centerness heatmap and postpone class identification to a subsequent specialized module. See \figref{fig:heatmaps} for an illustration of our parcel detection method.

We associate each parcel $p$ with a 
Gaussian kernel of deviations $\sigma^\text{ver}_p$ and $\sigma^\text{hor}_p$ taken respectively as $1/20$ of the height and width of the parcels' bounding box.
Unlike Law and Deng \cite{law2018cornernet}, we use heteroschedastic kernels to reflect the potential narrowness of parcels. 
We then define the target centerness heatmap $\hat{m} \in [0,1]^{H\times W}$ as the maximum value of all parcel kernels at each pixel $(i,j)$ in $H\times W$ :
\begin{align}\label{eq:kernel}
&\hat{m}_{i,j} = \max_{p \in P}\; \exp\left(-\left[\frac{(i-\hat{\imath}_p)^2}{2(\sigma^\text{ver}_p)^2}+\frac{(j-\hat{\jmath}_p)^2}{2(\sigma^\text{hor}_p)^2}\right] \right)
\end{align}
%see \eqref{eq:kernel}. 
%\VIVIEN{Note that this equation induces a direct mapping $\zeta : (i,j) \mapsto o$ ,  between each pixel position and the  ground truth  with highest kernel value at this position.}
%\LOIC{Replacing $\max$ operator by $\text{argmax}$ in \eqref{eq:kernel} defines a mapping $\zeta: H\times W \mapsto O$ associating each pixel with its \emph{closest} parcel.}
%\begin{align}\label{eq:kernel}
%&\hat{m}_{i,j} = \max_{o \in O} %\exp\left(-\left[\frac{(i-\hat{\i}_o)%^2}{2\sigma_\text{hor}^2}+\frac{(j-\h%at{\j}_o)^2}{2\sigma_\text{ver}^2}\ri%ght] \right)
%\end{align}
%
A convolutional layer takes the highest-resolution feature map $d^1$ as input and predicts a centerness heatmap ${m} \in [0,1]^{H\times W}$. 
The predicted heatmap is supervized using the loss defined in \eqref{eq:centerloss} with %$\alpha=2$ and 
$\beta=4$:
\begin{align}
&\cL_{\text{center}}\!=\! \frac{-1}{|P|}\!\!\sum_{\substack{i=1\cdots H \\j=1\cdots W}}\!\!\begin{cases}
\!\log({m}_{i,j}) \;\text{if}\; \hat{m}_{i,j}=1\\
\!(1\!-\!\hat{m}_{i,j})^\beta \log(1\!-\!{m}_{i,j}) \;\text{else.}\label{eq:centerloss}
    \end{cases}
\end{align}
%with $\alpha$ and $\beta$ set hyper-parameters chosen as $2$ and $4$ as in \cite{zhou2019centerNET}.
We define the predicted centerpoints as the local maxima of ${m}$, \ie pixels with larger values than their $8$ adjacent neighbors. This set can be efficiently computed with a single max-pooling operation. 
Replacing the $\max$ operator by $\argmax$ in \eqref{eq:kernel} defines a mapping $H\times W \mapsto P$ between pixels and  parcels.
During training, we associate each true parcel $p$ with the predicted centerpoint $\bc(p)$ with highest predicted centerness ${m}$ among the set of centerpoints  which coordinates are mapped to $p$.
If this set is empty, then $\bc(p)$ is undefined: the parcel $p$ is not detected. We denote by $P'$ the subset of detected parcels, \ie for which $\bc(p)$ is well defined.
%\VIVIEN{During training, each ground truth object is uniquely associated with the centerpoint with the highest predicted centerness $\hat{m}$ out of the centerpoints that are mapped to this object by $\zeta$. We denote by $\bc(o)$ the centerpoint thus associated with object $o$ when it exists, and by $O'$ the subset of ground truth objects for which such a centerpoint exists.}
%During training, each ground truth object is uniquely associated with the centerpoint within its %bounding box\VIVIEN{ground truth gaussian kernel with the highest predicted centerness $\hat{m}$,}
%the highest predicted centerness
%as defined in \equaref{eq:kernel}.
%Conversely, %\VIVIEN{the highest Gaussian kernel at a given position} this defines a mapping between each predicted centerpoint $p$ and a unique object $\text{obj}(p)$. Note that an object can potentially be associated with no predicted centerpoints, and vice versa. Centerpoints without objects are filtered out for the next steps.
%\LOIC{Conversely, we denote by $P'$ the set of centerpoints selected this way.
%During inference, as no ground truth is available, $P'$ is the set of all predicted centerpoints.}
%During inference, during which no ground truth object is available, we filter out predicted center with centerness under $0.2$.
%
%\begin{figure}[ht!]
%    \centering
%    \includegraphics[width=1\columnwidth]{images/Fig_He%atmapv2.pdf}
%    \caption{\textbf{Shape Prediction.}
%    Shows the resize shape patch, the cropped saliency %mask, the local shape and the cropped ground truth %masks.}
%    \label{fig:saliency}
%\end{figure}
%
\paragraph{Size and Class Prediction.}
{We associate to a predicted centerpoint $c$ of coordinate $({i_c},{j_c})$ the multi-scale feature vector $\tilde{d}_c$ of size $D_1 +\cdots +D_L$ by concatenating channelwise the pixel features at location $({i_c},{j_c})$ in all  maps $d^l$:}
%We start by computing a multi-scale feature vector for each selected centerpoint $p\in P'$ of coordinate $({i_p},{j_p})$. The pixel-wise feature vectors corresponding to location $({i_p},{j_p})$ at all resolutions of the decoder are concatenated along the channel dimension into a multi-scale descriptor $\tilde{d}_p$ of size $D_1 +\cdots +D_L$:  
\begin{align}
\tilde{d}_c=
\left[
  d^l\left(
    \left\lfloor
      {{i_c}}/{2^{l-1}}
    \right\rfloor
    ,
    \left\lfloor
      {{j_c}}/{2^{l-1}}
    \right\rfloor
  \right)
 \right]_{l=1}^L~,
 \end{align}
with $[\,\cdot\,]$ the channelwise concatenation.
This vector $\tilde{d}_c$ is then processed by four different multilayer perceptrons (MLP) to obtain three vectors of sizes $2$, $K$, and $S^2$ representing respectively:
%This vector $\tilde{d}_p$ is mapped by \VIVIEN{four  of respective output sizes $2$, $N$, $S^2$, and $1$ to predict 
(i) a bounding box size ${h}_c, {w}_c$, (ii) a vector of class probabilities ${k}_c$ of size $K$, and (iii) a shape patch ${\shape}_c$ of fixed size $S\times S$. The latter is described in the next paragraph.

%We denote by $\text{obj}(p)$ the ground truth object associated with the predicted centerpoint $p$. 
%\VIVIEN{During training, the class prediction $n_{\bc(o)}$ for each detected ground truth object is supervized with the cross-entropy loss,  }
The class prediction $k_{\bc(p)}$ associated to the true parcel $p$ is supervized with the cross-entropy loss,
and the size prediction with a normalized L1 loss. For all $p$ in $P'$, we have:
%a loss inspired by Redmon \etal \cite{redmon2016you}:
\begin{align}
    &\cL_{\text{class}}^p=-%(1-[k_{\text{obj}(p)}])^\gamma
    \log(k_{\bc(p)}[\hat{k}_{p}]) \\ 
    &\cL_{\text{size}}^p\!=\!\frac{|h_{\bc(p)} - \hat{h}_{p}|}{\hat{h}_{p}}+\frac{|w_{\bc(p)} - \hat{w}_{p}|}{\hat{w}_{p}}~.
    %&\cL_{\text{size}}^p\!=\! %\left(\sqrt{\vphantom{\hat{h}_{\text{obj}(p)}}{h}_p}\!-\sqrt{\vphantom{\h%at{h}_{\text{obj}(p)}}\hat{h}_{\text{obj}(p)}}\right)^2\!\!\!\!+\!
    %\left(\sqrt{\vphantom{\hat{h}_{\text{obj}(p)}}{w}_p}\!-\sqrt{\vphantom{\h%at{h}_{\text{obj}(p)}}\hat{w}_{\text{obj}(p)}}\right)^2\!\!.
\end{align}
\paragraph{Shape Prediction.} 
The idea of this step is to combine for a predicted centerpoint $c$ a rough shape patch ${\shape}_c$ with a full-resolution global saliency map ${z}$ to obtain a pixel-precise instance mask, see \figref{fig:cropmask}.
For a centerpoint $c$ of coordinates $(i_c,j_c)$, the predicted shape patch ${\shape}_c$ of size ${S\times S}$ is resized to the predicted size $\lceil {h}_c \rceil \times \lceil {w}_c \rceil$ with bilinear interpolation.
A convolutional layer maps the outermost feature map $d^1$ to a saliency map ${z}$ of size $H\times W$, which is shared by all predicted parcels. 
This saliency map is then cropped along the predicted bounding box $(i_c,j_c,\lceil {h}_c \rceil,\lceil {w}_c \rceil)$.
The resized shape and the cropped saliency are added (\ref{eq:mask:localtilde}) to obtain a first local shape $\tilde{l}_c$, which is then further refined with a residual convolutional network  $\text{CNN}$ (\ref{eq:mask:local}). We denote the resulting predicted shape by $l_c$:
\begin{align}\label{eq:mask:localtilde}
    \tilde{l}_c &= \text{resize}_c({s}_c)
        +
        \text{crop}_c({z})\\\label{eq:mask:local}
     {l}_c &= \sigmoid(\tilde{l}_c + \text{CNN}(\tilde{l}_c))~, 
\end{align}
\noindent with $\text{resize}_c$ and $\text{crop}_c$ defined by the coordinates $(i_c,j_c)$ and predicted bounding box size $(\lceil {h}_c \rceil,\lceil {w}_c\rceil)$.
%\VIVIEN{and combined with the resized shape. Instead of using simple pointwise multiplication as in CenterMask, we propose to use a small residual convolutional network to  obtain the predicted  shape mask ${l}_p \in [0,1]^{\lceil {h}_p \rceil,\lceil {w}_p \rceil}$, see \equaref{eq:mask:local}.}
%, and multiplied pointwise with the resized shape to obtain the predicted  shape mask ${l}_p \in [0,1]^{\lceil {h}_p \rceil,\lceil {w}_p \rceil}$, see \equaref{eq:mask:local}.
%
The shape and saliency predictions are supervized 
for each parcel $p$ in $P'$
%\VIVIEN{for each detected ground truth object}
by computing the pixelwise binary cross-entropy (BCE) between the predicted shape ${l}_{\bc(p)}$ and the corresponding true binary instance mask $\hat{s}_{p}$ cropped along the predicted bounding box $(i_{\bc(p)},j_{\bc(p)},\lceil {h}_{\bc(p)} \rceil,\lceil {w}_{\bc(p)} \rceil)$:
\begin{align}\label{eq:mask:localsupervized}
   \cL_{\text{shape}}^p &= \text{BCE}({l}_{\bc(p)}, \text{crop}_{\bc(p)}(\hat{s}_{p}))~.
\end{align}
\noindent
For inference, we associate a binary mask with a predicted centerpoint $c$ by thresholding ${l}_c$ with the value $0.4$.
%
%\paragraph{Quality Prediction.} Since our goal is panoptic prediction, we need a reliable quality score to combine potentially overlapping predicted masks. Similarly to the method proposed by \cite{redmon2016you}, our network associates to a predicted centerpoint $c$ a quality ${q}_c \in [0,1]$. We supervize ${q}_c$ with a binary target $\hat{q}_p$,
%equals to one if the Intersection over Union (IoU) of the predicted and ground truth masks is greater than $0.5$ and zero otherwise.
%that is positive if the Intersection over Union (IoU) of the predicted and ground truth masks is superior to $0.5$ and negative otherwise}
%\begin{align}
%\cL_{\text{conf}}^p = \text{BCE}({q}_{\bc(p)}, \hat{q}_p)~.
%\end{align}

\paragraph{Loss Function}:
These four losses are combined into a single loss with no weight and optimized end-to-end:
\begin{align}
    \cL =
    \cL_\text{center} +
    \frac1{|P'|} \sum_{p \in P'}\left(
    \cL_\text{class}^p +
    \cL_\text{size}^p +
    \cL_\text{shape}^p
    \right).
\end{align}
%\begin{align}
%    \cL \!=\! 
%    \cL_\text{center} \!+\!
%    \frac1{|P'|} \!\sum_{p \in P'} \!\!\left(
%    \cL_\text{class}^p \!\!+\!
%    \cL_\text{size}^p \!+\!
%    \cL_\text{shape}^p\!+\!
%    \cL_{\text{conf}}^p\right).
%\end{align}
%
\paragraph{Differences with CenterMask.} Our approach differs from CenterMask in several key ways: 
(i) We compute a single saliency map and heatmap instead of $K$ different ones. This represents the absence of parcel occlusion and the similarity of their shapes.
(ii) Accounting for the lower resolution of satellite images, centerpoints are computed at full resolution to detect potentially small parcels, thus dispensing us from predicting offsets. 
(iii) The class prediction is handled centerpoint-wise instead of pixel-wise for efficiency. 
(iv) Only the selected centerpoints predict shape, class, and size vectors, saving computation and memory. 
(v) We use simple feature concatenation to compute multi-scale descriptors instead of deep layer aggregation \cite{yu2018deep} or stacked Hourglass-Networks \cite{newell2016stacked}.
%(vi) The quality associated with a prediction is learned instead of being directly defined by the predicted heatmap. 
(vi) A convolutional network learns to combine the saliency and the mask instead of a simple term-wise product.
\begin{figure*}[ht!]
    \centering
        \begin{subfigure}{0.24\textwidth}
        \centering
        \begin{tikzpicture}
        \node[anchor=south west,inner sep=0] (image) at (0,0) {\includegraphics[width=\textwidth]{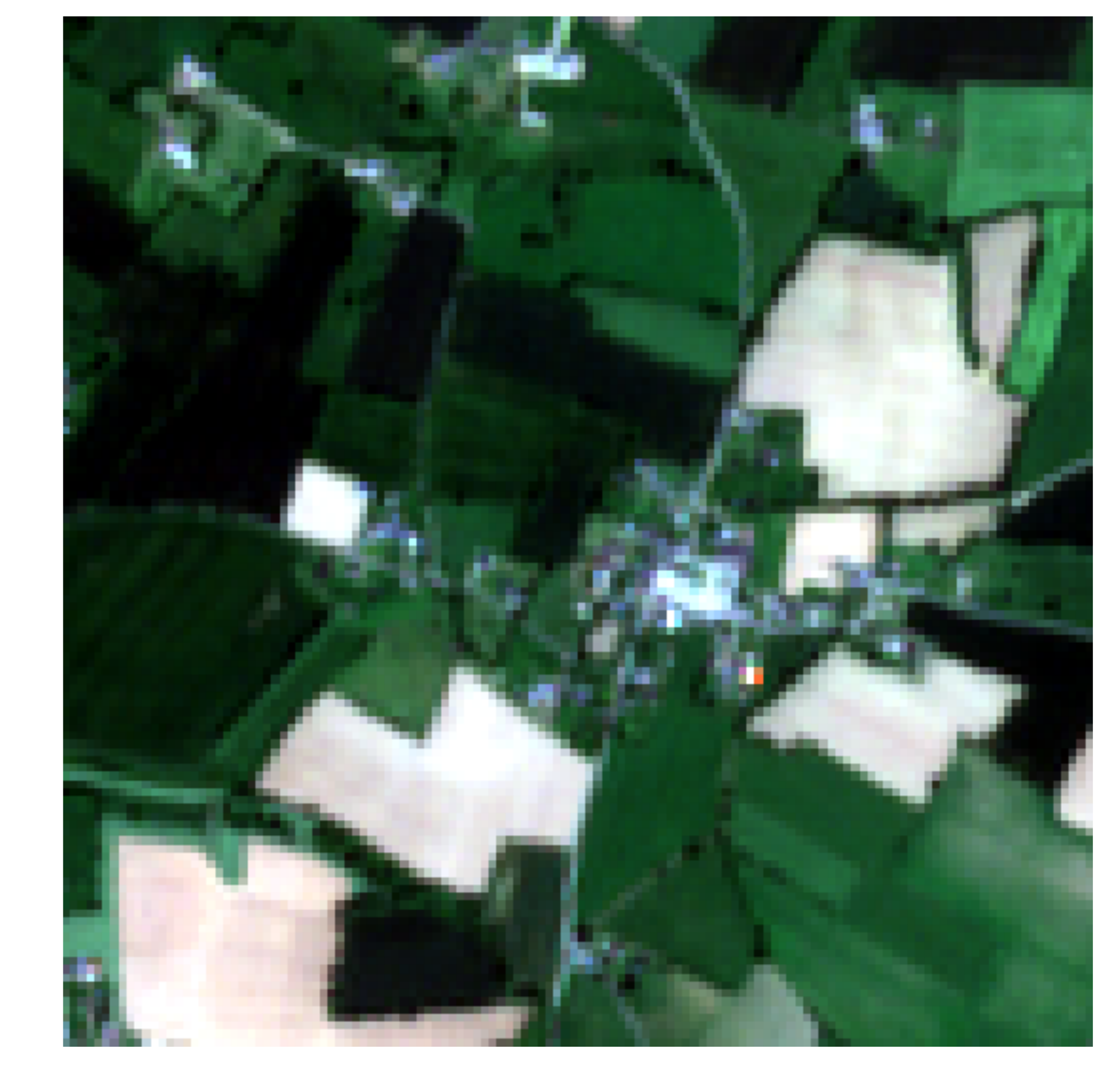}};
        \begin{scope}[x={(image.south east)},y={(image.north west)}]
        \draw[cyan,ultra thick] (0.22,0.7) circle (0.17);
        \draw[red,ultra thick] (0.65,0.15) circle (0.10);
    \end{scope}
\end{tikzpicture}
        
        \caption{Image from the sequence.}
        \label{fig:quali:rgb}
        \end{subfigure}
    \hfill
        \begin{subfigure}{0.24\textwidth}
        \includegraphics[width=\textwidth]{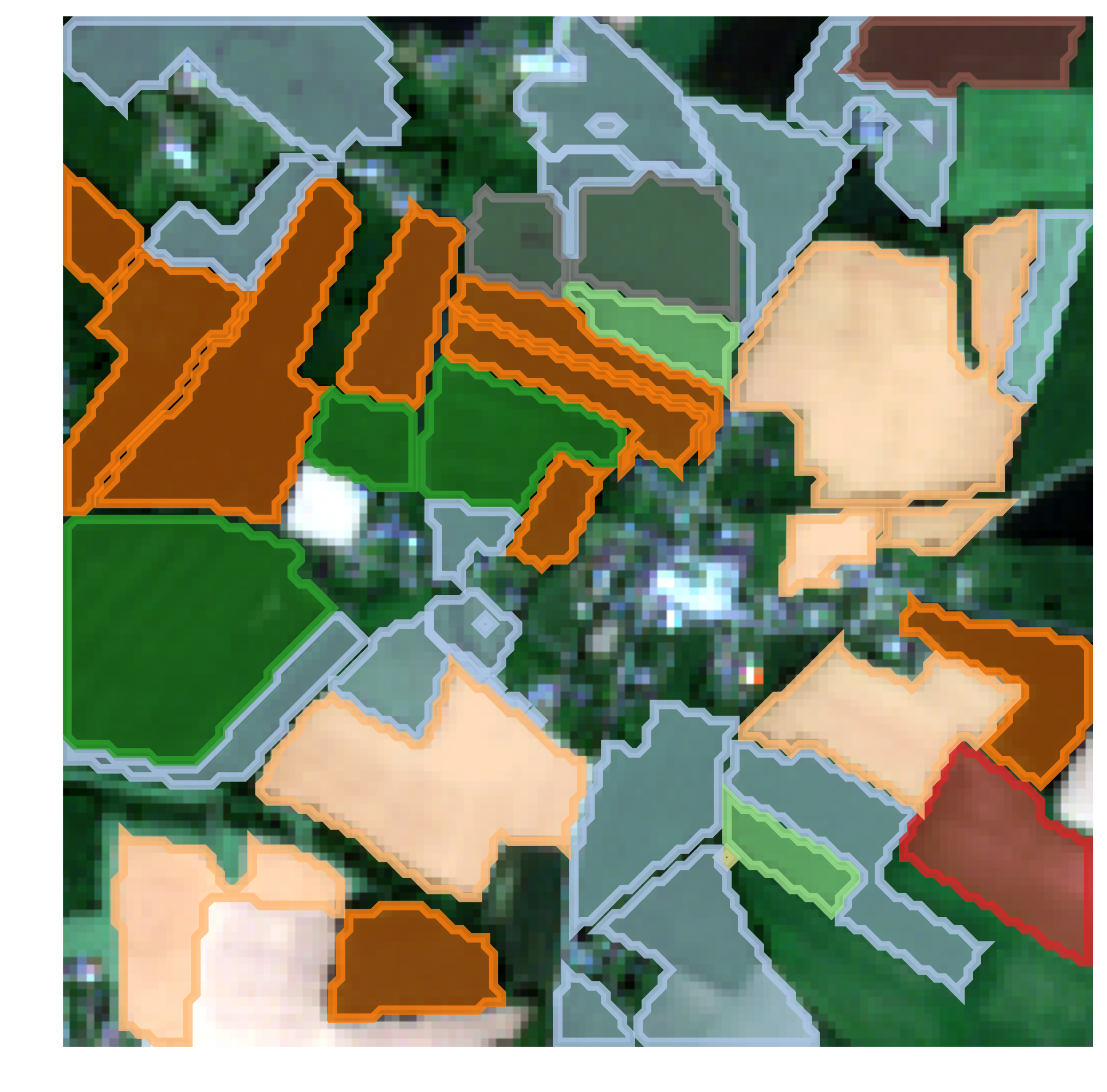}
        \caption{Panoptic annotation.}
        \label{fig:quali:gt}
        \end{subfigure}
    \hfill
        \begin{subfigure}{0.24\textwidth}
        \includegraphics[width=\textwidth]{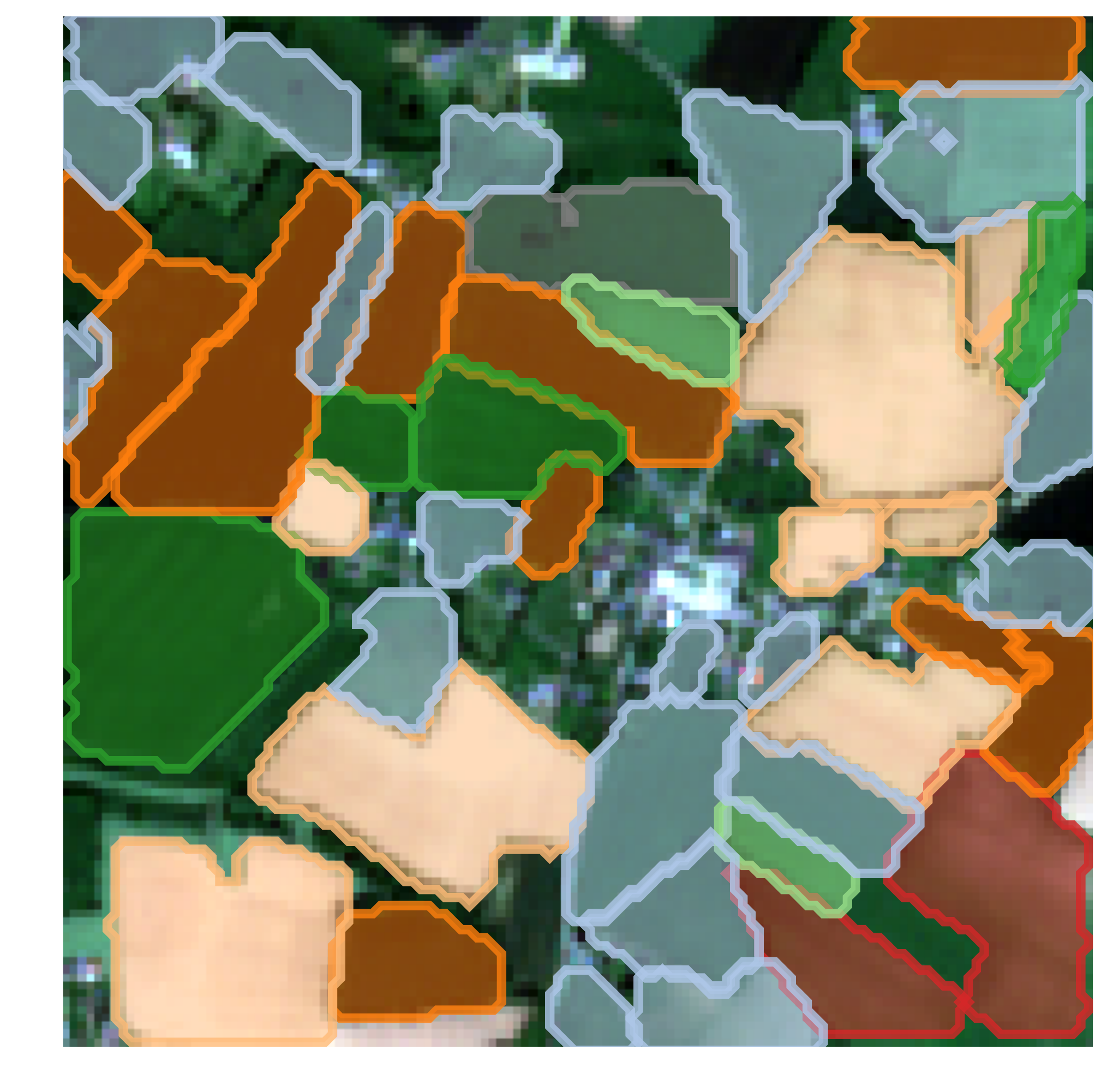}
        \caption{Panoptic segmentation.}
        \label{fig:quali:pan}
        \end{subfigure}
    \hfill
        \begin{subfigure}{0.24\textwidth}
        \includegraphics[width=\textwidth]{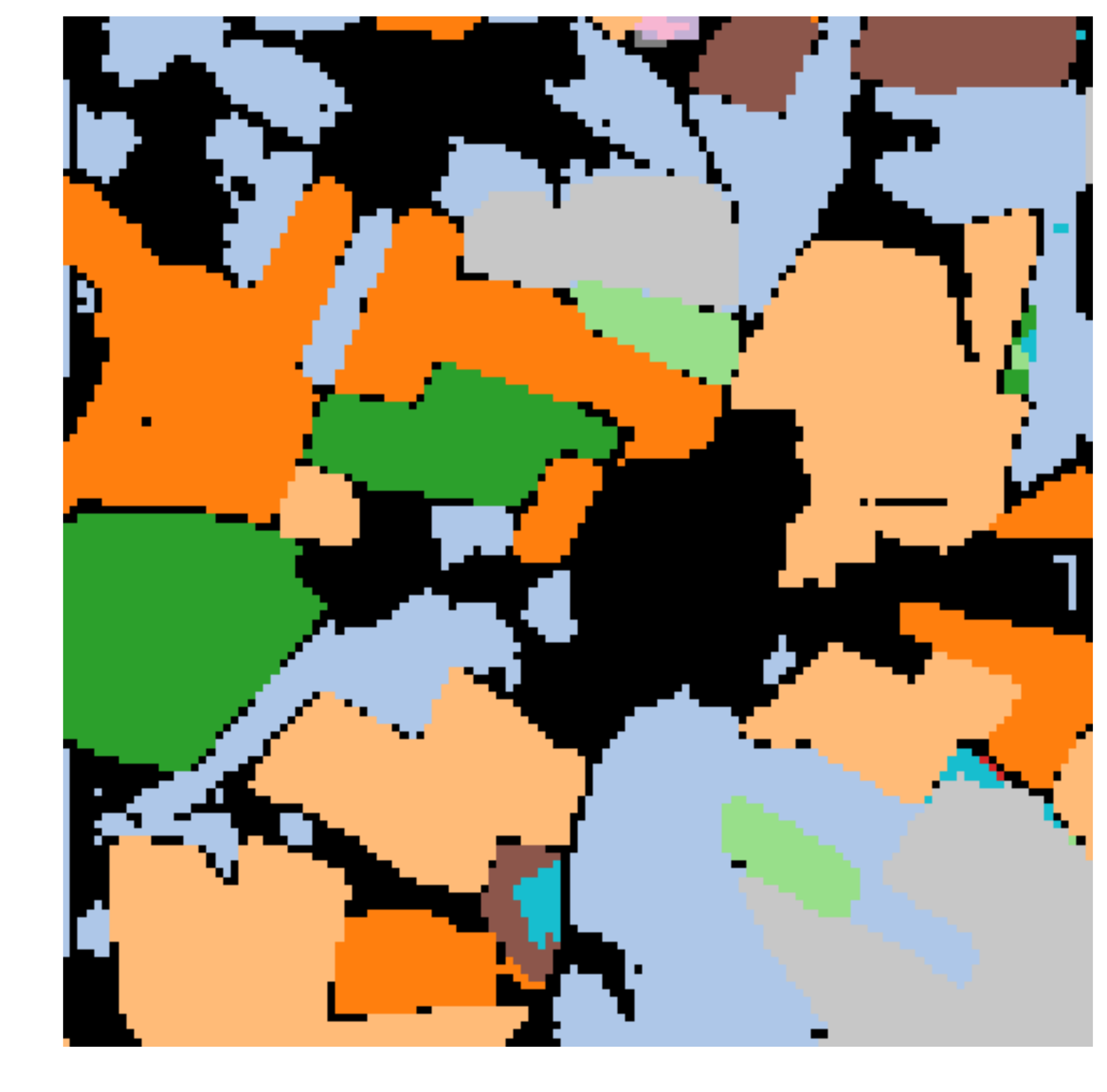}
        \caption{Semantic segmentation.}
        \label{fig:quali:seg}
        \end{subfigure}
    \caption{\textbf{Qualitative results.} We consider an image sequence \Subref{fig:quali:rgb} with panoptic annotations \Subref{fig:quali:gt}. We represent the results of our method in terms of  panoptic segmentation \Subref{fig:quali:pan} and semantic segmentation \Subref{fig:quali:seg}. The parcels' and pixels' color corresponds to the crop type, according to a legend given in the appendix. The predominantly correct class predictions highlight the fact that the difficulty of panoptic segmentation lies in the precise delineation of each individual parcel.
    We observe cases where the temporal structure of the SITS was successfully leveraged to resolve boundary ambiguities that could not be seen from a single image
    (cyan circle \protect\tikz \protect\node[circle, thick, draw = cyan, fill = none, scale = 0.7] {};). Conversely, some visually fragmented parcels are annotated as a single instance (red circle \protect\tikz \protect\node[circle, thick, draw = red, fill = none, scale = 0.7] {};).
    }
    \label{fig:quali}
\end{figure*}

\paragraph{Converting to Panoptic Segmentation.}
Panoptic segmentation consists of associating to each pixel a semantic label and, for non-background pixels (our only \emph{stuff} class), an instance label \cite{kirillov2019panoptic}.
Our predicted binary instance masks can have overlaps, which we resolve by associating to each predicted parcel a quality measure equal to the predicted centerness $m$ at its associated centerpoint. Masks with higher quality overtake the pixels of overlapping masks with lesser predicted quality. If a mask loses more than $50\%$ of its pixels through this process, it is removed altogether from the predicted instances. Predicted parcels with a quality under a given threshold are dropped. This threshold can be tuned on a validation set to maximize the parcel detection F-score. All pixels not associated with a parcel mask are labelled as background.

%based on the predicted quality: masks with higher quality overtake the pixels of overlapping masks with lesser predicted quality. If a mask loses more than $50\%$ of its pixels through this process, it is removed altogether from the predicted instances. Predicted parcels with a quality under a given threshold are dropped. This threshold can be tuned on a validation set to maximize the parcel detection F-score. All pixels not associated with a parcel mask are labelled as background.
%All pixels not associated with an instance label when all predicted  with a quality of over $0.15$ have been processed are identified as background. 

%The confidence of a prediction can be directly defined as the centerness of a predicted centerpoints. However, the centerness is not directly supervized to represent the confidence of a prediction. We thus define the confidence of a prediction by multiplying the centerness of the centerpoint with the contrast of the predicted mask. We define this contrast as the difference between the average values of the pixels above and the pixels below the binary threshold ($0.4$) in the predicted masks. Such measure of contrast should favor parcels with clear-cut borders.
%\VIVIEN{Since centerness does not directly encode for the quality of the predicted masks, we found it beneficial to multiply it by the contrast of the predicted mask, defined as the difference between the average values of the positive and negative zones of the predicted instance masks.}
%
\paragraph{Implementation Details.}
Our implementation of U-TAE allows for batch training on sequences of variable length thanks to a simple padding strategy.
The complete configuration and training details can be found in the Appendix. 
A Pytorch implementation is available at \url{https://github.com/VSainteuf/utae-paps}. 
%
%In all experiments, we use a U-TAE with $L=4$ resolution levels and a $G=16$-head LTAE. When using the PaPs module, we take shape patches of size $S=8$. See appendix for the exact configuration.
%Across our experiments, we use Adam %\cite{kingma2014adam} optimizer with %default parameters and a batch size of %$4$ sequences. The semantic segmentation %experiments use a fixed learning rate of %$0.001$ for $100$ epochs. For the %panoptic segmentation experiments, we %start with a higher learning rate of %$0.01$ for $50$ epochs, and decrease it %to $0.001$ for the last $50$ epochs.
%Our network is implemented in PyTorch, use Adam \cite{kingma2014adam} optimizer with default parameters, a batch size of $4$ sequences and a learning rate of $0.001$ for $200$ epochs. See appendix for the exact configurations.

%==================================================
\section{Experiments}
%==================================================

%auto-ignore
%-------------------------------------------
\subsection{The PASTIS Dataset}
%-------------------------------------------
%We are not aware of publicly available datasets allowing for evaluating panoptic or instance segmentation methods operating on SITS. Therefore, 
We present PASTIS (Panoptic Agricultural Satellite TIme Series), the first large-scale, publicly available SITS dataset with both semantic and panoptic annotations.
This dataset, as well as more information about its composition, are publicly available at \url{https://github.com/VSainteuf/pastis-benchmark} .

\paragraph{Description.} 
PASTIS is comprised of $2\,433$ sequences of multi-spectral images of shape $10 \times 128 \times 128$. Each sequence contains between $38$ and $61$ observations taken between September $2018$ and November $2019$, for a total of over 2 billion pixels. The time between acquisitions is uneven with a median of $5$ days. This lack of regularity is due to the automatic filtering of acquisitions with extensive cloud cover
%of tiles entirely covered by clouds
by the satellite data provider \href{https://www.theia-land.fr/en/product/sentinel-2-surface-reflectance/}{THEIA}. 
%\VIVIEN{Note that our implementation of U-TAE %and PaPs allows for training in batches of sequences of variable lengths.}
The $10$ channels correspond to the {non-atmospheric} spectral bands of the Sentinel-2 satellite, after atmospheric correction and re-sampling at a spatial resolution of $10$ meters per pixel. The dataset spans over $4000$ km$^2$, with images taken from four different regions of France with diverse climates and crop distributions, covering almost $1\%$ of the French Metropolitan territory.
We estimate that close to $28$\% of images have at least partial cloud cover.
%\VIVIEN{A visual inspection of $600$ images finds that
%close to $26$\% of images are at least partially covered by clouds and $18\%$ are totally obstructed}. 
%The uncompressed size of the dataset is approximately $40$GB.

\paragraph{Annotation.}
Each pixel of PASTIS is associated with a semantic label taken from a nomenclature of $18$ crop types plus a background class. As is common in remote sensing applications, the 
dataset is highly unbalanced, with a ratio of over $50$ between the most and least common classes.
%classes are imbalanced, with $50$ times more samples between the most and least populated classes. }
Each non-background pixel also has a unique instance label corresponding to its parcel index. In total, $124\,422$ parcels are individualized, each with their bounding box, pixel-precise mask, and crop type.
All annotations are taken from the publicly available French Land Parcel Identification System. The French Payment Agency estimates the accuracy of the crop annotations via in situ control over $98$\% and the relative error in terms of surfaces  under $0.3$\%. To allow for cross-validation, the dataset is split into $5$ folds, chosen with a $1$km buffer between images to avoid  cross-fold contamination.
%\paragraph{Ghana} We also evaluated the semantic segmentation performance of ours  and several other competing approaches on the publicly available Ghana dataset \cite{ghana} used by Rustowicz \etal \cite{m2019semantic}. It comprises XXX parcels for which we have a sequence of Sentinel-2 superspetral imaaages and a label among $4$ crop types. For better visual results, we added a fifth background class.
%----------------------------------------
\subsection{Semantic Segmentation}
%----------------------------------------
\label{sec:xp:segsem}
\begin{table}[t]
    \caption{\textbf{Semantic Segmentation.} We report for our method and six competing methods the model size in trainable parameters, Overall Accuracy (OA), mean Intersection over Union (mIoU), and Inference Time for one fold of $\sim490$ sequences (IT). The second part of the table report results from our ablation study.}
    \label{tab:semseg}
    \centering
    \begin{tabular}{lrrrr}
    \toprule
    \multirow{2}{*}{Model}
    &\small  \#~param & \multirow{2}{*}{OA}& \multirow{2}{*}{mIoU}& \multirow{2}{*}{IT (s)}
    \\
    & $\times 1000$
    \\\cmidrule{1-5}
    \small U-TAE (ours)        &   1\,087   &    \textbf{83.2}   &          \textbf{63.1}  & \bf 25.7\\
    \small 3D-Unet \cite{m2019semantic}        &  1\,554    &  81.3     &    58.4   & 29.5     
    \\
    \small U-ConvLSTM  \cite{m2019semantic}       &   1\,508 & 82.1   &   57.8   &    28.3
    \\
    \small FPN-ConvLSTM \cite{martinez171fully} & 1\,261      &    81.6     &  57.1   &   103.6    
    \\
    \small U-BiConvLSTM  \cite{martinez171fully}       &  1\,434    &    81.8   &       55.9  & 32.7  
    \\
    \small ConvGRU \cite{ballas2015delving}        &  1\,040    &    79.8   & 54.2 &    49.0       
    \\
    \small ConvLSTM \cite{russwurm2018convolutional, shi2015convolutionallstm}        &  1\,010 & 77.9   &49.1  & 49.1
    \\\midrule
    \small Mean Attention  &  1\,087  & 82.8 & 60.1 & 24.8\\
    \small Skip Mean + Conv &  1\,087 & 82.4 & 58.9& 24.5 \\
    \small Skip Mean & 1\,074  & 82.0 & 58.3& 24.5\\
    \small BatchNorm &  1\,087  & 71.9 & 36.0& 22.3\\ 
    \small Single Date (August) & 1\,004  & 65.6 & 28.3 & {1.3}\\ 
    \small Single Date (May) & 1\,004  & 58.1 & 20.6 & {1.3}\\ 
    \bottomrule
    \end{tabular}
\end{table}
Our U-TAE has $L=4$ resolution levels and a LTAE with $G=16$ heads, see appendix for an exact configuration. For the semantic segmentation task, the feature map $d_1$ with highest resolution is set to have $K$ channels, with $K$ the number of classes.  We can then interpret $d_1$ as pixel-wise predictions to be supervized with the cross-entropy loss. In this setting, we do not use the PaPs module.
\paragraph{Competing Methods.} 
We reimplemented six of the top-performing SITS encoders proposed in the literature:
\begin{itemize}
    \item %\textbf{Convolutional-Recurrent Architectures}: 
    \emph{ConvLSTM} \cite{russwurm2018convolutional, shi2015convolutionallstm} and \emph{ConvGRU}\cite{ballas2015delving}.
    These approaches are recurrent neural networks in which all linear layers are replaced by spatial convolutions.
    %\vspace{-.2cm}
    \item %\textbf{UNet-Based Architectures}:
    \emph{U-ConvLSTM} \cite{m2019semantic} and \emph{U-BiConvLSTM} \cite{martinez171fully}.
    To reproduce these UNet-Based architectures, we replaced the L-TAE in our architecture by either a convLSTM \cite{shi2015convolutional} or a bidirectional convLSTM. Skip connections are temporally averaged. In contrast to the original methods, we replaced the batch normalization in the encoders with group normalization which significantly improved the results across-the-board.
    %\vspace{-.2cm}
    \item 
    \emph{3D-Unet} \cite{m2019semantic}. A U-Net in which the convolutions of the encoding branch are three-dimensional to handle simultaneously the spatial and temporal dimensions.
    %\vspace{-.2cm}
    \item  %\textbf{Pyramid-Based Architectures:}
    \emph{FPN-ConvLSTM} \cite{martinez171fully}. This model combines a feature pyramid network \cite{lin2017feature} to extract spatial features and a bidirectional ConvLSTM for the temporal dimension. 
\end{itemize}
\paragraph{Analysis.} In \tabref{tab:semseg}, we detail the  performance obtained with $5$-fold cross validation of our approach and the six reimplemented baselines. We report the Overall Accuracy (OA) as the ratio between correct and total predictions, and (mIoU) the class-averaged classification IoU. We observe that the convolutional-recurrent methods \emph{ConvGRU} and \emph{ConvLSTM} perform worse. Recurrent networks embedded in an U-Net or a FPN  share similar performance, with a much longer inference time for FPN. Our approach significantly outperforms all other methods in terms of precision. In \figref{fig:quali}, we present a qualitative illustration of the semantic segmentation results.
%===================================================================
\paragraph{Ablation Study.}
%===================================================================
We first study the impact of using spatially interpolated attention masks to collapse the temporal dimension of the spatio-temporal feature maps at different levels of the encoder simultaneously. Simply computing the temporal average of skip connections for levels without temporal encoding as proposed by \cite{stoian2019land,m2019semantic}, we observe a drop of $4.8$ mIoU points (Skip Mean). This puts our method performance on par with its competing approaches. Adding a $1 \times 1$ convolutional layer after the temporal average reduces this drop to $4.2$ points (Skip Mean + Conv). 
Lastly, using interpolated masks but foregoing the channel grouping strategy by averaging the masks group-wise into a single attention mask per level results in a drop of $3.1$ points (Mean Attention). This implies that our network is able to use the grouping scheme at different resolutions simultaneously. 
In conclusion, the main advantage of our proposed attention scheme 
is that the temporal collapse is controlled at all resolutions, in contrast to recurrent methods.
%is that it can be used to supervize the temporal collapse at all resolutions, in contrast to recurrent methods.

Using batch normalization in the encoder leads to a severe degradation of the performance of $27.1$ points (BatchNorm). We conclude that the temporal diversity of the acquisitions requires special considerations. This was observed for all U-Net models alike.
We also train our model on a single acquisition date (with a classic U-Net and no temporal encoding) for two different cloudless dates in August and May (Single Date). We observe a drop of $24.8$ and $42.5$ points respectively, highlighting the crucial importance of the temporal dimension for crop classification. We also observed that images with at least partial cloud cover received on average $58\%$ less attention than their cloud-free counterparts. This suggests that our model is able to use the attention module to automatically filter out corrupted data.
%\begin{itemize}
%    \item \emph{Mean Skip}:
%     We average temporally the skip connection instead of %using the interpolated attention masks.
%     \item \emph{Mean Skip + Conv}:
%     Same as \emph{Mean Skip}, but we add a $1\times1$ %convolution for all skip connections.
%     \item \emph{Mean Attention Skip}:
%     We average group-wise the attention masks outputted %by the L-TAE before up or downsampling them, %breaking the head structure.
%     \item \emph{BatchNorms:} We replace the group %normalization of the encoders with batch %normalization.
%    \item \emph{Single Image:} We select a single %cloudless acquisition in May or August and run the %U-Net using a single encoder, \ie the classical U-Net %architecture.
%\end{itemize}
%We report the impact of these design choices in %\tabref{tab:ablation}. Surprisingly, we observe that %batch normalization leads to a severe degradation of the %results. We conclude that the temporal diversity of the %acquisition requires special considerations. This was %observed for all models alike.
%We observe that our inter-level attention scheme is %responsible for a large part of the \emph{ConvTAE} %performance. Interestingly, keeping the group structure %inherited from the TAE at all resolution levels is %beneficial. Using a single image leads to severe %performance  degradation, which is very dependant on the %chosen date.
%===================================================================
\subsection{Panoptic Segmentation}
%===================================================================
\begin{table}[t]
    \caption{\textbf{Panoptic Segmentation Experiment.} We report class-averaged panoptic metrics: SQ, RQ, PQ (see Metric Correction paragraph before references). }
    \label{tab:instance_seg}
    \centering
    {
    \begin{tabular}{lccc}
    \toprule
        &SQ&RQ&PQ
    \\\cmidrule{1-4}
    \textbf{U-TAE + PaPs} & 81.5 & 53.2 & 43.8 \\
%    {Unet-3d + PaPs} && & \\
    {U-ConvLSTM + Paps} & 80.2 & 43.9 & 35.6\\\midrule
    %{No Quality} & -1.5 & +0.1 & -0.6 \\
    {$S=24$} &80.7 & 50.6&41.3\\
    {$S=8$} &80.9 & 52.3 & 42.7 \\
    {Multiplicative Saliency} &74.6 &49.9 & 37.5 \\
    {Single-image} & 72.3 & 18.7 & 14.1 \\
    \bottomrule           
    \end{tabular}
    }
\vspace{-.4cm}
\end{table}
We use the same U-TAE configuration for panoptic segmentation, and select a PaPs module with $190$k parameters and a shape patch size of $16\times16$.
In \tabref{tab:instance_seg}, we report the  class-averaged Segmentation Quality (SQ), Recognition Quality (RQ), and Panoptic Quality (PQ)~\cite{kirillov2019panoptic}.
We observe that while the network is able to correctly detect and classify most parcels, the task remains difficult. In particular, the combination of ambiguous borders and hard-to-classify parcel content makes for a challenging panoptic segmentation problem.
We illustrate these difficulties in \figref{fig:quali}, along with qualitative results.

Replacing the temporal encoder by a U-BiConvLSTM as described in \secref{sec:xp:segsem} (U-BiConvLSTM+PaPs), we observe a noticeable performance drop of $8.2$ PQ, which is consistent with the results of \tabref{tab:semseg}. 
%{Training the model with different values for $S$ shows a low sensitivity to this hyper-parmeter. }
As expected, our model's performance is not too sensitive to changes in the size $S$ of the shape patch. Indeed, the shape patches only determine the rough outline of parcels while the pixel-precise instance masks are derived from the saliency map.
Performing shape prediction with a simple element-wise multiplication as in \cite{wang2020centerMASK} (Multiplicative Saliency) instead of our residual CNN results in a drop of over $-6.9$ SQ.
%\VIVIEN{Using simple element-wise multiplication instead of our residual CNN for shape prediction results in a decrease of $4.9$PQ, largely caused by a poorer segmentation performance $-7.1$ SQ.}
%{Directly using the predicted centerness as quality  results in a drop of $0.6$ PQ.}
%Our definition of the prediction quality as the combined centerness and contrast of the instance mask accounts for a $1.1$PQ increase compared to a model simply using centerness.
Using a single image (August) leads to a low panoptic quality. Indeed, identifying crop types and parcel borders from a single image at the resolution of Sentinel-2 is particularly difficult.

Inference on $490$ sequences takes $129$s: $26$s to generate U-TAE embeddings, $1$s for the heatmap and saliency, $90$s for instance proposals, and $12$s to merge them into a panoptic segmentation. Note that the training time is also doubled compared to simple semantic segmentation. 
\FloatBarrier
%==================================================
\section{Conclusion}
%==================================================
We introduced U-TAE, a novel spatio-temporal encoder using a combination of spatial convolution and temporal attention. This model can be easily combined with \emph{PaPs}, the first panoptic segmentation framework operating on SITS. Lastly, we presented PASTIS, the first large-scale panoptic-ready SITS dataset. Evaluated on this dataset, our approach significantly outperformed all other approaches for semantic segmentation, and set up the first state-of-the-art for panoptic segmentation of satellite image sequences.

We hope that the combination of our open-access dataset and promising results will encourage both remote sensing and computer vision communities to consider the challenging problem of panoptic SITS segmentation, whose economic and environmental stakes can not be understated.

\newpage

\section*{Metric Correction}
The values reported in this version of the article for the Panoptic Segmentation experiment differ from the version published in the ICCV 2021 proceedings. Indeed, a bug in the computation of the Recognition Quality (RQ) metric was present in the original implementation resulting in the \emph{void} target instances not being properly ignored. Instead, all predictions matched to \emph{void} target instances were counted as false positives, thus artificially reducing the RQ score.
Since the panoptic metrics are not involved in the training loss, this bug did not impact the overall training procedure. All models of \tabref{tab:instance_seg} were re-evaluated with the corrected implementation. Across methods this resulted in a $\sim3$ PQ increase, driven by a similar increase in RQ. Refer to \url{github.com/VSainteuf/utae-paps/issues/11} for more details. 

\balance
{\small
\bibliographystyle{ieee_fullname}
\bibliography{convsa}
}

\newpage
\nobalance

\ARXIV{
\section*{\Large Supplementary Material}
\renewcommand{\thesection}{A.\arabic{section}}
\setcounter{section}{0}
%auto-ignore
In this appendix, we provide additional information on the PASTIS dataset and our exact model configuration. We also provide complementary qualitative experimental results.

\section{PASTIS Dataset}

\begin{figure}[ht!]
    \centering
    \begin{subfigure}{\linewidth}
    \centering
    \includegraphics[width=.78\linewidth,trim=0cm 3cm 0cm 3cm, clip, angle=90]{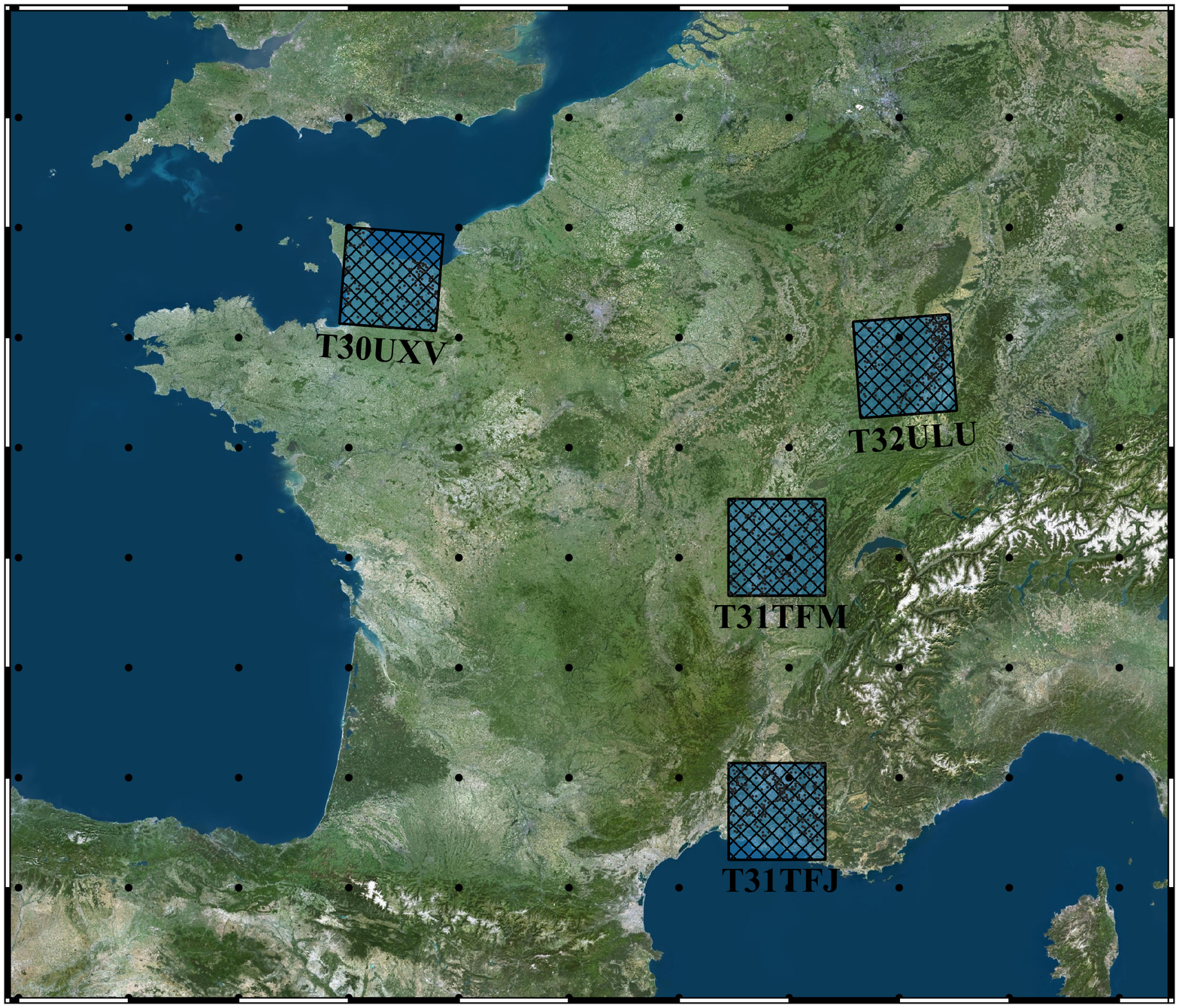}
    \caption{Location of the four tiles.}
    \label{fig:data:global}
    \end{subfigure}
    \vfill
    \begin{subfigure}{0.48\linewidth}
    \centering
    \includegraphics[width=\linewidth, trim=0cm 0cm 9cm 0cm, clip]{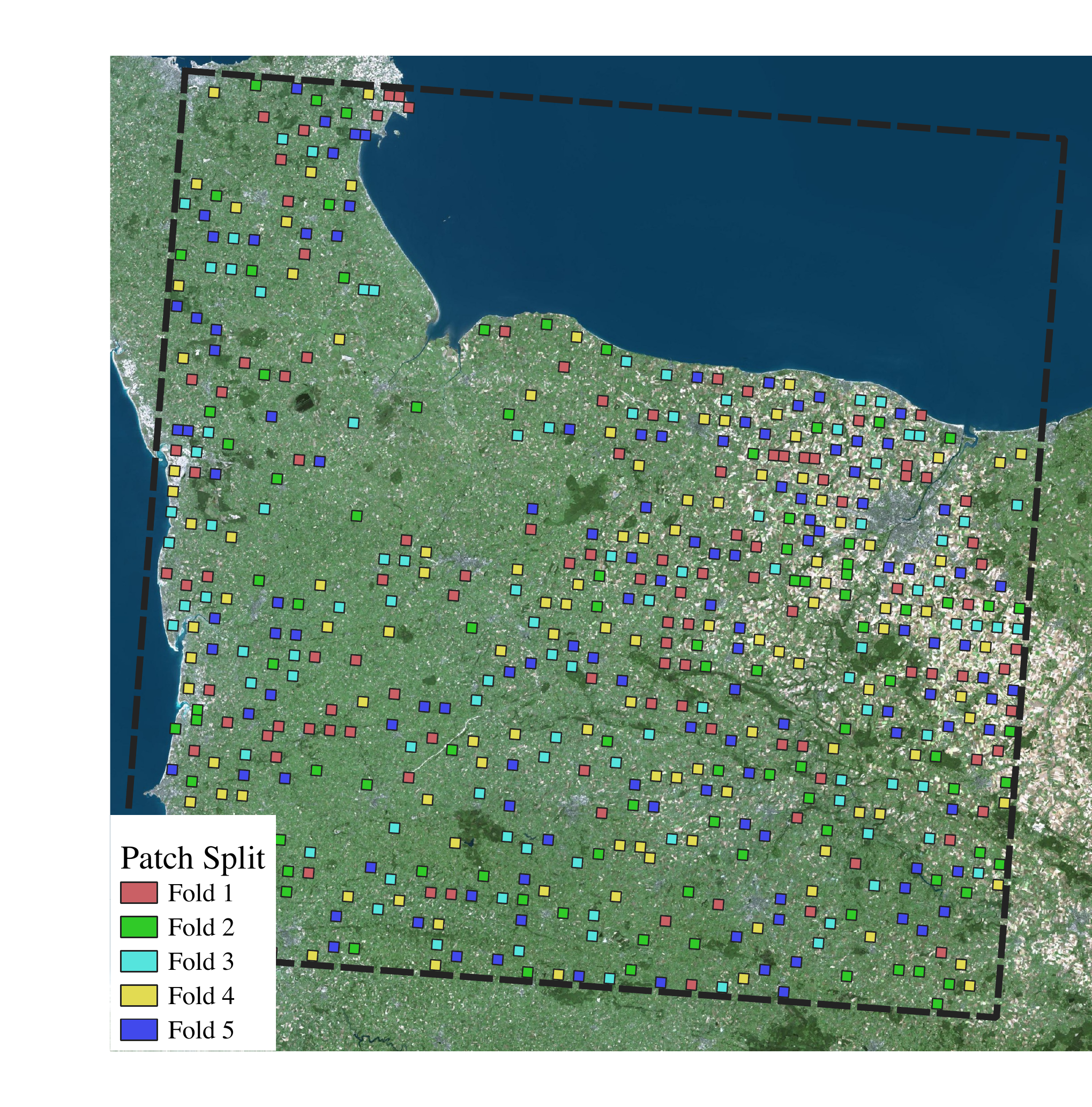}
    \caption{Selected patches.}
    \label{fig:data:zoom1}
    \end{subfigure}
    \hfill
    \begin{subfigure}{0.485\linewidth}
    \centering
    \includegraphics[width=\textwidth, trim=0cm 0cm 6.4cm 0cm, clip]{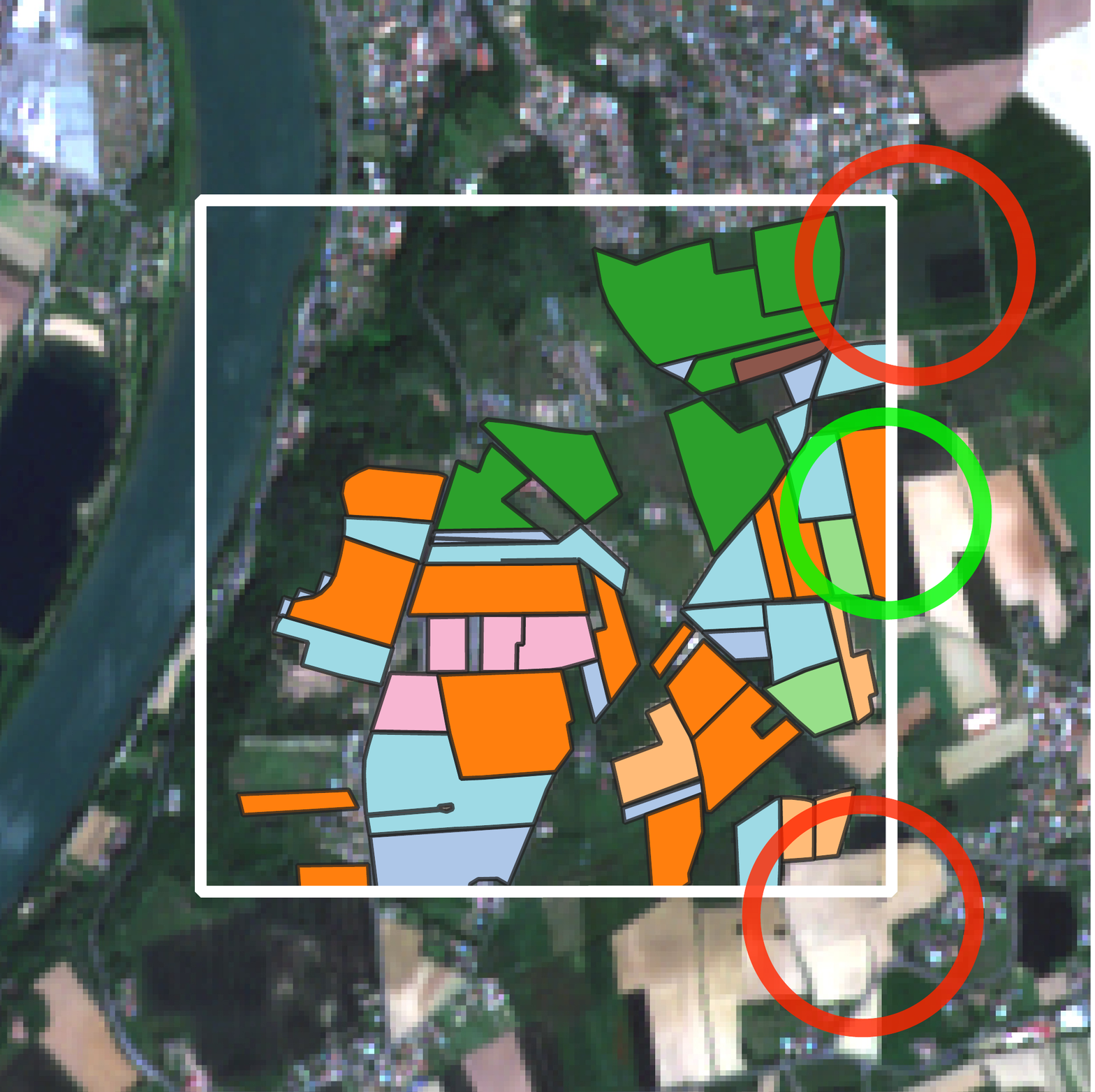}
    \caption{Single patch.}
    \label{fig:data:zoom3}
    \end{subfigure}
\caption{\textbf{Data Location.}~Spatial distribution of the four Sentinel tiles used in PASTIS \ref{fig:data:global}, and of the selected patches of tile T30UXV \ref{fig:data:zoom1}. We show an example of patch in \ref{fig:data:zoom3}, and highlight with red circles examples of parcels that are mostly outside of the patch's extent and thus annotated with the void label. The green circle \protect\tikz \protect\node[circle, thick, draw = green!90!black, fill = none, scale = 0.7] {}; highlight a parcel partially cut off by the patch borders, but with sufficient overlap to be kept as a valid parcel.
}
\end{figure}

\paragraph{Overview.} The PASTIS dataset is composed of $2433$ square $128\times128$ patches with $10$ spectral bands and at $10$m resolution, obtained from the open-access Sentinel-2 platform. \footnote{\url{https://scihub.copernicus.eu}} 
For each patch, we stack all available acquisitions between September $2018$ and November $2019$, forming our four dimensional multi-spectral SITS: $T \times C \times H \times W$. 
The publicly available French Land Parcel Identification System (FLPIS) allows us to retrieve the extent and content of all parcels within the tiles, as reported by the farmers.
Each patch pixel is annotated with a semantic label corresponding to either the parcels' crop type or the background class. The pixels of each unique parcel in the patch receive a corresponding instance label.

\paragraph{Dataset Extent.} 
The SITS of PASTIS are taken from 4 different Sentinel-2 tiles in different regions of the French metropolitan territory as depicted in Figure \ref{fig:data:global}. These regions cover a wide variety of climates and culture distributions.
Sentinel tiles span $100\times100$km and have a spatial resolution of $10$ meter per pixel. Each pixel is characterized by $13$ spectral bands. We select all bands except the atmospheric bands B01, B09, and B10.
%Sentinel-2 images are distributed under the form of $100\times100$km tiles. We choose 4 such tiles in different regions of the French metropolitan territory as depicted on Figure \ref{fig:data:global}. 
Each of these tiles is subdivided in square patches of size $1.28\times1.28$km ($128\times128$ pixels at $10$m/pixel), for a total of around $24,000$ patches.
We then select $2,433$ patches ($~10\%$ of all available patches, see \figref{fig:data:zoom1}), favoring patches with rare crop types in order to decrease the otherwise extreme class imbalance of the dataset.

%The selection process is biased to favor patches containing rare classes and thus reduce the imbalance of the dataset. We show the selected patches for one of the four tiles in Figure \ref{fig:data:zoom1}. 

\paragraph{Nomenclature} 
The FLPIS uses a $73$ class breakdown for crop types. 
%The crop types provided with the French Land Parcel Identification System are grouped in $73$ different classes.
We select classes with at least $400$ parcels and with samples in at least $2$ of the $4$ Sentinel-2 tiles. This leads us to adopt a $18$ classes nomenclature, presented in Figure \ref{fig:nomenc}.
Parcels belonging to classes not in our $18$-classes nomenclature are annotated with the \emph{void} label, see below.

\paragraph{Patch Boundaries.}
The FLPIS allows us to retrieve the pixel-precise borders of each parcel. We also compute bounding boxes for each parcel.
The parcels' extents are cropped along the extent of their $128 \times 128$ patch, and the bounding boxes are modified accordingly.
Parcels whose surface is more than $50$\% outside of the patch are annotated with the \emph{void} label, see \figref{fig:data:zoom3}.
%meaning that it will be ignored in the panoptic loss and metric,Note that the void label is different from the background label, as we do not want to penalize predictions overlapping the partial parcel. }
%In cases where an agricultural parcel is cropped by the patchs' boundary (see Figure \ref{fig:data:zoom3}), it is kept as a valid target annotation only if the cropped part contains more than half of the complete parcel surface. Conversely, parcels that are mostly outside of the patch are annotated with a void label (different than background), that will be ignored in both losses and metrics.

\paragraph{Void and Background Labels.} Pixels which are not within the extent of any declared parcel are annotated with the background ``stuff" label, corresponding to all non-agricultural land uses. For the semantic segmentation task, this label becomes the $20$-th class to predict. In the panoptic setting, this label is associated with pixels not within the extent of any predicted parcel. 
We do not compute the panoptic metrics for the background class, since our focus is on retrieving the parcels' extent rather than an extensive land-cover prediction. In other words, the reported panoptic metrics are the ``things" metrics, which already penalize parcels predicted for background pixels by counting them as false positives.

The void class is reserved for \emph{out-of-scope} parcels, either because their crop type is not in our nomenclature or because their overlap with the selected square patch is too small. We remove these parcels from all semantic or panoptic metrics and losses. Predicted parcels which overlap with an IoU superior to $0.5$ with a void parcel are not counted as false positive or true positive, but are simply ignored by the metric, as recommended in \cite{kirillov2019panoptic}.

\begin{figure}
    \centering
    \includegraphics[width=.85\linewidth, trim=0cm 5cm 38cm 0cm, clip]{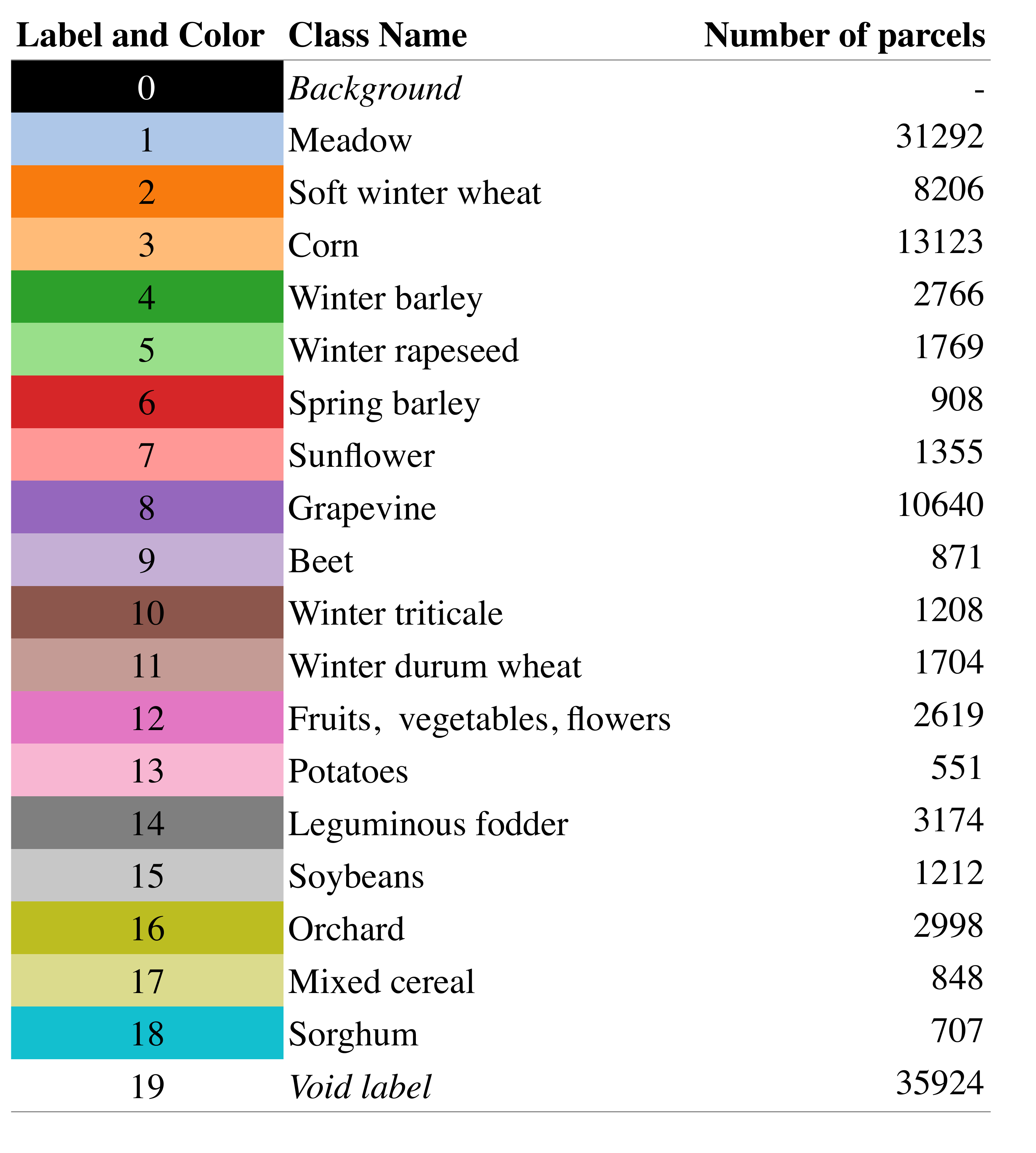}
    \caption{Color code of our class nomenclature, and the number of parcel per class.}
    \label{fig:nomenc}
\end{figure}

%We select those classes that have a support of at least $400$ parcels and are present on at least $2$ of the $4$ Sentinel-2 tiles. This restrains our nomenclature to $19$ classes that are presented in Figure \ref{fig:nomenc}.

\begin{figure}[ht!]
    \centering
    \includegraphics[width=\linewidth, trim=0cm 0cm 0cm 0cm, clip]{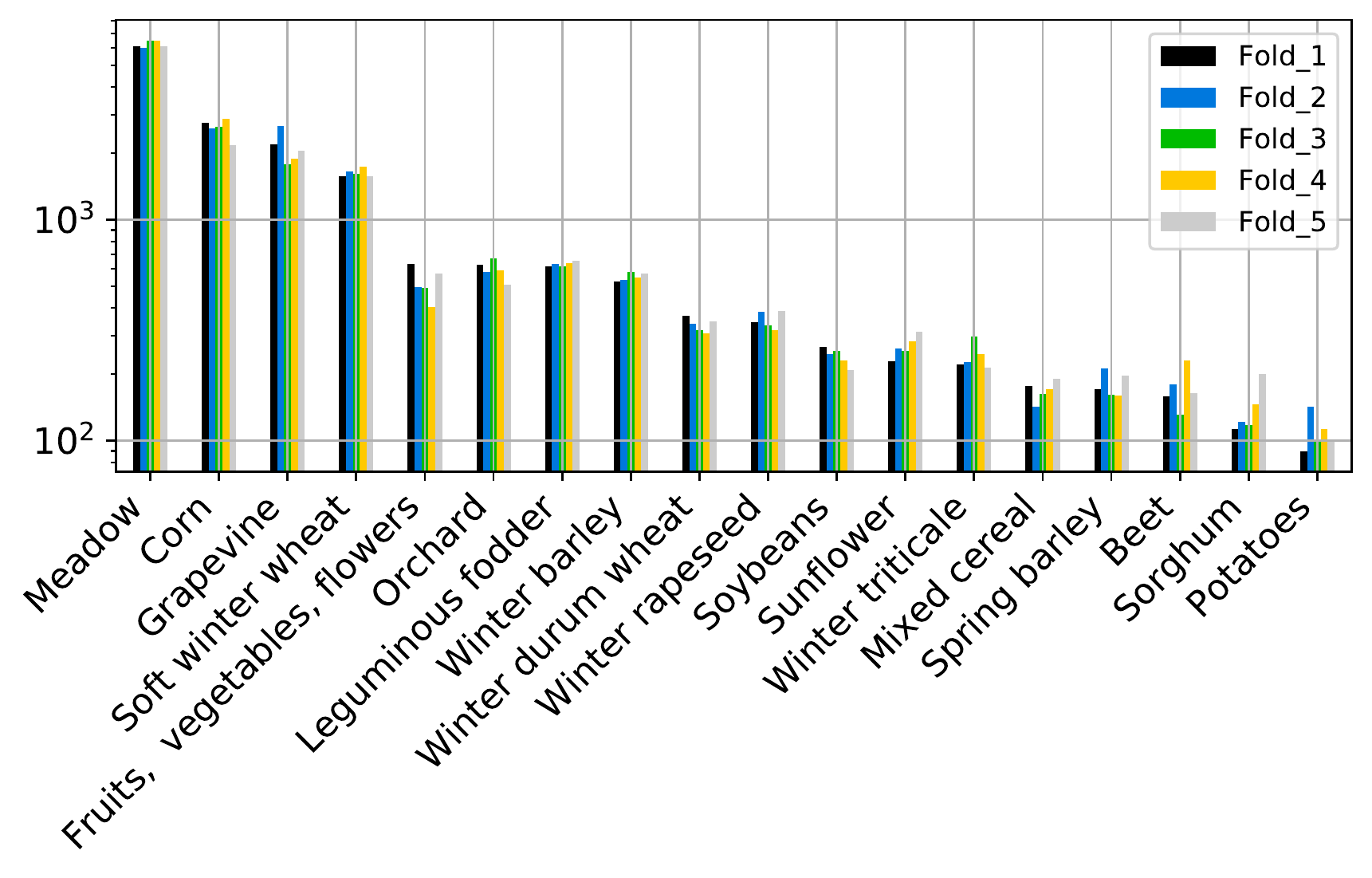}
    \caption{Class distribution for the five folds (in log-scale).}
    \label{fig:class_counts}
\end{figure}

\paragraph{Cross-Validation.} The $2,433$ selected patches are randomly subdivided into $5$ splits, allowing us to perform cross-validation. The official $5$-fold cross-validation scheme used for benchmarking is given in Table \ref{tab:cv}.
In order to avoid heterogeneous folds, 
each fold is constituted of patches taken from all four Sentinel tiles.
We also chose folds with comparable class distributions, as measured by their pairwise Kullback-Leiber divergence. We show the resulting class distribution for each fold in Figure \ref{fig:class_counts}. 
Finally, we prevent adjacent patches from being in different folds to avoid data contamination.
Geo-referencing metadata of the patches and parcels is included in PASTIS, allowing for the constitution of geographically consistent folds to evaluate spatial generalization. However, this is out of the scope of this paper.
%Neighboring patches are assigned to the same fold, ensuring a $1,28$km buffer between folds.
%Out of $100$ different random splits, we kept the one with minimal Kullback-Leiber divergence between the class distributions of the different folds. We show the resulting class distribution for each fold on Figure \ref{fig:class_counts}. 
\begin{table}[h]
    \centering
    \begin{tabular}{c|ccc}
       Fold  & Train & Val & Test  \\\midrule
        I &1-2-3 & 4 & 5\\
        II &2-3-4 & 5 & 1\\
        III &3-4-5 & 1 & 2\\
        IV &4-5-1 & 2 & 3\\
        V &5-1-2 & 3 & 4\\    \end{tabular}
    \caption{Official $5$-fold cross validation scheme. Each line gives the repartition of the splits into train, validation and test set for each fold. }
    \label{tab:cv}
\end{table}

\paragraph{Temporal Sampling.} The temporal sampling of the sequences in PASTIS is irregular: depending on their location, patches are observed a different number of times and at different intervals.
%depending on their location different patches are not observed at the same dates, and not for the same number of times. 
This is a result of both the orbit schedule of Sentinel-2 and the policy of Sentinel data providers not to process tile observations identified as covered by clouds for more than $90\%$ of the tile's surface.
As this corresponds to the \emph{real world} setting, we decided to leave the SITS as is, and thus to encourage methods that can favourably address this technical challenge. As a result, the proposed SITS are constituted of $33$ to $61$ acquisitions.
In order to assess how our model handles lower sampling frequencies, we limited the number of available acquisitions at inference time\footnote{This can be interpreted as the test set having an increased cloud cover.}, and observed a  drop of performance of $-0.7$, $-2.0$, $-5.5$, and $-14.6$ points of mIoU with $32$, $24$, $16$, and $8$ available dates, respectively.

\paragraph{Clouds Cover. } Even after the automatic filtering of predominantly cloudy acquisitions, some patches are still partially or completely obstructed by cloud cover. We opt to not apply further pre-processing or cloud detection, and produce the raw data in PASTIS. Our reasoning is that an adequate algorithm should be able to learn to deal with such acquisitions. Indeed, robustness to cloud-cover has been experimentally demonstrated for deep learning methods by Ru{\ss}wurm and K\"orner \cite{russwurm2018convolutional, russwurm2020self}.
%.................................................................
\section{Implementation Details}
%.................................................................
In this section, we detail the exact configuration of our method as well as the competing algorithms evaluated.

\paragraph{Training Details.} 
Across our experiments, we use Adam \cite{kingma2014adam} optimizer with default parameters and a batch size of $4$ sequences. The semantic segmentation experiments use a fixed learning rate of $0.001$ for $100$ epochs. For the panoptic segmentation experiments, we start with a higher learning rate of $0.01$ for $50$ epochs, and decrease it to $0.001$ for the last $50$ epochs.

\paragraph{U-TAE.} 
In \tabref{tab:unet_conf}, we report the width of the feature maps outputted by each level of the U-TAE's encoder and decoder.
In both networks, we use the the same convolutional block shown in \figref{fig:convlblock} and constituted of one $3\times3$ convolution from the input to the output's width, and one residual $3\times3$ convolution. In the encoding branch, we use Group Normalisation with $4$ groups and Batch Normalisation in the decoding branch .

%We give the number of channels used for each level of the U-TAE in table \ref{tab:unet_conf}. We use the same convolutional block shown in Figure \ref{fig:convlblock} across the encoding and decoding branches. In the encoding branch, we use Group Normalisation with $4$ groups and in the decoding branch Batch Normalisation.
For the temporal encoding, we chose a L-TAE with $16$ heads, and a key-query space of dimension $d_k=4$. We use Group Normalisation with $16$ groups at the input and output of the L-TAE, meaning that that the inputs of each head are layer-normalized.
%The L-TAE network used for temporal encoding is set with $16$ heads, and a key-query space of dimension $d_k=4$. We use Group Normalisation at the input and output of the L-TAE with one group per head.

\begin{figure}[h!]
    \centering
    \includegraphics[width=.7\linewidth, trim=0cm 30cm 42cm 0cm, clip]{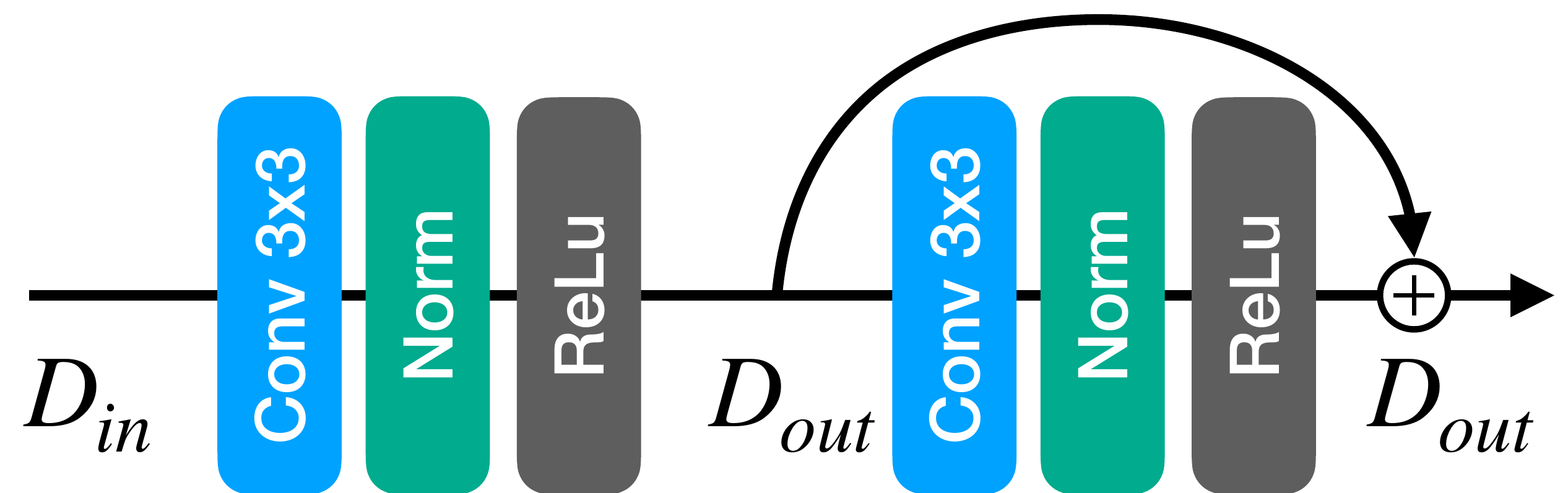}
    \caption{Structure of the convolutional block used in the spatial encoder-decoder network. This block maps a feature map with $D_{in}$ channels to a feature map with $D_{out}$ channels. }
    \label{fig:convlblock}
\end{figure}

\begin{table}[]
    \centering
    \caption{Width of the feature maps outputted at each level of the encoding and decoding branches of the spatial module.}
    \begin{tabular}{ccccc}
      \multicolumn{2}{c}{Encoder}&\phantom{a}&\multicolumn{2}{c}{Decoder}\\\cmidrule{1-2}\cmidrule{4-5}
       $e_1$ & 64 && $d_1$ & 32 \\
        $e_2$ & 64 && $d_2$ & 32 \\
       $e_3$ & 64 && $d_3$ & 64 \\
       $e_4$ & 128 && $d_4$ & 128 \\
    \end{tabular}
    \label{tab:unet_conf}
\end{table}

\paragraph{Recurrent Models.} 
We use the same U-Net architecture for our models and \emph{U-BiConvLSTM} and \emph{U-ConvLSTM}, but simply replace the L-TAE by a ConvLSTM or BiConvLSTM respectively. The hidden state's size of the biConvLSTM is chosen as $32$ in both directions, and $64$ for the convLSTM.
For the recurrent-convolutional methods \emph{ConvLSTM} and \emph{ConvGRU} not using a U-Net, we set hidden sizes of $160$ and $188$ respectively. 
%U-BiConvLSTM and U-ConvLSTM use the same U-Net structure as our U-TAE. For U-BiConvLSTM we replace the L-TAE with a bi-directional ConvLSTM with a hidden size of $32$ in each direction. For U-ConvLSTM, we replace it with a ConvLSTM with a hidden size of $64$. The ConvLSTM and ConvGRU models are set with a hidden size of  $160$ and $188$ respectively. 

\paragraph{3D-Unet.} 
For this network, we use the official PyTorch implementation of Rustowicz \etal \cite{m2019semantic}. This network is constituted of 
%For the 3d-Unet  model we use the PyTorch implementation of \cite{m2019semantic}. This implementation uses 
five successive 3D-convolution blocks with spatial down-sampling after the $2$nd and $4$th blocks. Each convolutional block doubles the number of channels of the processed feature maps, and the innermost feature maps have a channel dimension set to $128$. Leaky ReLu and 3D Batch Normalisation are used across the convolutional blocks of this architecture. The sequence of feature maps is averaged along the temporal dimension to produce the final embedding of the input image sequence. In their implementation, the authors used a linear layer to collapse the temporal dimension, yet this was not a valid option for PASTIS as the sequences have highly variable lengths and
the sequence indices do not correspond to the same acquisition date from one sequence to another.
%observation dates do not have a fixed position.
 
\paragraph{FPN-ConvLSTM.} 
For this architecture, the input sequence of images is first mapped to feature maps of channel dimension $64$ with two consecutive $3\times3$ convolution layers, followed by Group Normalization and ReLu. A $5$-level feature pyramid is then constructed for each date of the sequence
by applying to the feature maps $4$ different $3\times3$ convolution of respective dilation rates $1$, $2$, $4$ and $8$, and computing the spatial average of the feature map.
%\VIVIEN{and a constant feature map equal to the global average pooling of the input maps}. 
These $5$ maps are concatenated along the channel dimension, and processed by a ConvLSTM with a hidden state size of $88$. We found it beneficial to use a supplementary convolution before the ConvLSTM to reduce the number of channels of the feature pyramid by a factor $2$.
%the output of global average pooling, and of $4$ different $3\times3$ convolutional layers with respective dilation rates $1$,$2$,$4$, and $8$ are concatenated along the channel dimension. The sequence of feature pyramids is then fed to a ConvLSTM with hidden state size of $88$. We found it beneficial to use an intermediary convolution before the ConvLSTM to reduce the number of channels of the feature pyramids by a factor $2$.

\paragraph{PaPs module.} In the PaPs module, the saliency and heatmap predictions are obtained with two separate convolutional blocks operating on the high resolution feature map $d_1$ with $32$ channels. These blocks are composed of two convolutional layers
of width $32$ and $1$  respectively. We use Batch Normalisation and ReLu after the first convolution, and a sigmoid after the second.
%, one with $32$ kernel and second with only $1$ output dimension. Batch Normalisation and ReLu is used between the two convolutions, while only sigmoid is applied to the output of the second convolution.

The $256$-dimensional multi-scale feature vector ($128+64+32+32)$ is mapped to the shape, class and size predictions by three different MLPs described in Table \ref{tab:mlp_config}. The inner layers use Batch Normalisation and ReLu activation.

{The residual CNN used for shape refinement is composed of three convolutional layers : $1 \mapsto 16 \mapsto 16 \mapsto 1$, with ReLu activation and instance normalisation on the first layer only.}
%with a MLP $:256 \mapsto 128 \mapsto S^2+N+2$. The hidden layer uses Batch Normalisation and ReLu activation.
%set with a first linear layer with $128$ output neurons, Batch Normalisation, ReLu, and a second linear layer that outputs an $S^2+K+2$ vector.
\begin{table}[]
    \caption{Configuration of the four MLPs of PaPs}

    \centering
    \begin{tabular}{l|lc}
        MLP & Layers & Final Layer\\\midrule
        Shape & $256 \mapsto 128 \mapsto S^2$ & - \\
        Size & $256 \mapsto 128 \mapsto 2$ & Softplus\\
        Class & $256 \mapsto 128 \mapsto 64 \mapsto K$ & Softmax\\
    \end{tabular}
    \label{tab:mlp_config}
\end{table}

\begin{figure}[h!]
    \centering
    \includegraphics[width=\linewidth, trim=0cm .9cm 0cm 0cm, clip]{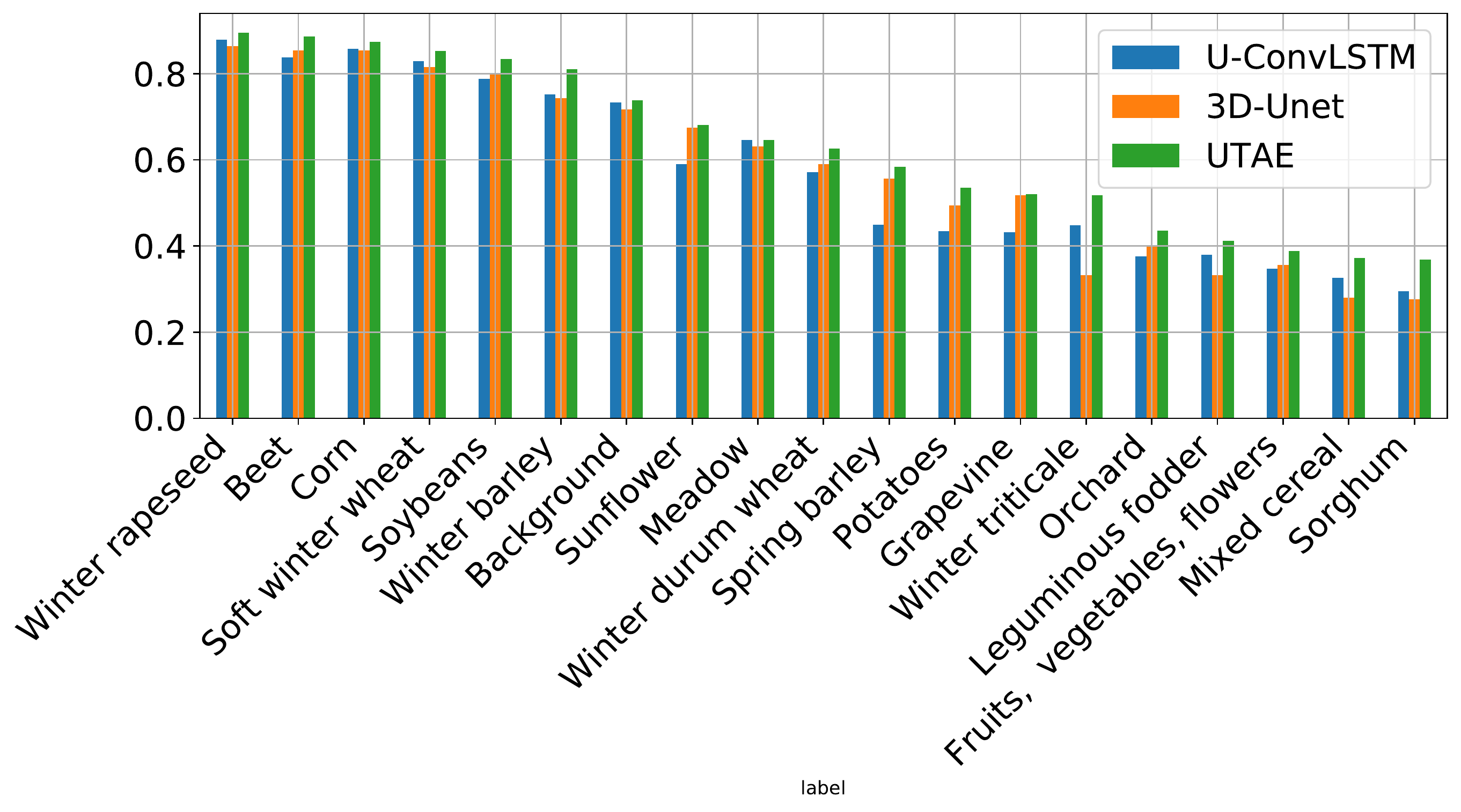}
    \caption{Per class IoU of the three best performing semantic segmentation models. Our U-TAE outperforms the other two approaches on every classes, and brings noticeable improvement on hard classes such as Mixed cereal and Sorghum. }
    \label{fig:per_class}
\end{figure}

\begin{figure}[h!]
    \centering
    \includegraphics[width=\linewidth]{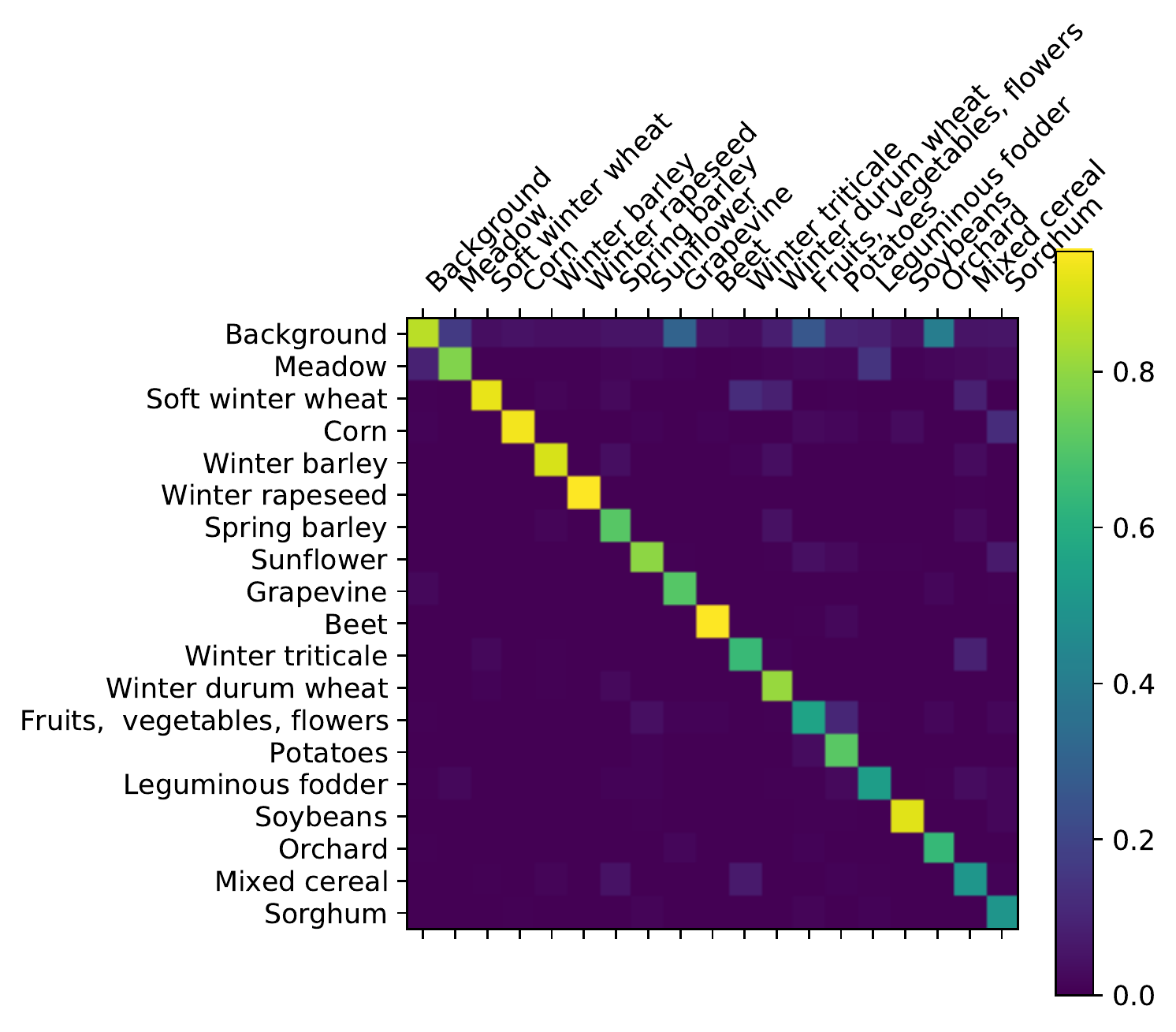}
    \caption{Confusion matrix of U-TAE for semantic segmentation on PASTIS. The color of each pixel at line $i$ and column $j$ corresponds to the proportion of samples of the class $i$  that were attributed to the class $j$.}
    \label{fig:confmat}
\end{figure}

\paragraph{Handling Sequences of Variable Lengths.} All models are trained on batches of sequences of variable length.
To facilitate the handling of batches by the GPU, we append all-zeroes images at the end of shorter sequences to match the length of the longer sequence in the batch.
We retain a padding mask to prevent the spatial and temporal encoding of padded values, and to exclude these padded values from temporal averages.
%We thus pad shorter sequences to match the length of the longer sequence in the batch. We use a padding mask to make sure that no trainable layer processes padded values, and also to exclude padded values when computing temporal averages. In our U-TAE we set the key-query compatibility of padded positions to $10^{-16}$. In recurrent-based architectures, we take the hidden state of the last non-padded position to represent the input sequence.

\section{Additional Results}
{In \figref{fig:per_class}, we show the class-wise performance of the three best performing semantic segmentation models, displaying an improvement of U-TAE compared to the other methods across all crop types. We also show on \figref{fig:confmat} the confusion matrix of U-TAE. Unsurprisingly, confusions seem to occur between semantically close classes such as different cereal types, or \textit{Sunflower} and \textit{Fruits, Vegetable, Flower}.  }

In \figref{fig:qualisup}, we present qualitative results illustrating the predicted panoptic and semantic segmentations compared to the ground truth. In particular, we show some failure cases in which thin or visually fragmented parcels are not recovered correctly.

In \figref{fig:sem}, we illustrate the results of the semantic segmentation for our method and three other competing approaches: \emph{3D-Unet}, \emph{U-BiConvLSTM}, and \emph{convGRU}. We show how our multi-scale temporal attention masks allow our predictions to be both pixel-precise and consistent for large parcels.

Finally, we present in \figref{fig:mono} an example of inference using a single image from the sequence. As expected for mono-temporal segmentation, the parcel classification is poor. Furthermore, we show a case of a border that is essentially invisible on a single image, but that our full model is able to detect using the entire sequence of satellite images.

\section*{Acknowledgments}
The satellite images used in PASTIS were gathered from  \href{www.theia.land.fr}{THEIA}: 
``\emph{Value-added data processed by the CNES for the Theia data cluster using Copernicus data.
The treatments use algorithms developed by Theia’s Scientific Expertise Centres.}''
The annotations of PASTIS were taken from the French \href{https://www.data.gouv.fr/en/datasets/registre-parcellaire-graphique-rpg-contours-des-parcelles-et-ilots-culturaux-et-leur-groupe-de-cultures-majoritaire/}{LPIS} produced
 by IGN, the French mapping agency.
This work was partly supported by \href{https://www.asp-public.fr}{ASP}, the French Payment Agency.

\begin{figure*}
    \begin{subfigure}{0.24\textwidth}
        \centering
        \includegraphics[width=\textwidth]{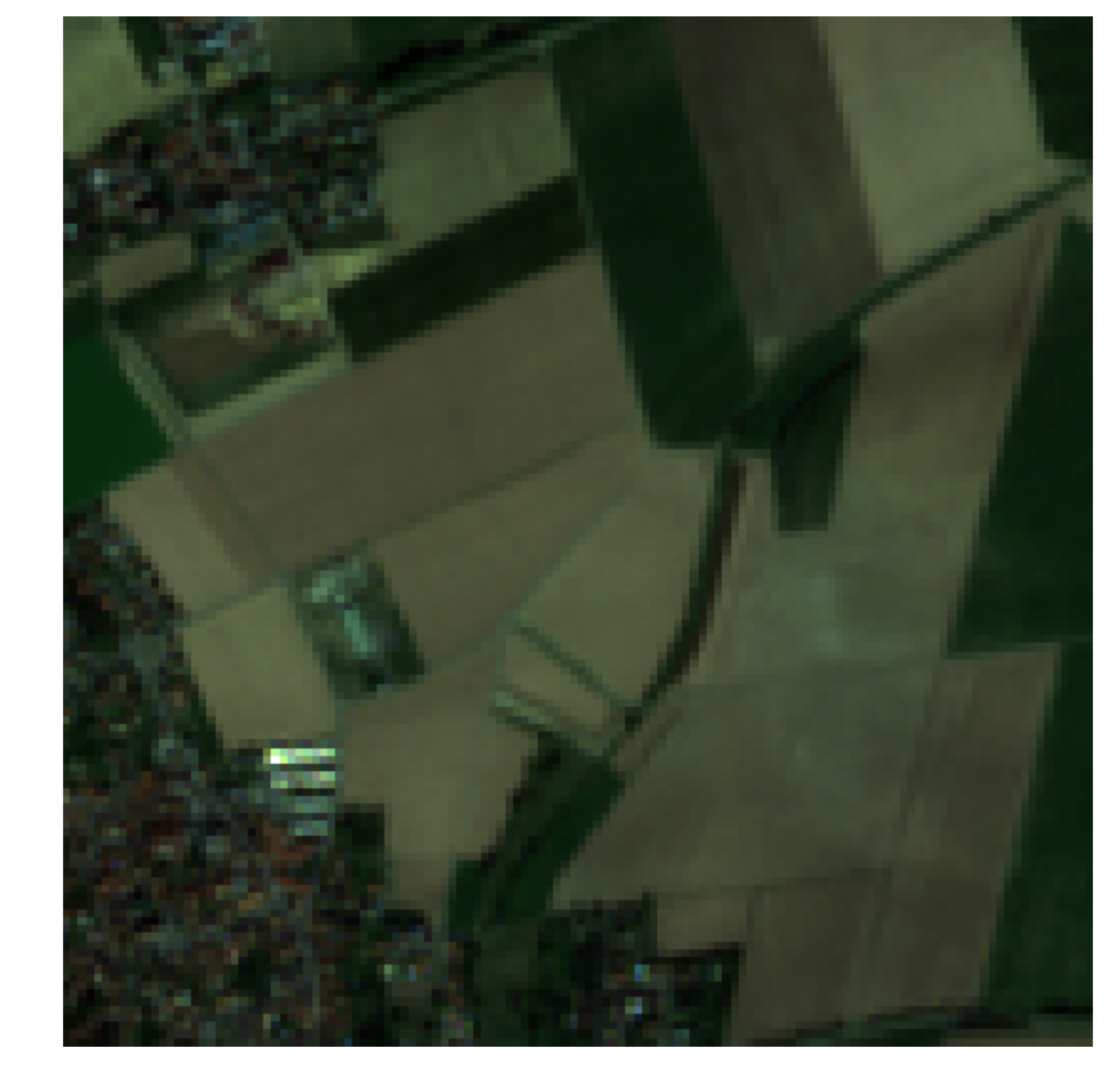}
        %\caption{Image from the sequence.}
        \label{fig:qualisup:rgb}
        \end{subfigure}
    \hfill
        \begin{subfigure}{0.24\textwidth}
        \includegraphics[width=\textwidth]{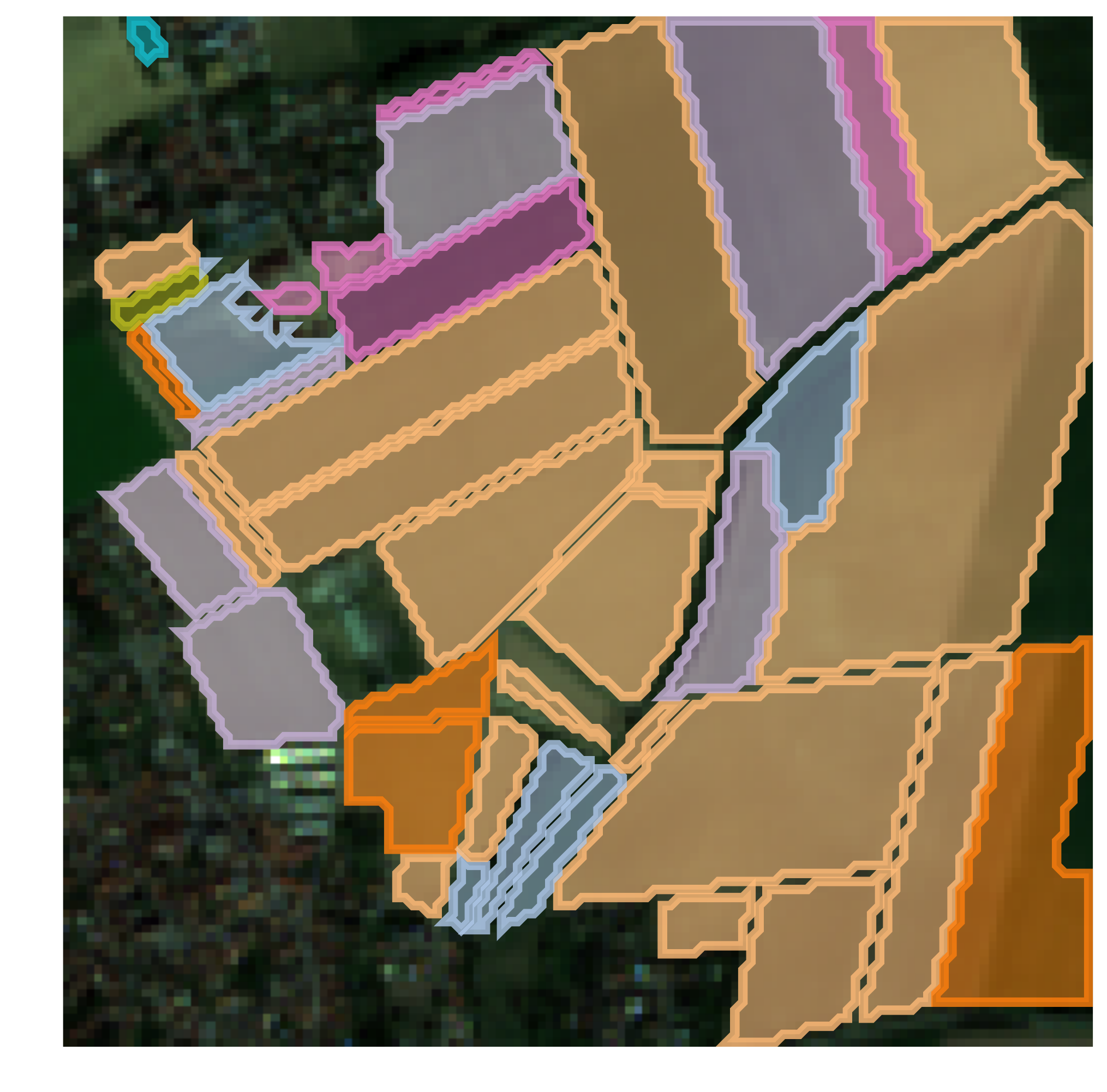}
        %\caption{Panoptic annotation.}
        \label{fig:qualisup:gt}
        \end{subfigure}
    \hfill
        \begin{subfigure}{0.24\textwidth}
        \includegraphics[width=\textwidth]{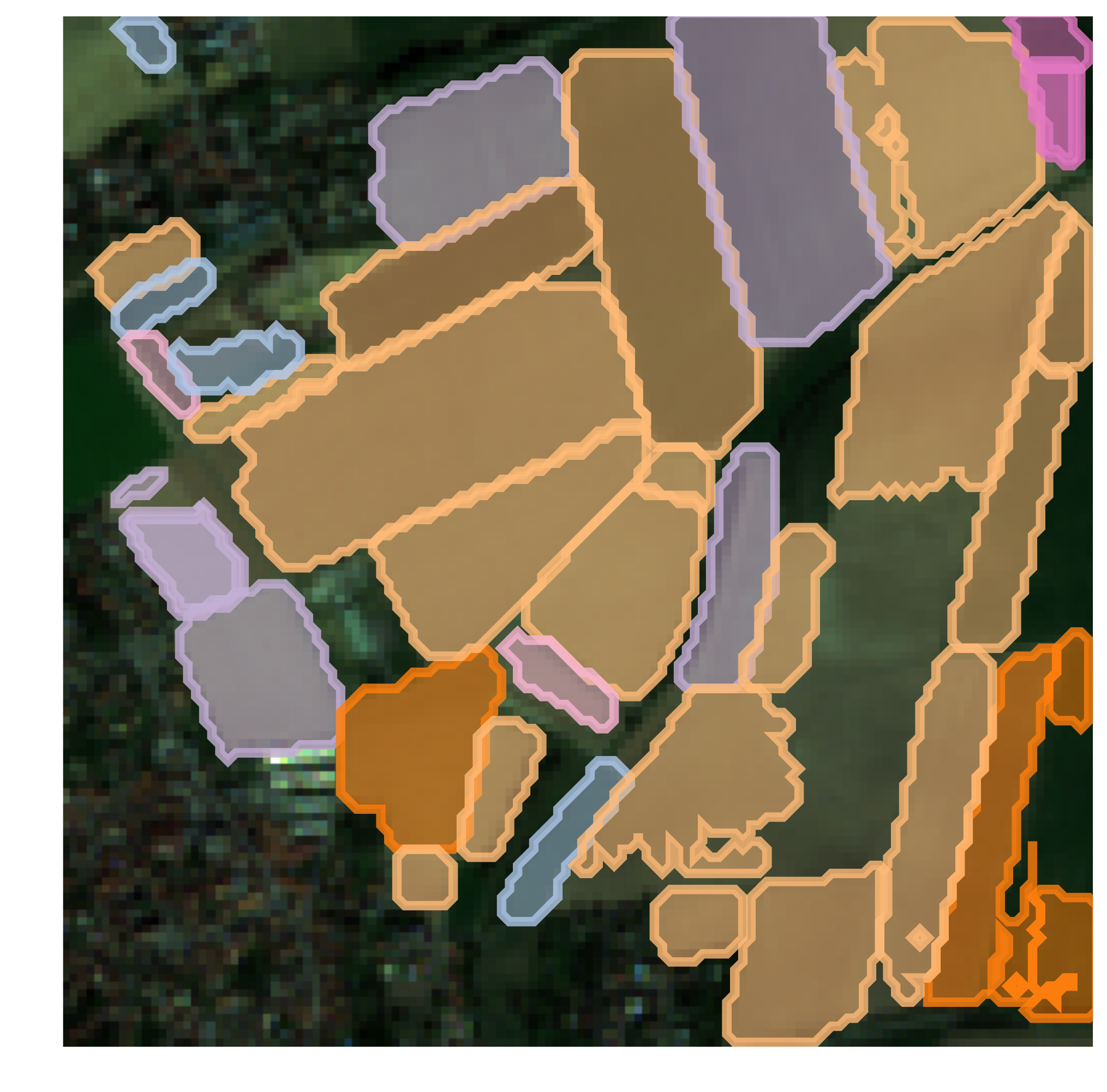}
        %\caption{Panoptic segmentation.}
        \label{fig:qualisup:pan}
        \end{subfigure}
    \hfill
        \begin{subfigure}{0.24\textwidth}
        \includegraphics[width=\textwidth]{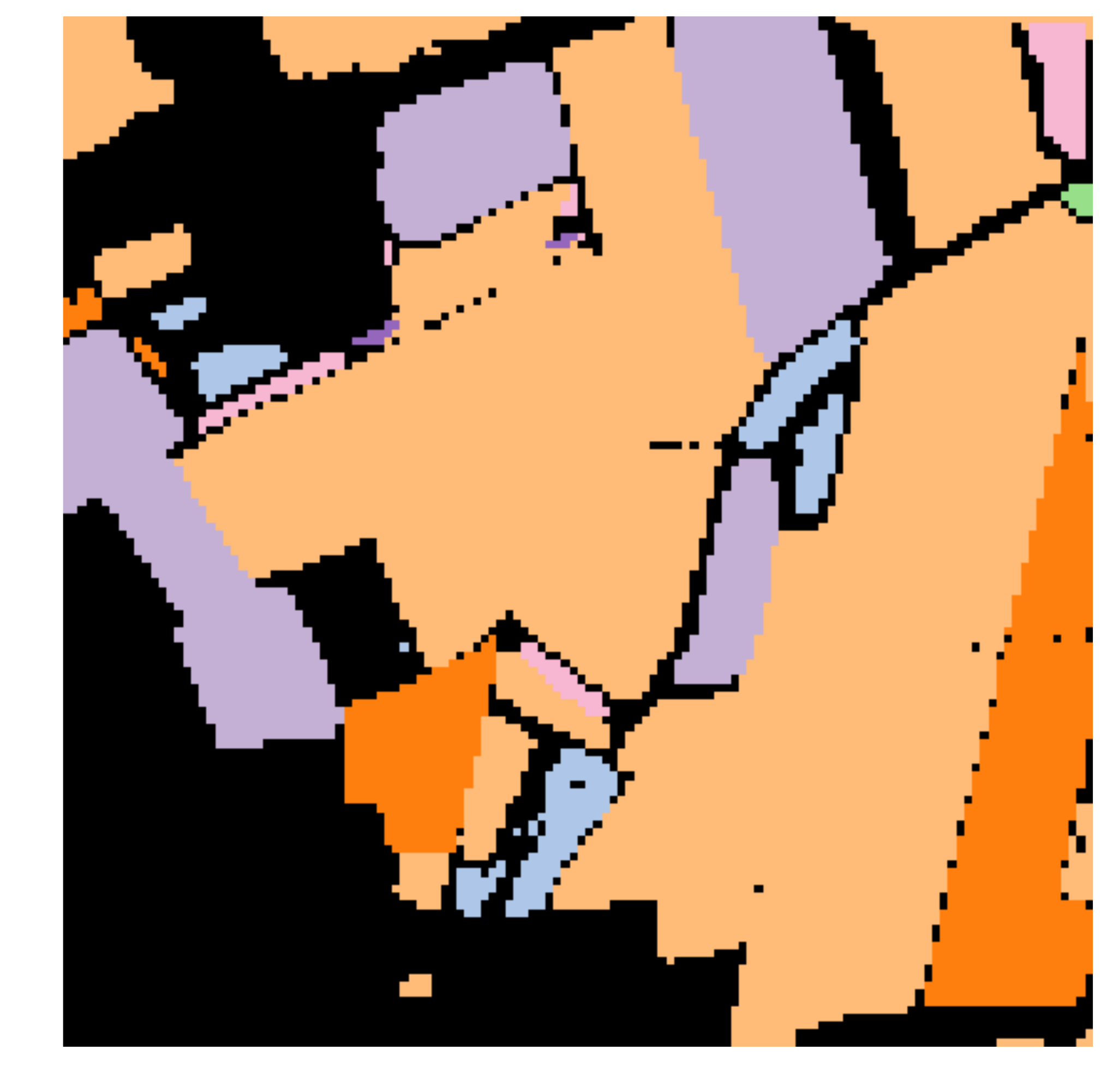}
        %\caption{Semantic segmentation.}
        \label{fig:qualisup:seg}
        \end{subfigure}
    \vfill
    \vspace{-.4cm}
        \begin{subfigure}{0.24\textwidth}
        \centering
        \includegraphics[width=\textwidth]{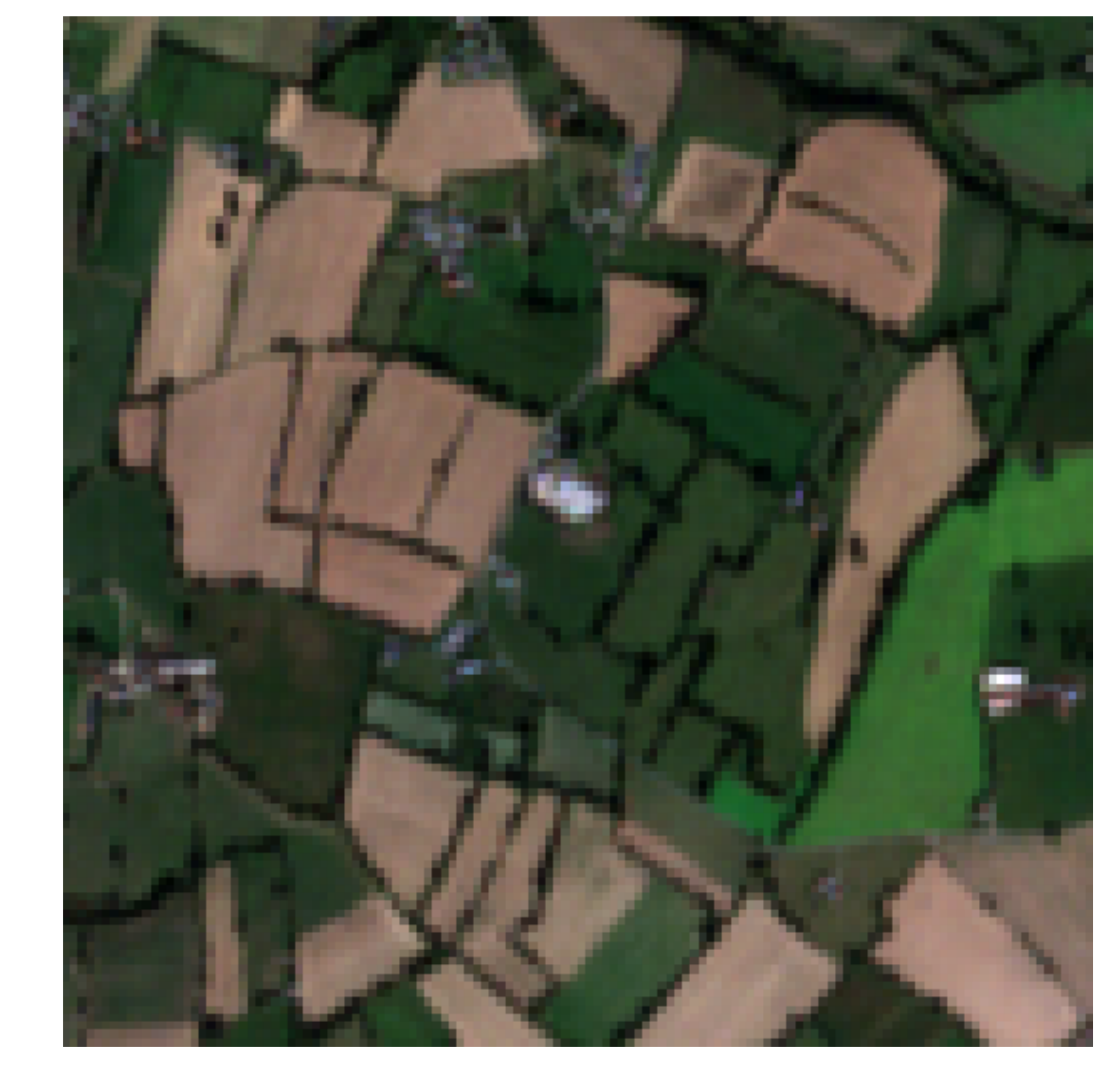}
        %\caption{Image from the sequence.}
        \label{fig:qualisup2:rgb}
        \end{subfigure}
    \hfill
        \begin{subfigure}{0.24\textwidth}
        \centering
        \begin{tikzpicture}
         \node[anchor=south west,inner sep=0] (image) at (0,0) {\includegraphics[width=\textwidth]{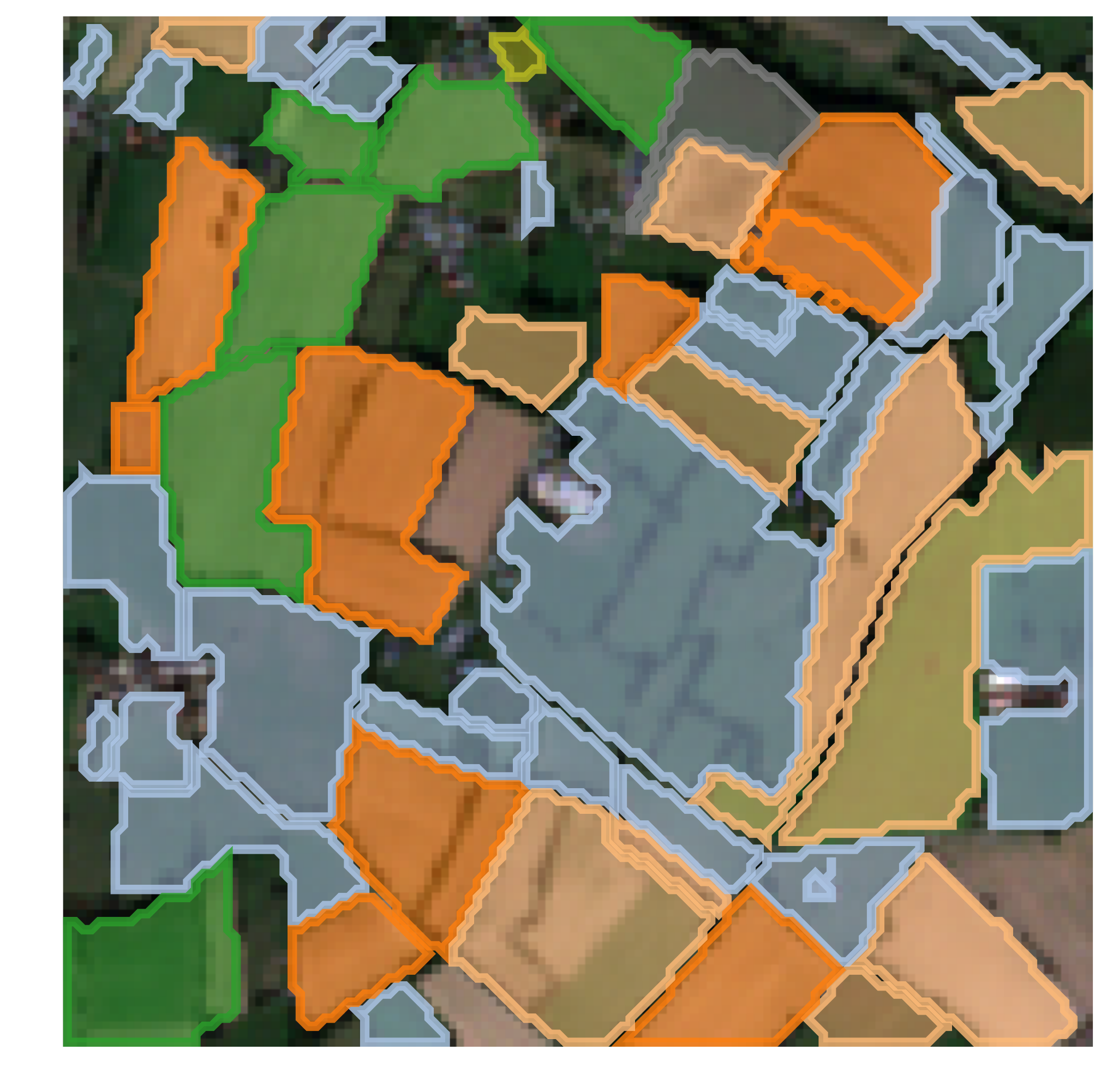}};
        \begin{scope}[x={(image.south east)},y={(image.north west)}]
        \draw[green,ultra thick] (0.6,0.45) circle (0.18);
        \end{scope}
        \end{tikzpicture}   
        %\includegraphics[width=\textwidth]{images/fig5_v2/Fig4_2_10.pdf}
        %\caption{Panoptic annotation.}
        \label{fig:qualisup2:gt}
        \end{subfigure}
    \hfill
        \begin{subfigure}{0.24\textwidth}
        \includegraphics[width=\textwidth]{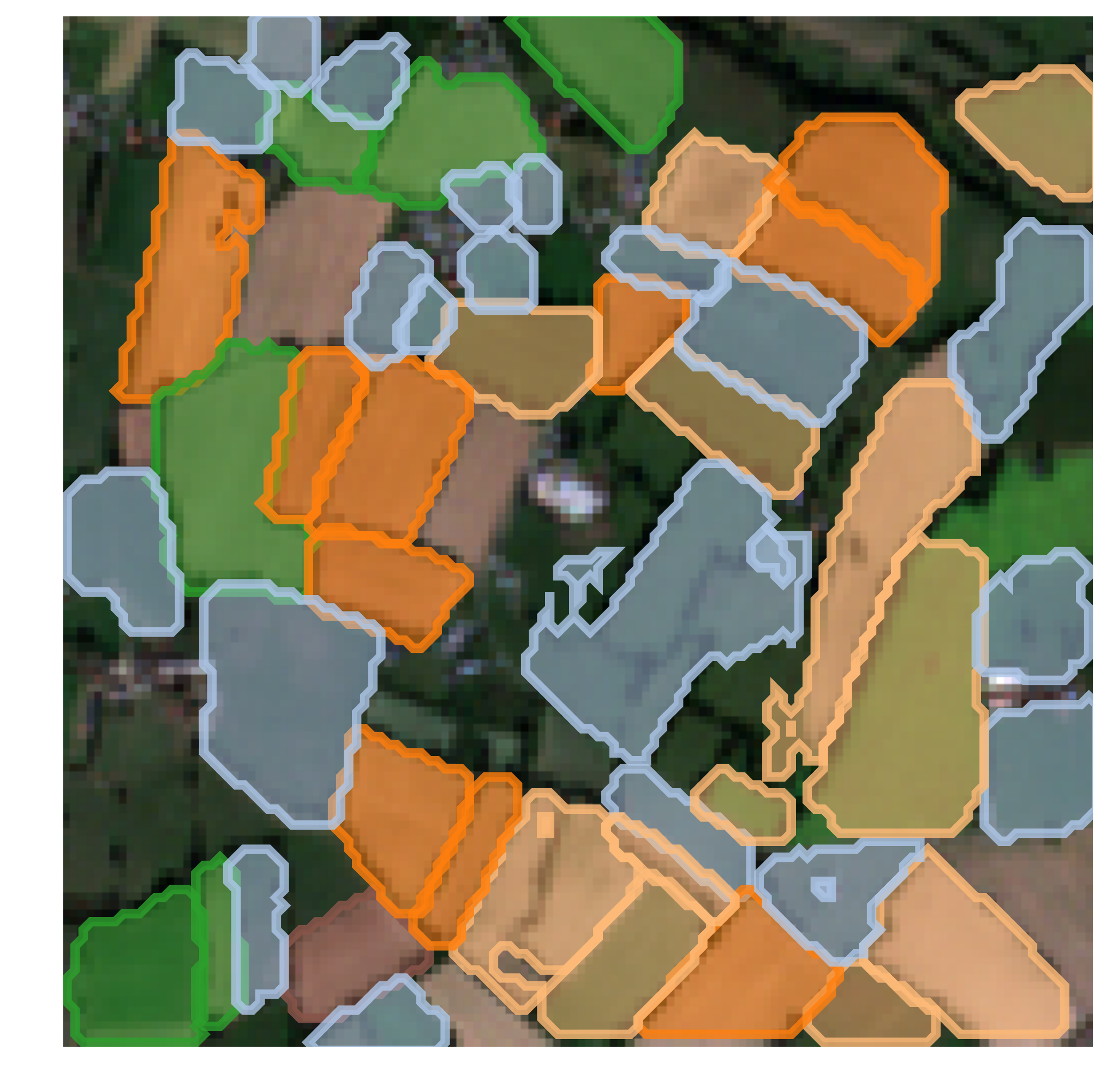}
        \label{fig:qualisup2:pan}
        \end{subfigure}
    \hfill
        \begin{subfigure}{0.24\textwidth}
        \includegraphics[width=\textwidth]{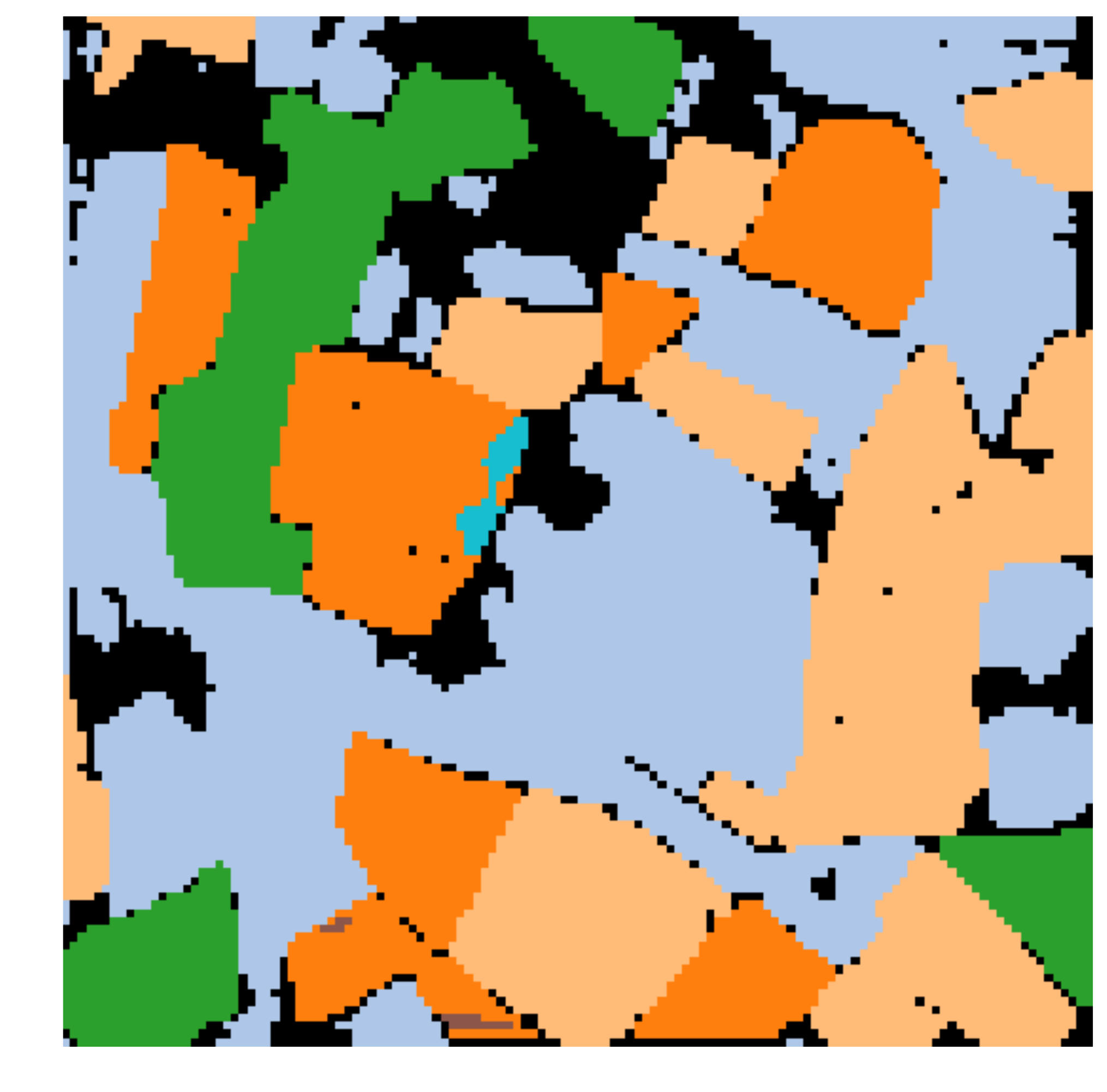}
        \label{fig:qualisup2:seg}
        \end{subfigure}
        
    \vfill
    \vspace{-.4cm}
        \begin{subfigure}{0.24\textwidth}
        \centering
        \includegraphics[width=\textwidth]{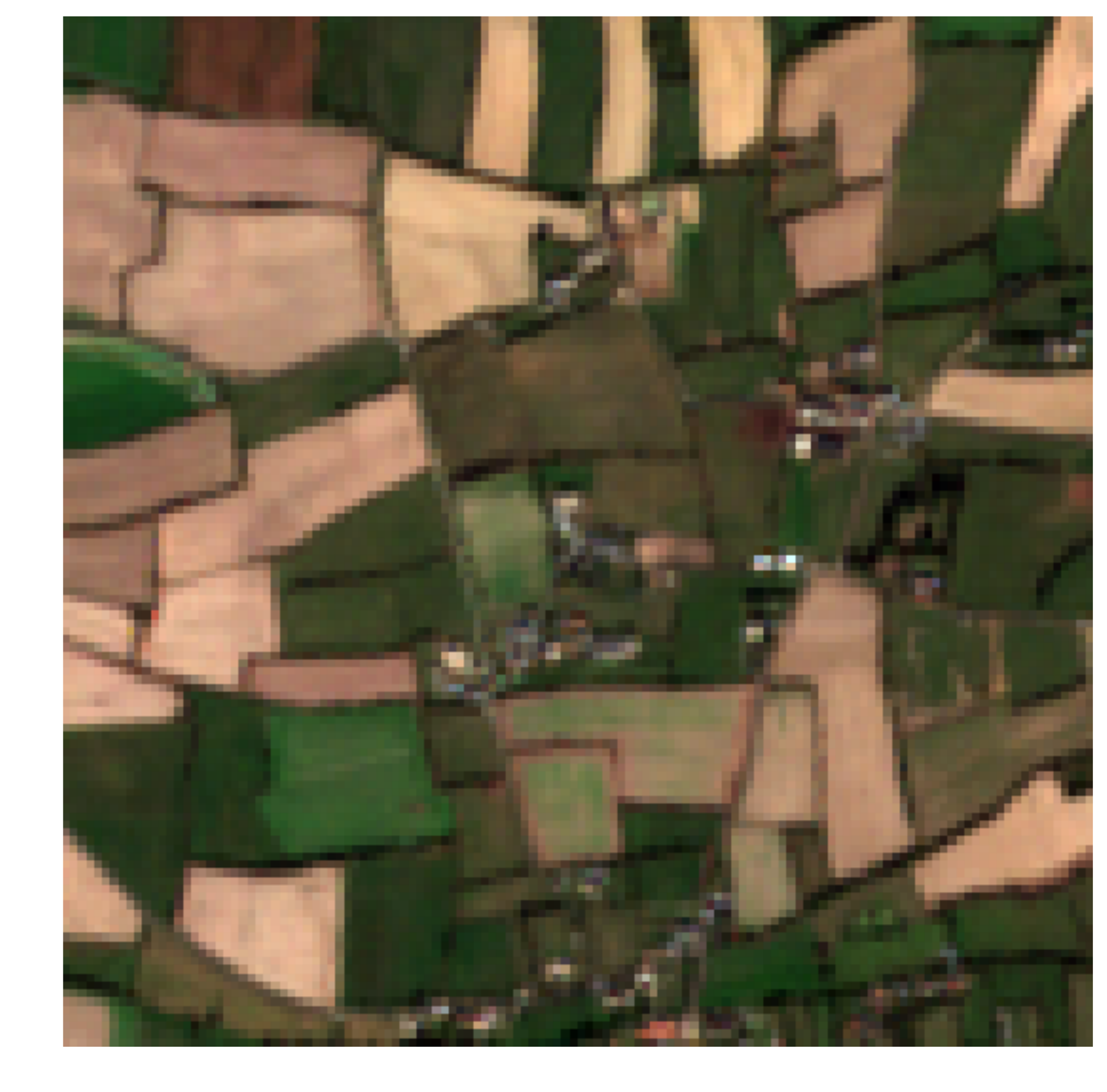}
        %\caption{Image from the sequence.}
        \label{fig:qualisup3:rgb}
        \end{subfigure}
    \hfill
        \begin{subfigure}{0.24\textwidth}
        \centering
        \begin{tikzpicture}
         \node[anchor=south west,inner sep=0] (image) at (0,0) {\includegraphics[width=\textwidth]{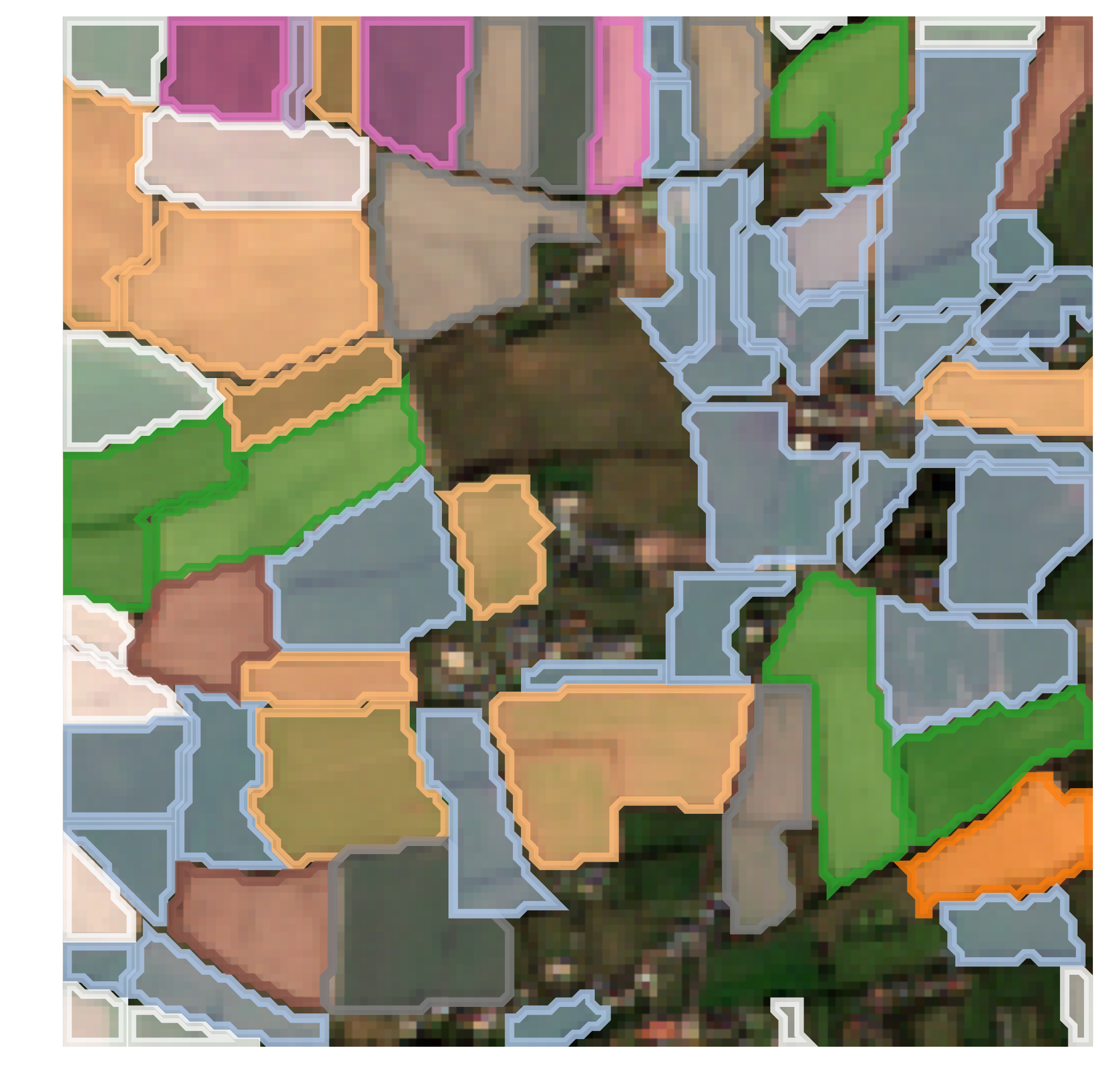}};
        \begin{scope}[x={(image.south east)},y={(image.north west)}]
        \draw[cyan ,ultra thick] (0.55,0.3) circle (0.12);
        \end{scope}
        \end{tikzpicture}
        %\includegraphics[width=\textwidth]{images/Fig10_v2/Fig4_2_1.pdf}
        %\caption{Panoptic annotation.}
        \label{fig:qualisup3:gt}
        \end{subfigure}
    \hfill
        \begin{subfigure}{0.24\textwidth}
        \includegraphics[width=\textwidth]{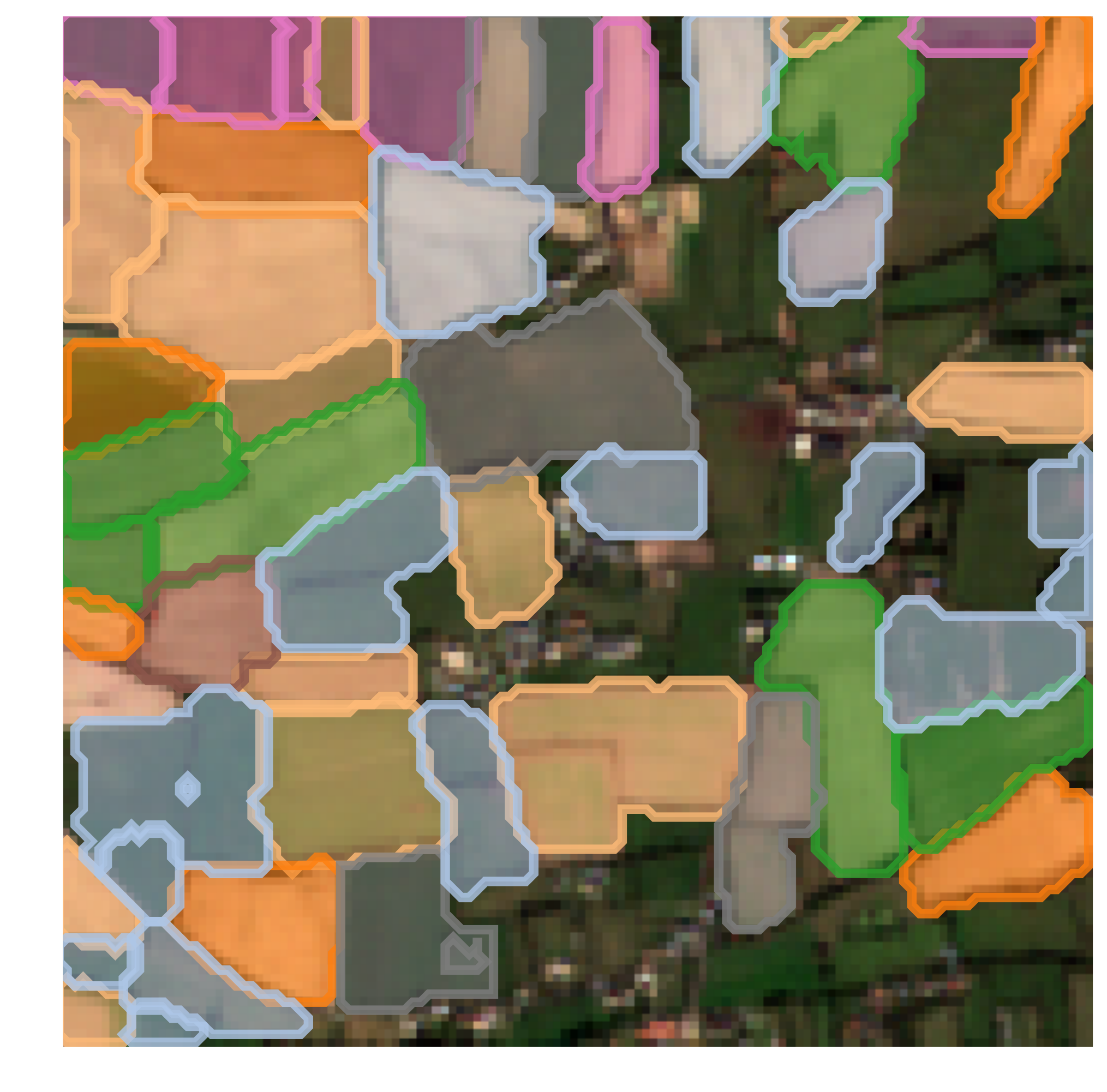}
        \label{fig:qualisup3:pan}
        \end{subfigure}
    \hfill
        \begin{subfigure}{0.24\textwidth}
        \includegraphics[width=\textwidth]{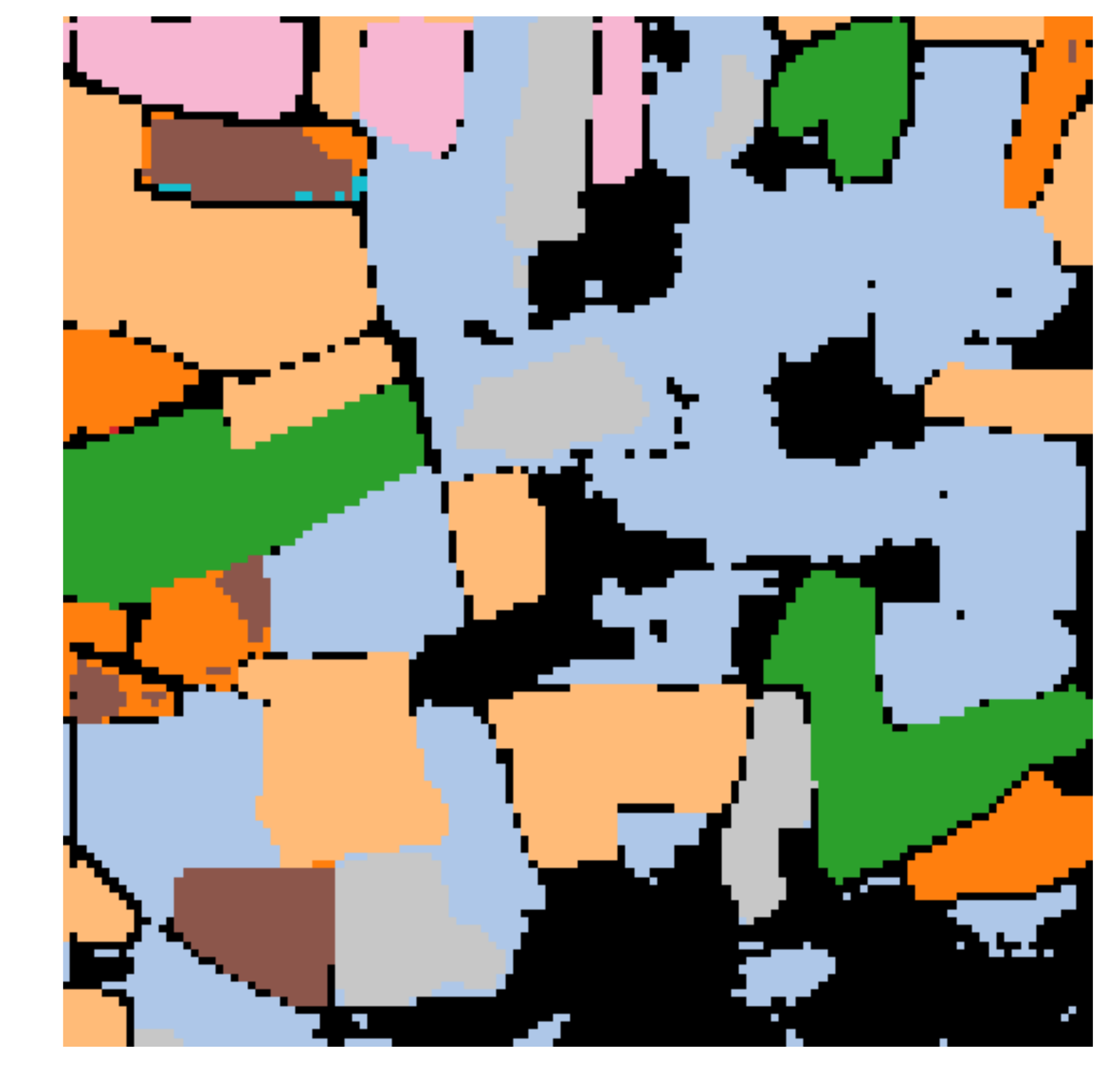}
        \label{fig:qualisup3:seg}
        \end{subfigure}
    
    \vfill 
    \vspace{-.4cm}

        \begin{subfigure}{0.24\textwidth}
        \centering
        \includegraphics[width=\textwidth]{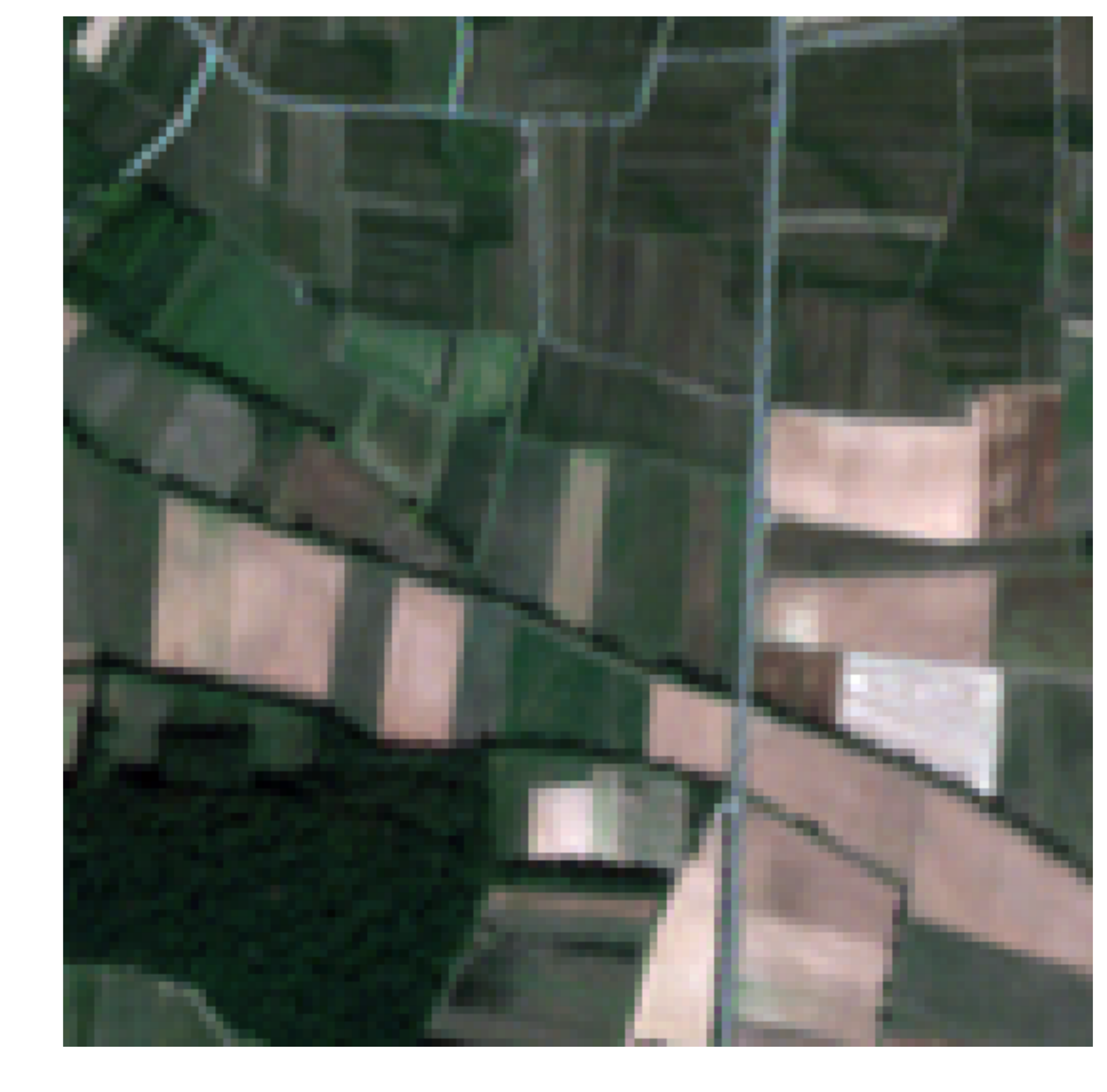}
        \caption{Image from the sequence.}
        \label{fig:qualisup4:rgb}
        \end{subfigure}
    \hfill
        \begin{subfigure}{0.24\textwidth}
        \begin{tikzpicture}
         \node[anchor=south west,inner sep=0] (image) at (0,0) {\includegraphics[width=\textwidth]{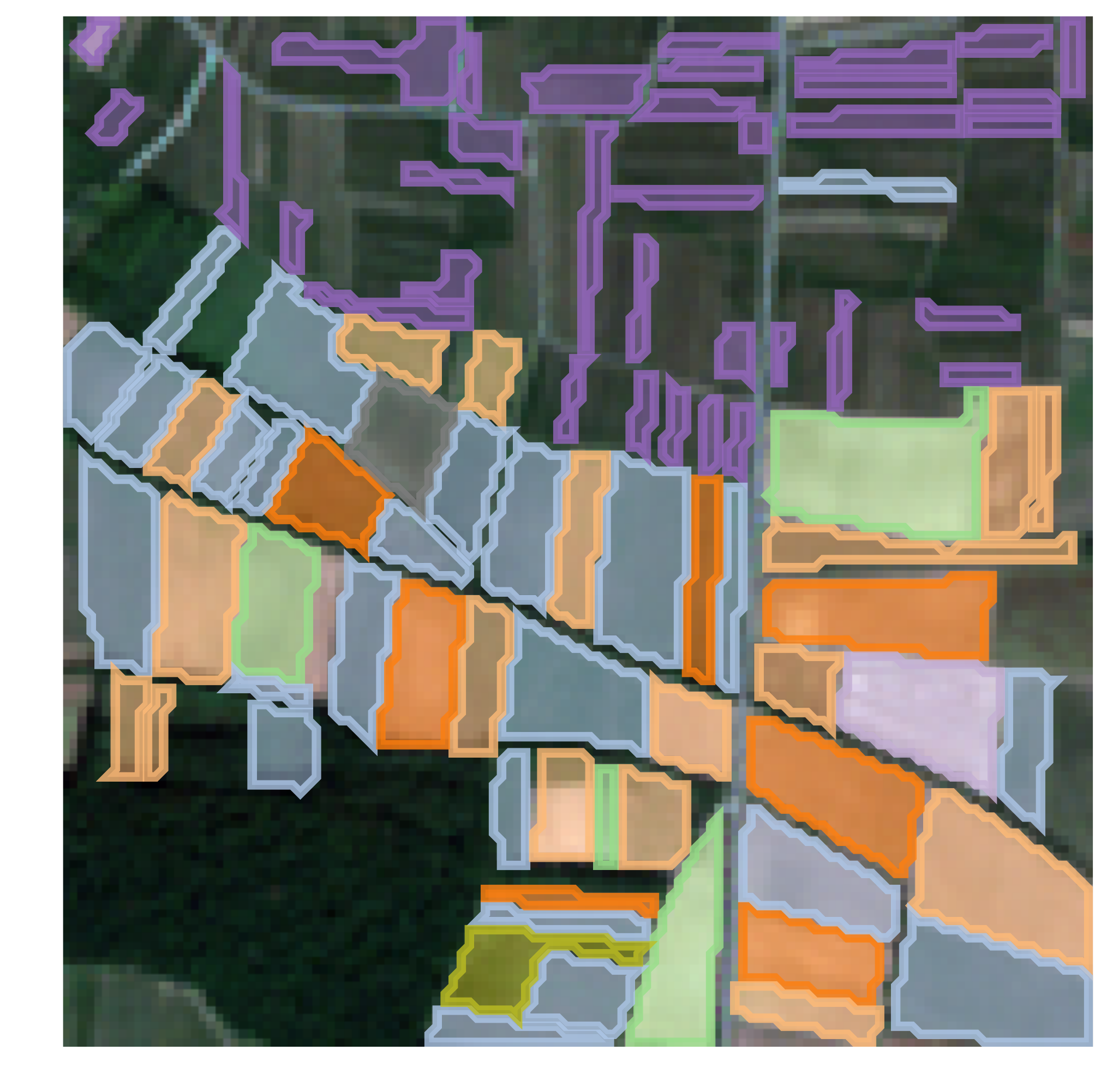}};
        \begin{scope}[x={(image.south east)},y={(image.north west)}]
        \draw[red,ultra thick] (0.7,0.8) circle (0.18);
        \end{scope}
        \end{tikzpicture}

        \caption{Panoptic annotation.}
        \label{fig:qualisup4:gt}
        \end{subfigure}
    \hfill
        \begin{subfigure}{0.24\textwidth}
        \begin{tikzpicture}
         \node[anchor=south west,inner sep=0] (image) at (0,0) {\includegraphics[width=\textwidth]{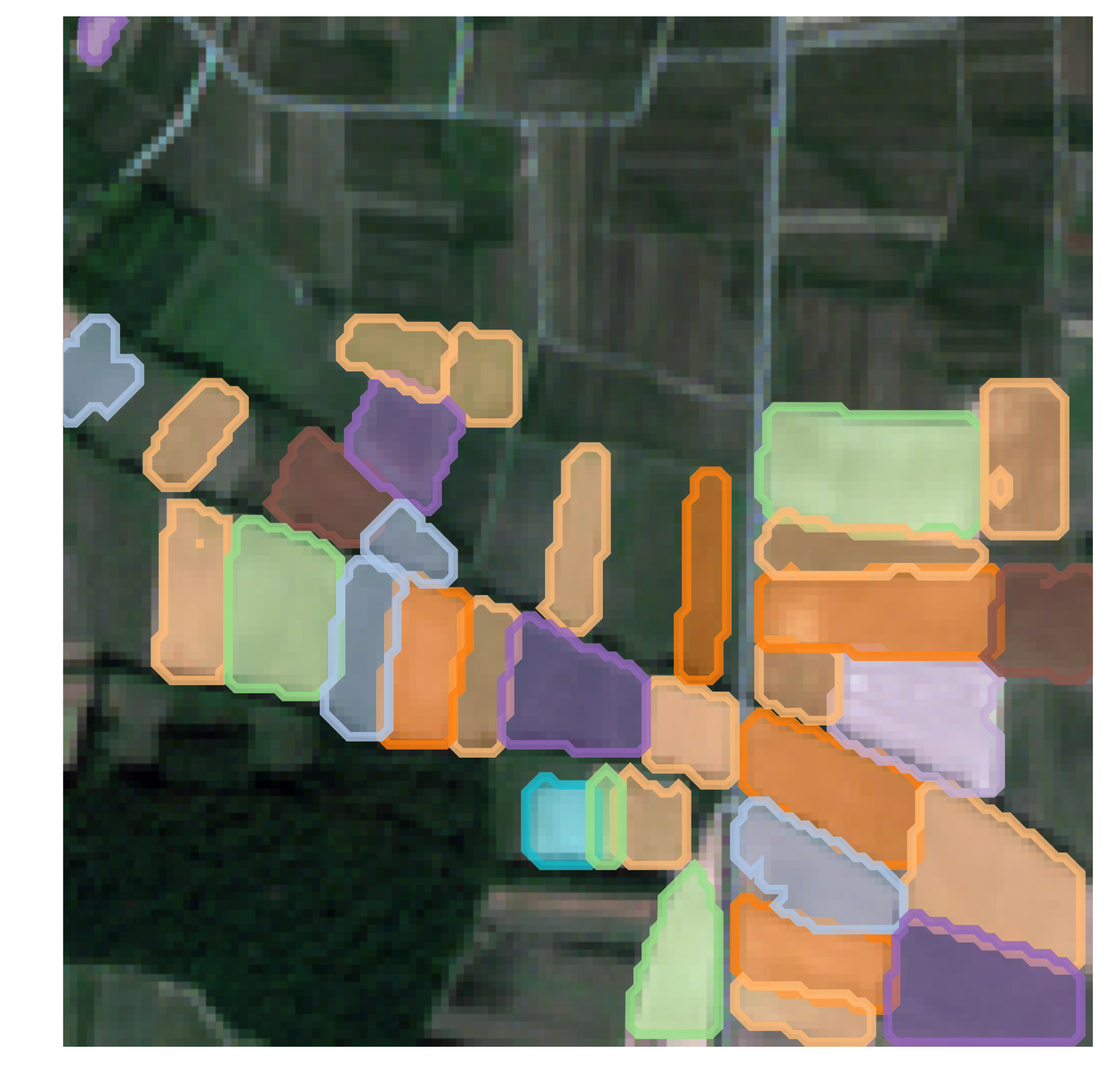}};
        \begin{scope}[x={(image.south east)},y={(image.north west)}]
        \draw[blue ,ultra thick] (0.4,0.4) circle (0.1);
        \end{scope}
        \end{tikzpicture}

        \caption{Panoptic segmentation.}
        \label{fig:qualisup4:pan}
        \end{subfigure}
    \hfill
        \begin{subfigure}{0.24\textwidth}
        \includegraphics[width=\textwidth]{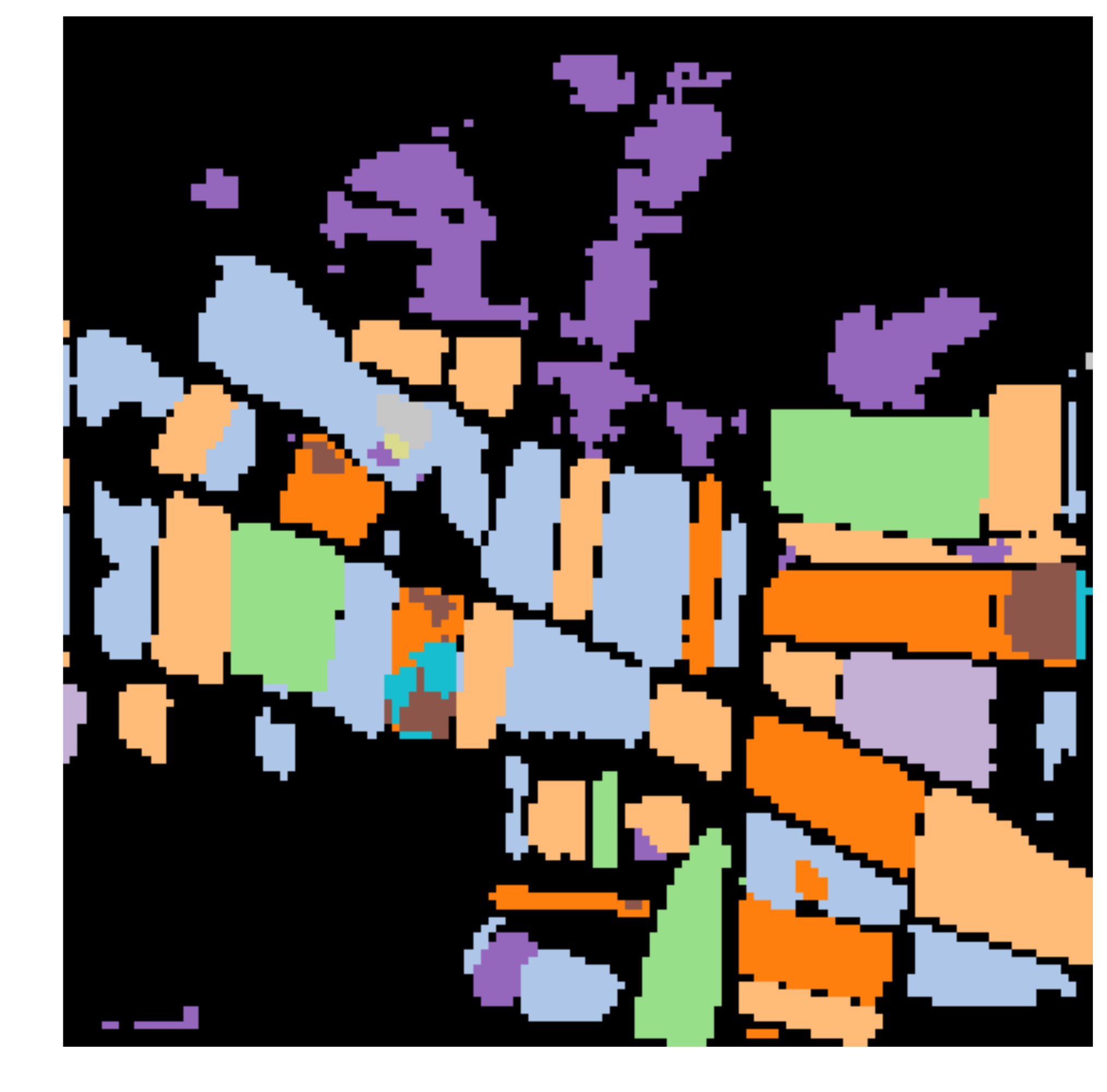}
        \caption{Semantic segmentation.}
        \label{fig:qualisup4:seg}
        \end{subfigure}
\caption{\textbf{Qualitative  Panoptic Segmentation Results.} We represent a single image from the sequence using the RGB channels (\subref{fig:qualisup4:rgb}), and whose ground truth parcel's limit and types are known (\subref{fig:qualisup4:gt}). We then represent the parcels predicted by our panoptic segmentation module (\subref{fig:qualisup4:pan}), and the pixelwise prediction of our semantic segmentation module (\subref{fig:qualisup4:seg}). See \figref{fig:nomenc} for the color to crop type correspondence. 
We highlight with a green circle \protect\tikz \protect\node[circle, thick, draw = green!90!black, fill = none, scale = 0.7] {}; a large, fragmented parcel declared as one single field. This leads to predictions with low confidence and a low panoptic quality.
Conversely, the cyan circle \protect\tikz \protect\node[circle, thick, draw = cyan, fill = none, scale = 0.7] {}; highlights such fragmented parcel which is correctly predicted as a single instance. This suggests that our network is able to use the temporal dynamics to recover ambiguous borders.
We highlight a failure case with the red circle \protect\tikz \protect\node[circle, thick, draw = red, fill = none, scale = 0.7] {};, for which many thin parcels are not properly detected, resulting in a low panoptic quality. {We observe that the semantic segmentation model struggles as well for such thin parcels.}  Finally, we 
highlight with a blue circle \protect\tikz \protect\node[circle, thick, draw = blue, fill = none, scale = 0.7] {}; an example in which the panoptic prediction is superior to the semantic segmentation, indicating that detecting parcels' boundaries and extent can be informative for their classification.
}
\label{fig:qualisup}
\end{figure*}

\begin{figure*}

\begin{subfigure}{0.16\textwidth}
        \centering
        \includegraphics[width=\textwidth]{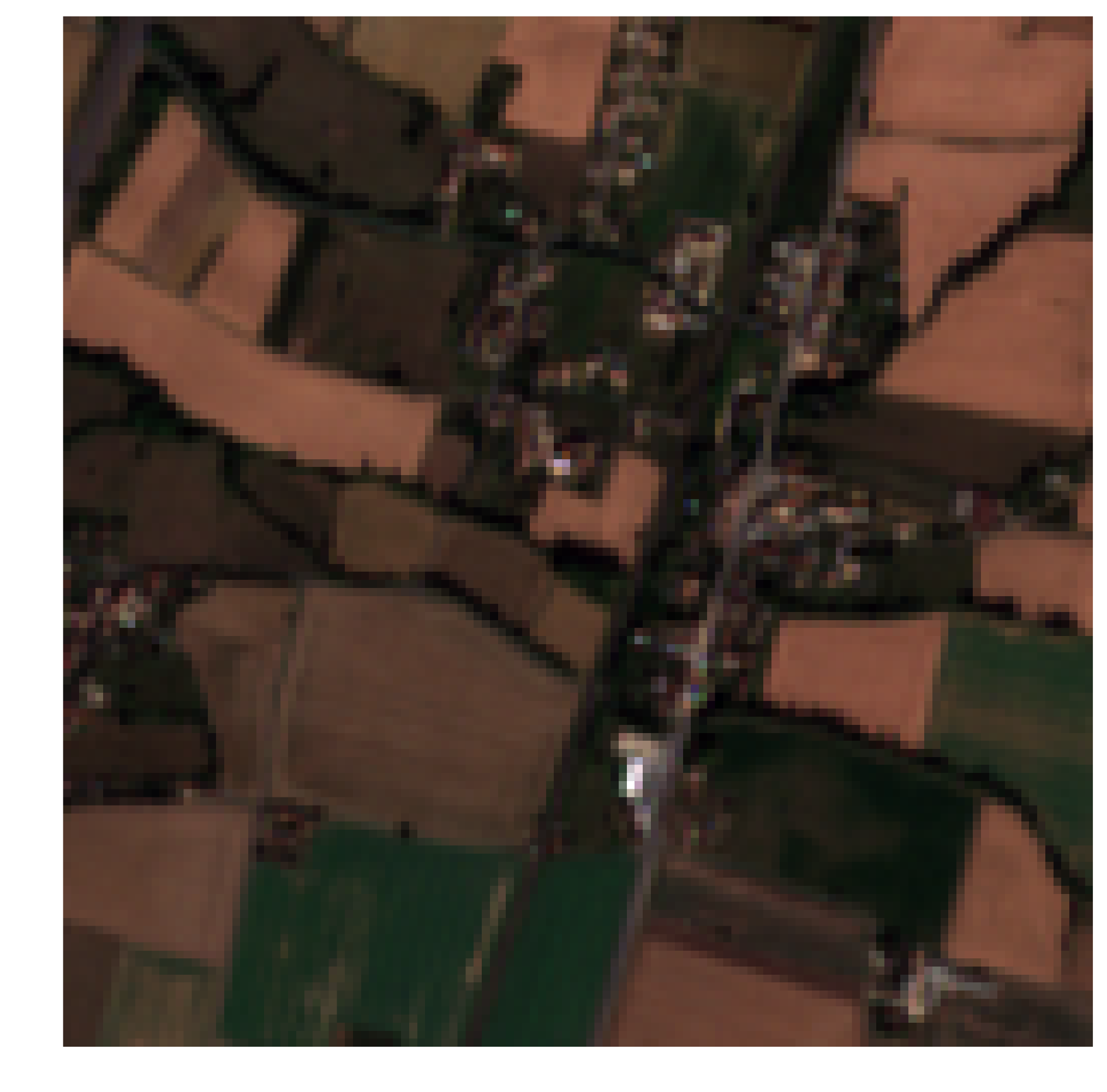}
        \end{subfigure}
    \hfill
        \begin{subfigure}{0.16\textwidth}
        \includegraphics[width=\textwidth]{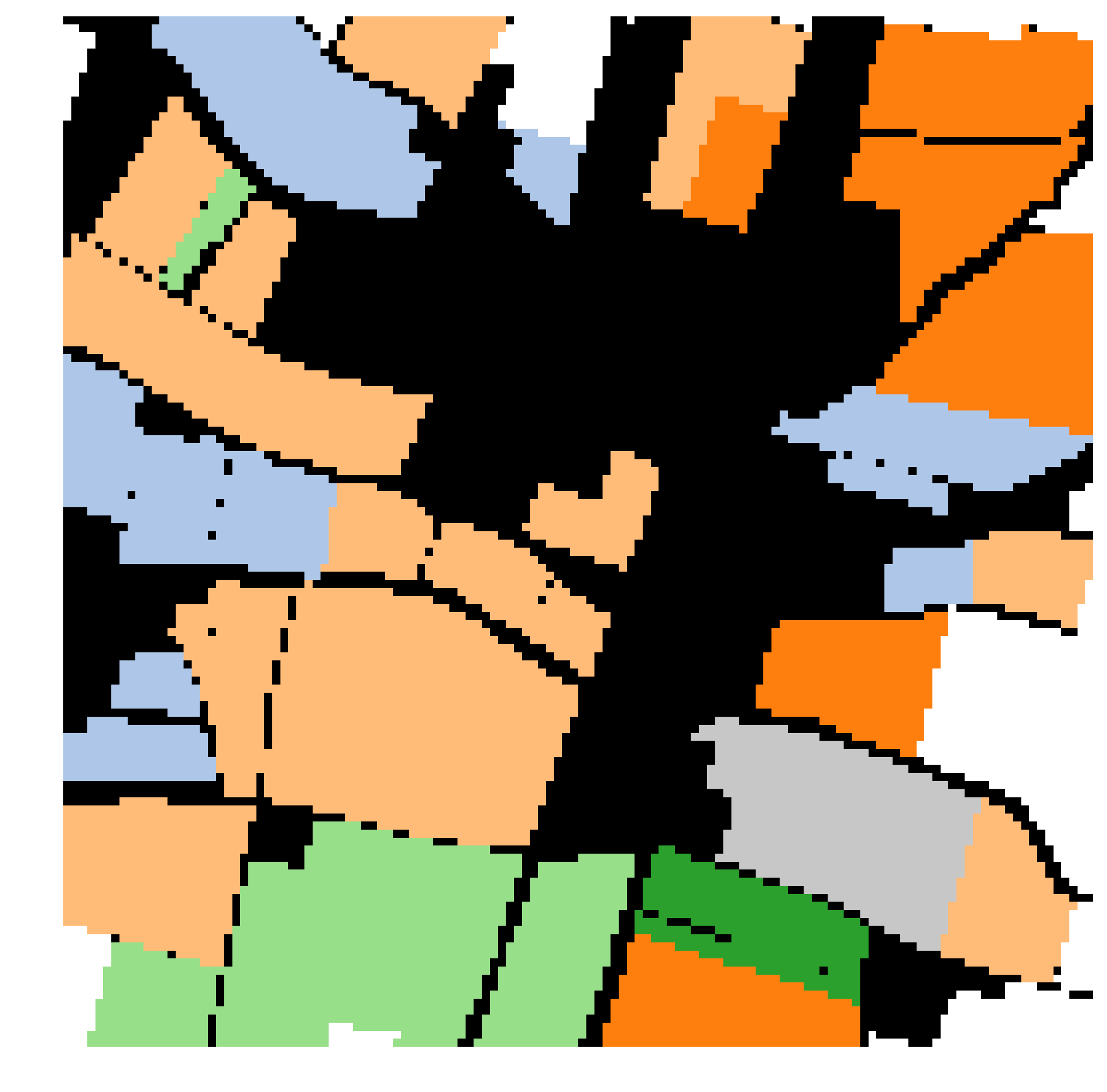}
        \end{subfigure}
    \hfill
        \begin{subfigure}{0.16\textwidth}
        \includegraphics[width=\textwidth]{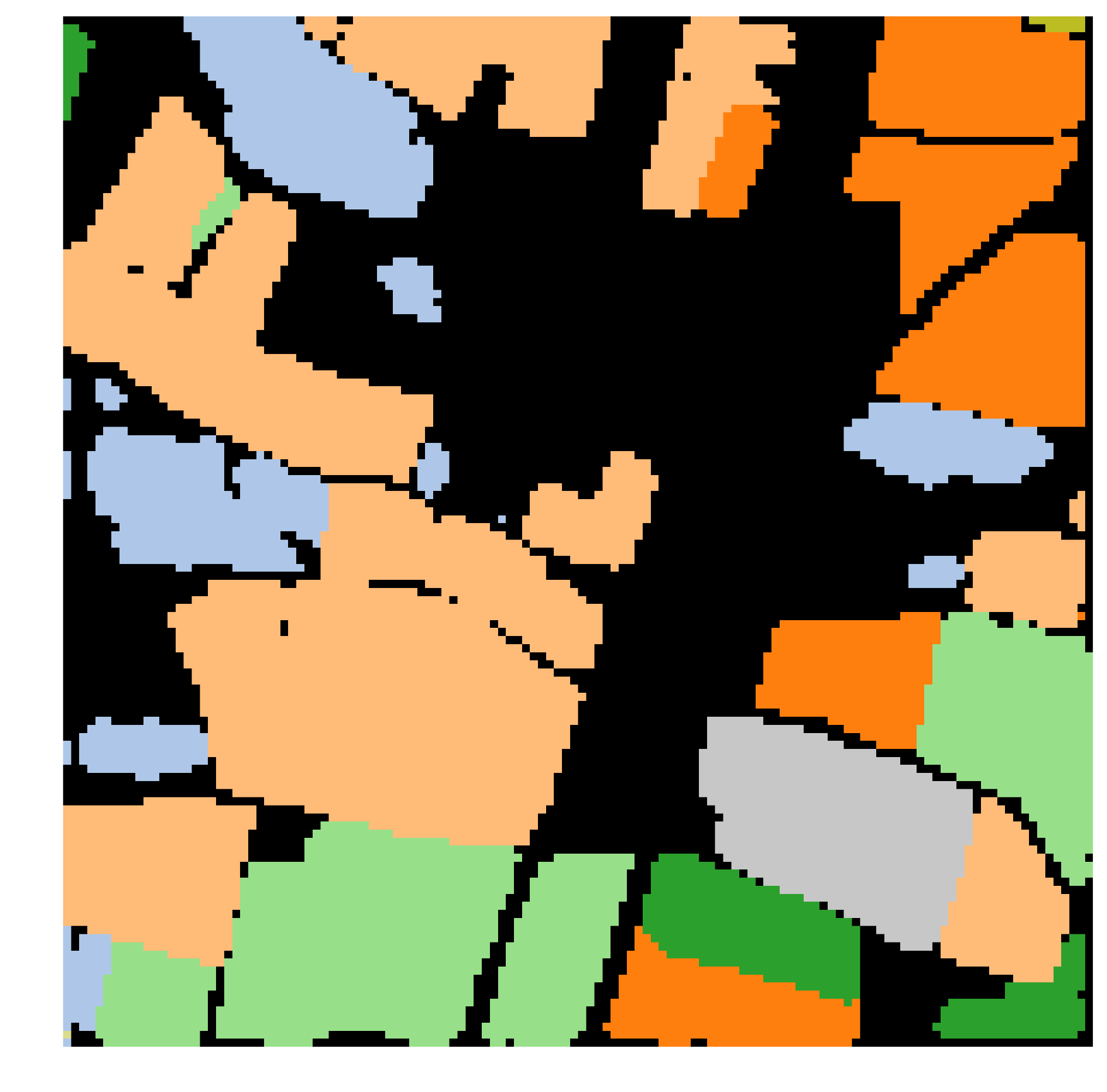}
        \end{subfigure}
    \hfill
        \begin{subfigure}{0.16\textwidth}
        \includegraphics[width=\textwidth]{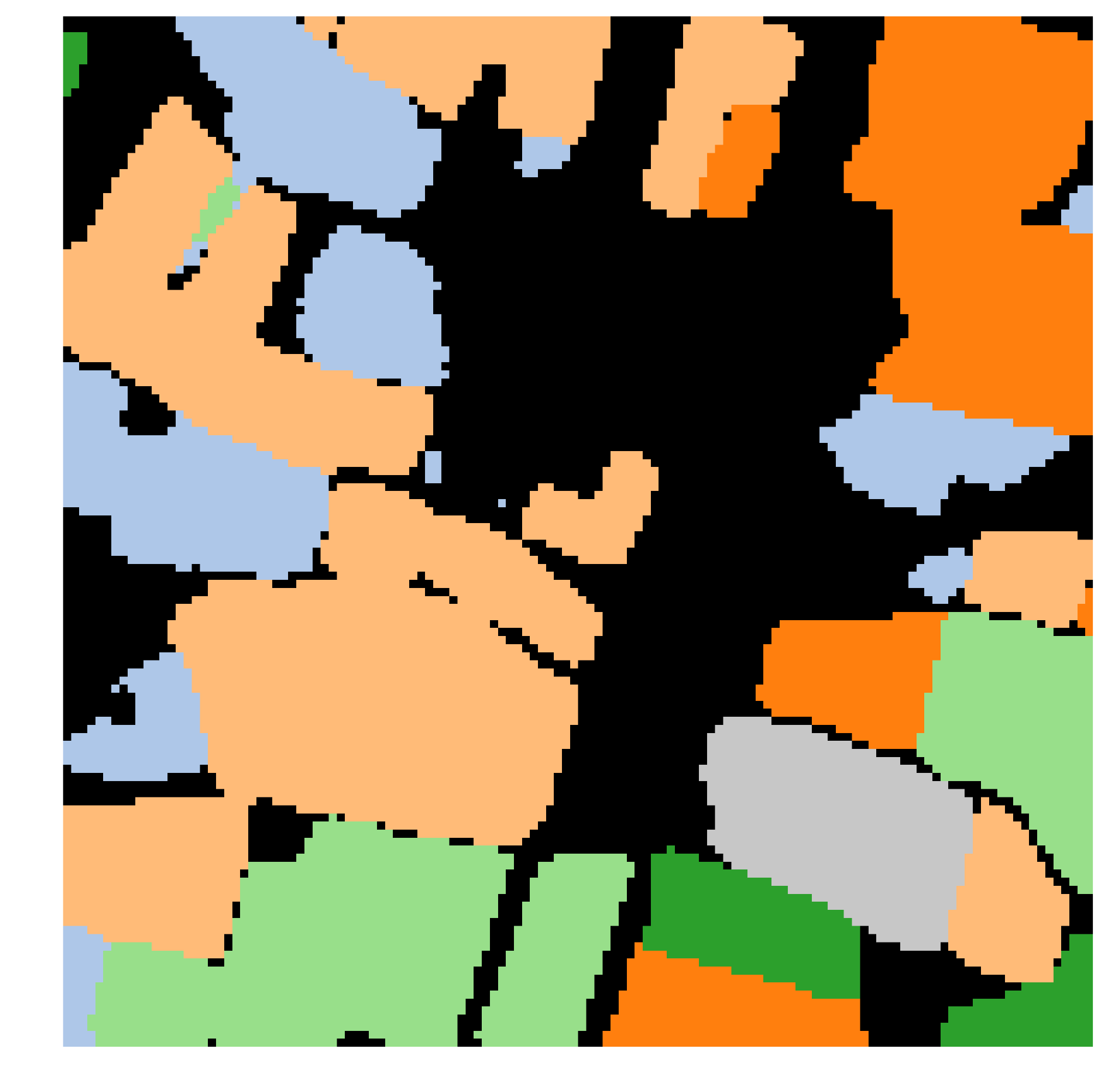}
        \end{subfigure}
    \hfill
        \begin{subfigure}{0.16\textwidth}
        \includegraphics[width=\textwidth]{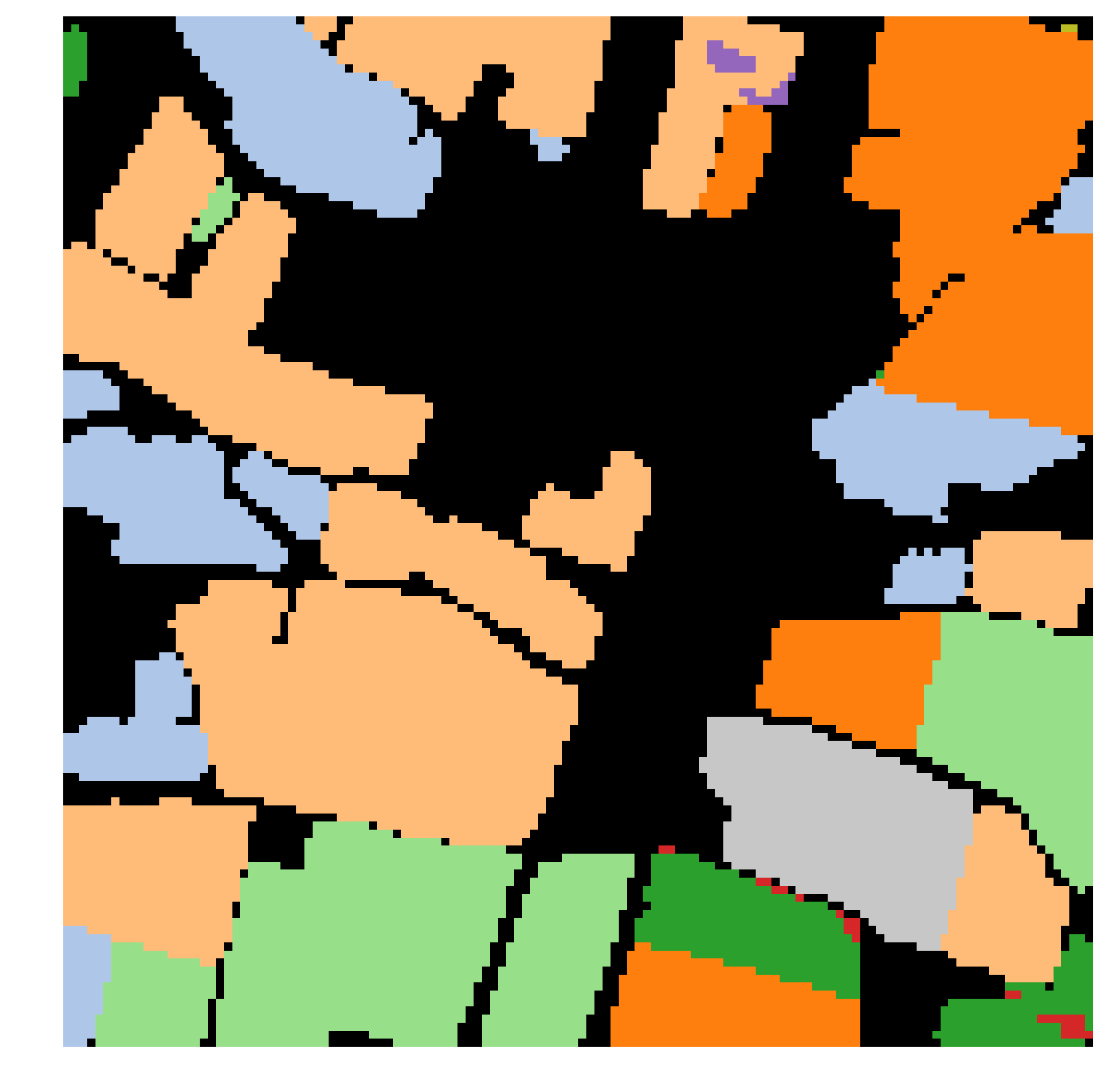}
        \end{subfigure}
\hfill
        \begin{subfigure}{0.16\textwidth}
        \includegraphics[width=\textwidth]{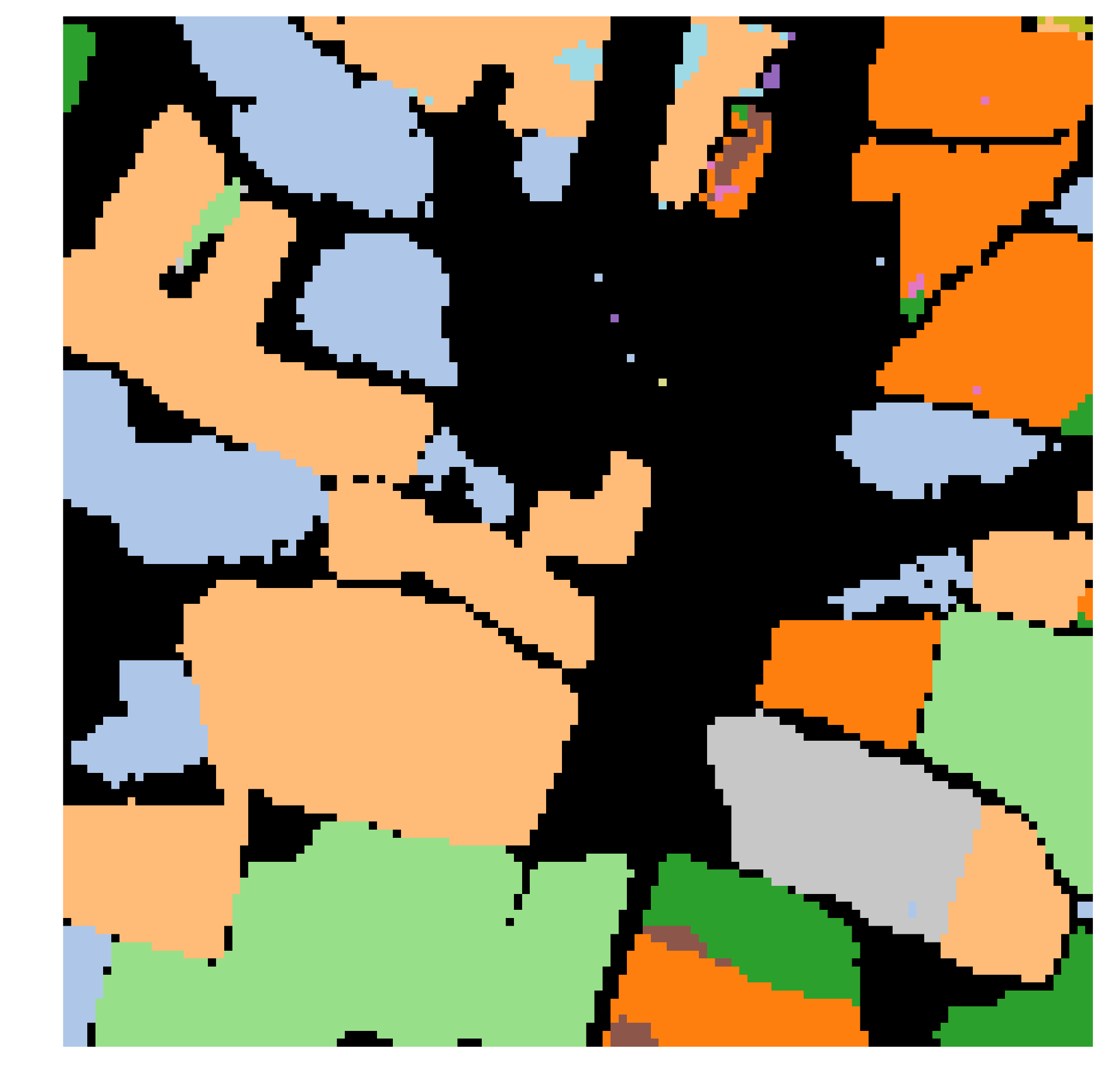}
        \end{subfigure}
\vfill
\begin{subfigure}{0.16\textwidth}
        \centering
        \includegraphics[width=\textwidth]{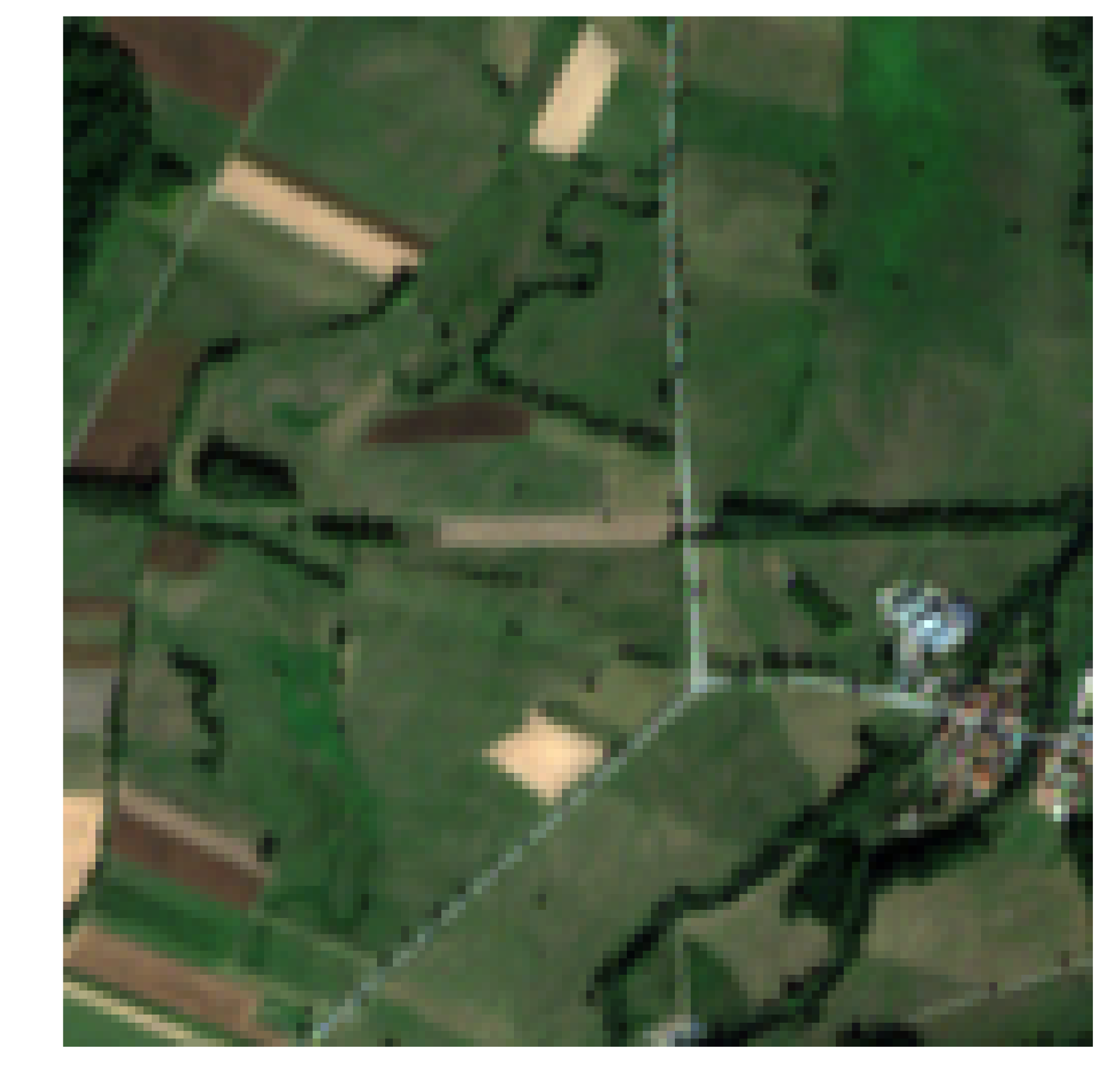}
        \end{subfigure}
    \hfill
        \begin{subfigure}{0.16\textwidth}
        \includegraphics[width=\textwidth]{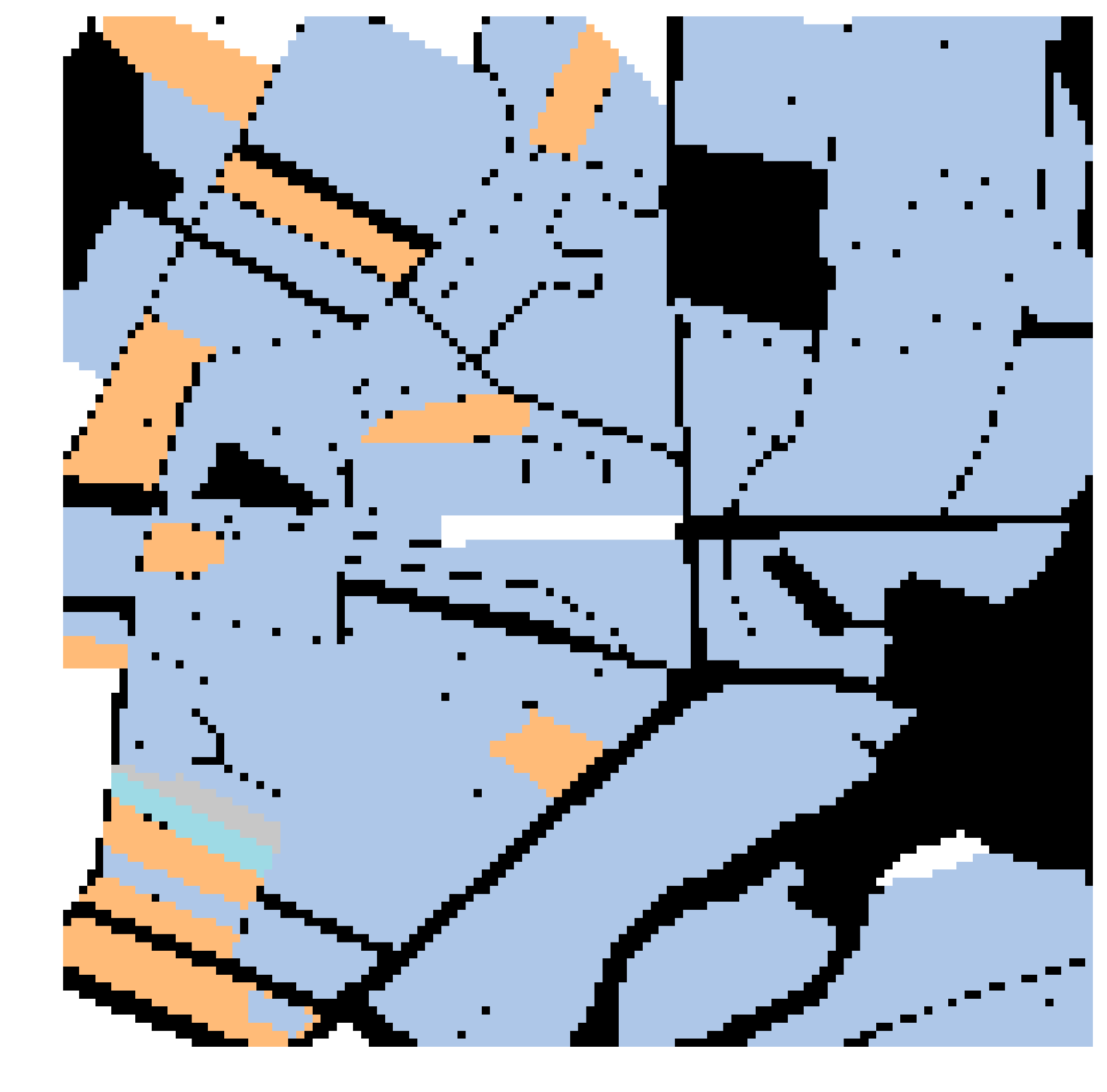}
        \end{subfigure}
    \hfill
        \begin{subfigure}{0.16\textwidth}
        \includegraphics[width=\textwidth, trim=0cm 4cm 0cm 4cm, clip]{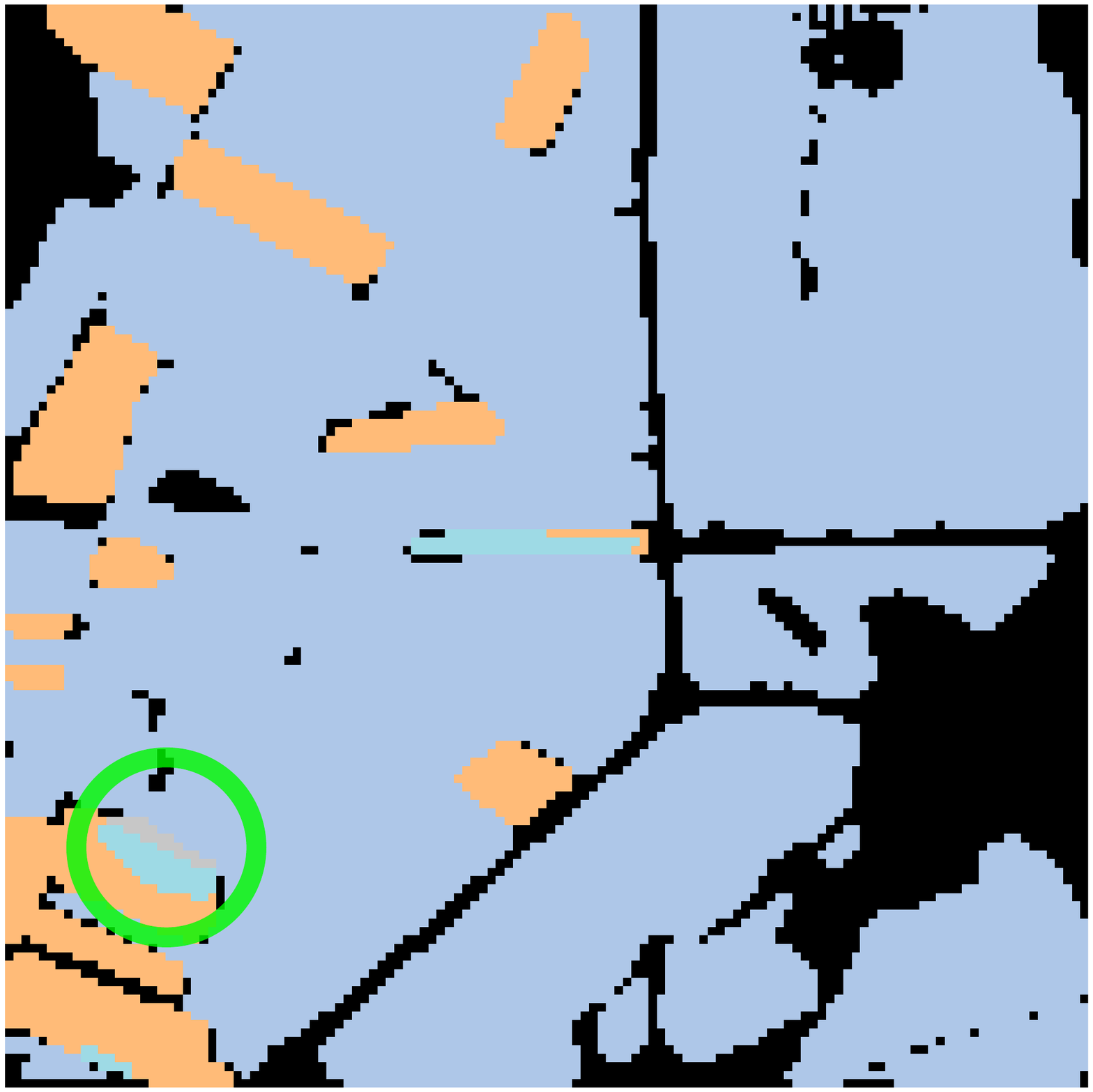}
        \end{subfigure}
    \hfill
        \begin{subfigure}{0.16\textwidth}
        \includegraphics[width=\textwidth]{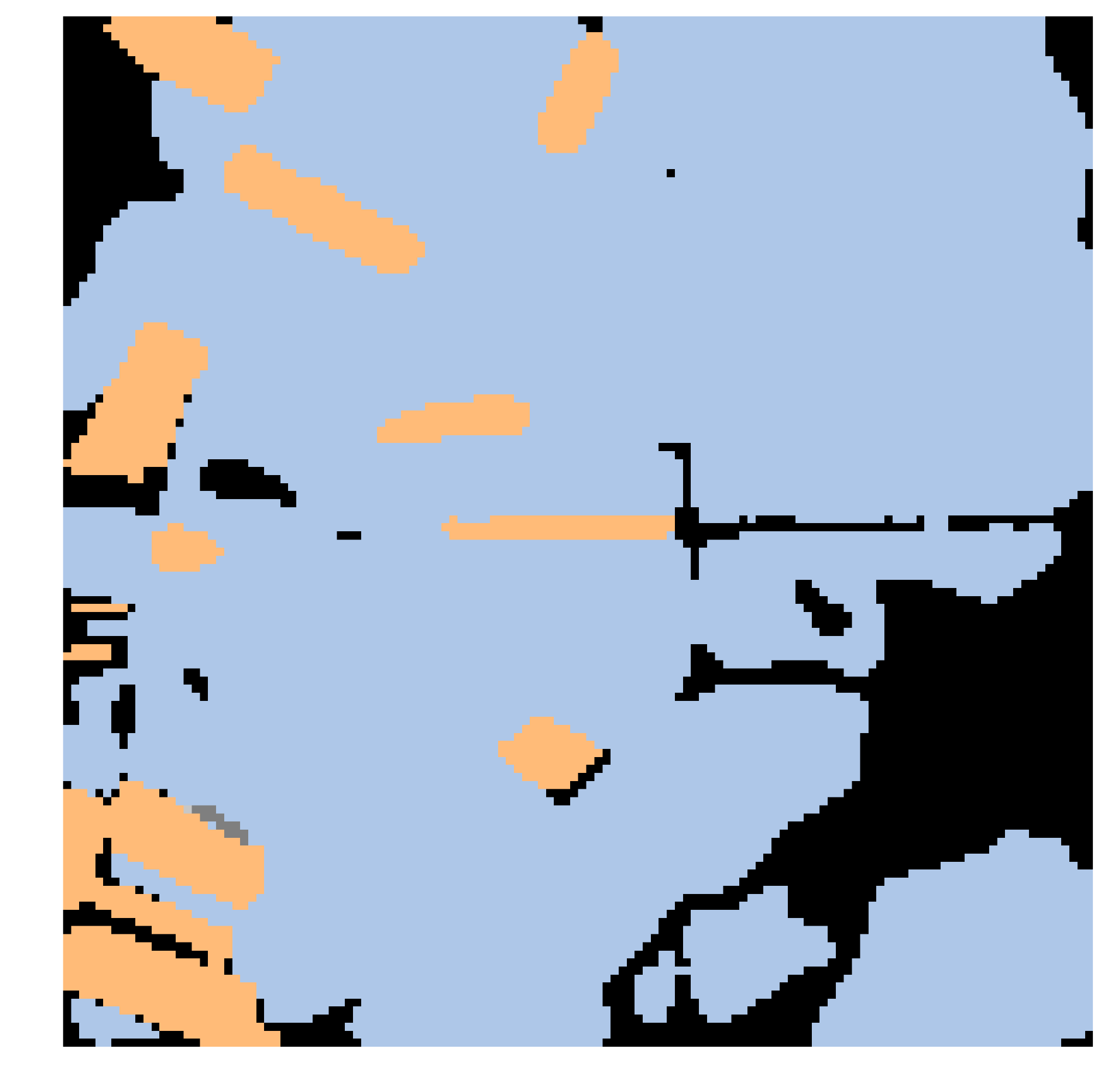}
        \end{subfigure}
    \hfill
        \begin{subfigure}{0.16\textwidth}
        \includegraphics[width=\textwidth]{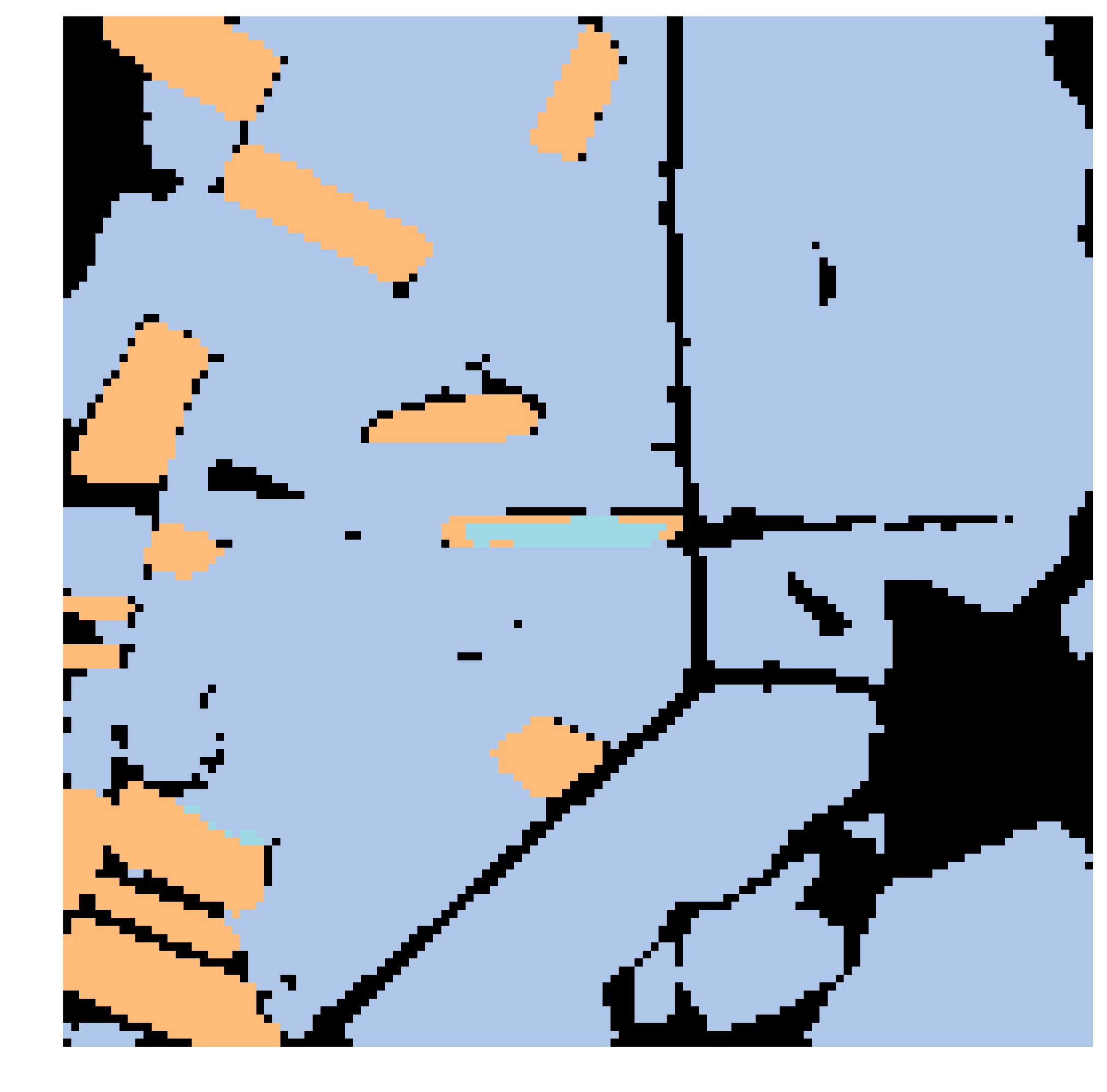}
        \end{subfigure}
\hfill
        \begin{subfigure}{0.16\textwidth}
        \includegraphics[width=\textwidth]{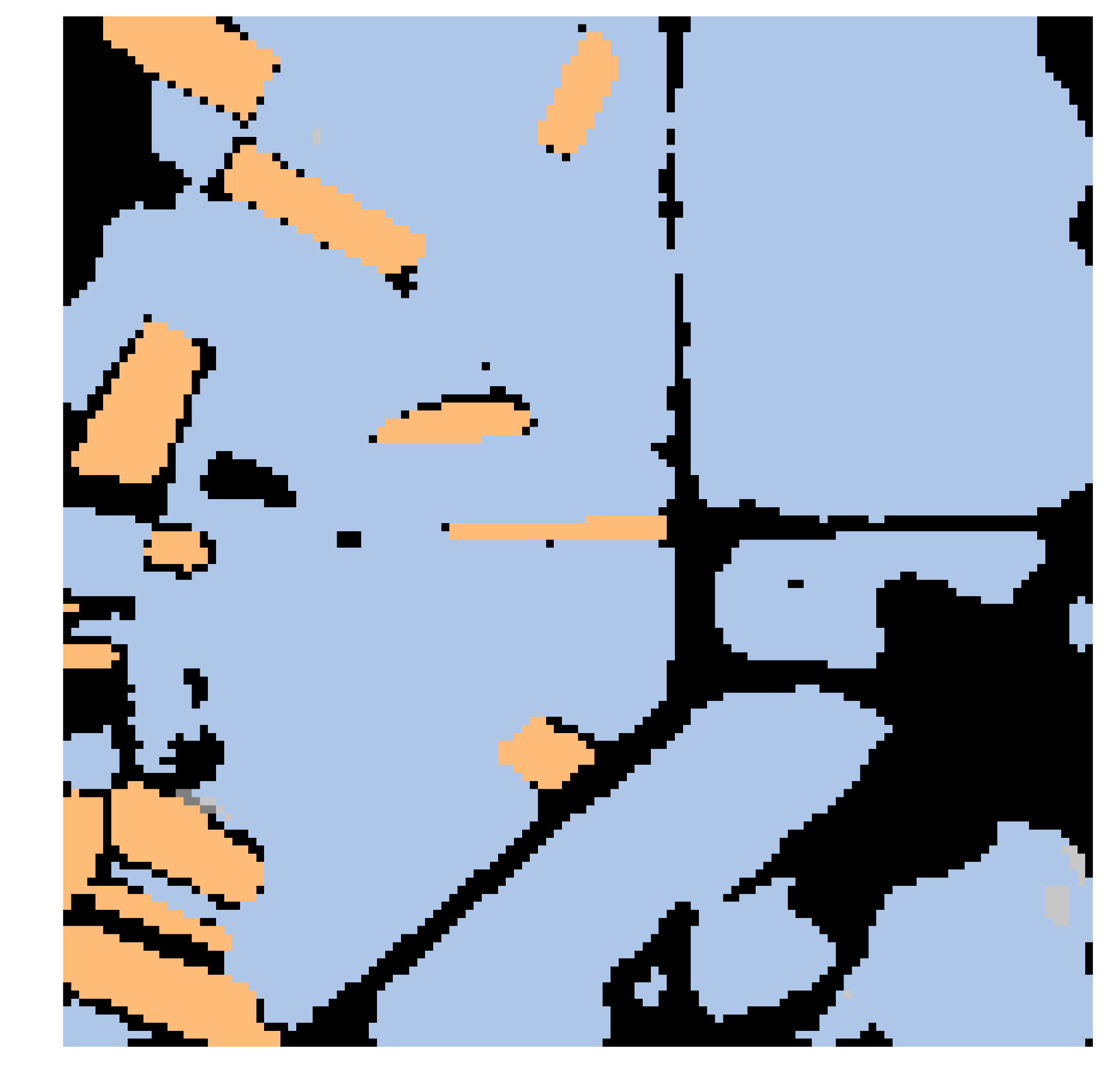}
        \end{subfigure}
\vfill
\begin{subfigure}{0.16\textwidth}
        \centering
        \includegraphics[width=\textwidth]{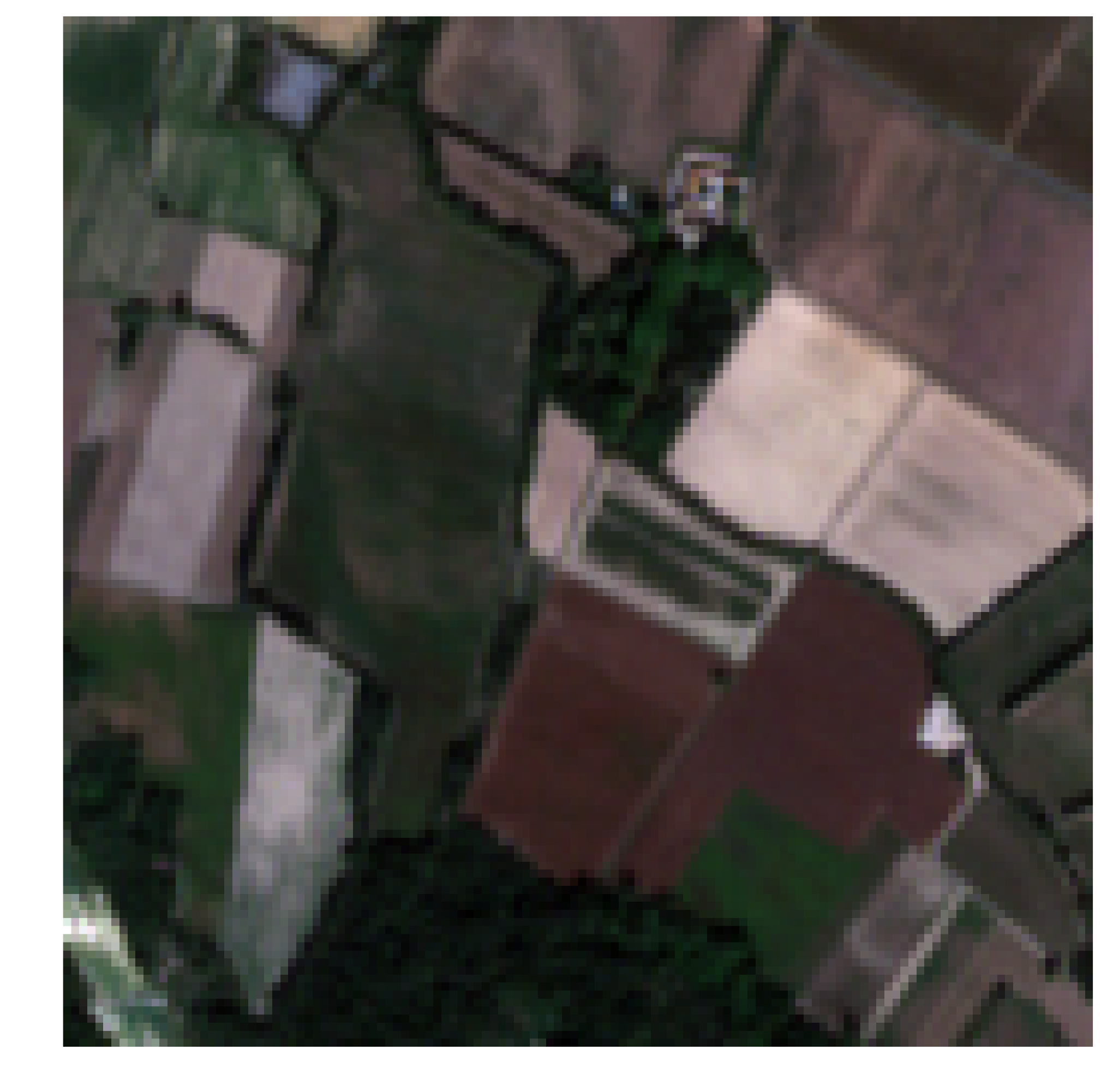}
        \end{subfigure}
    \hfill
        \begin{subfigure}{0.16\textwidth}
        \includegraphics[width=\textwidth]{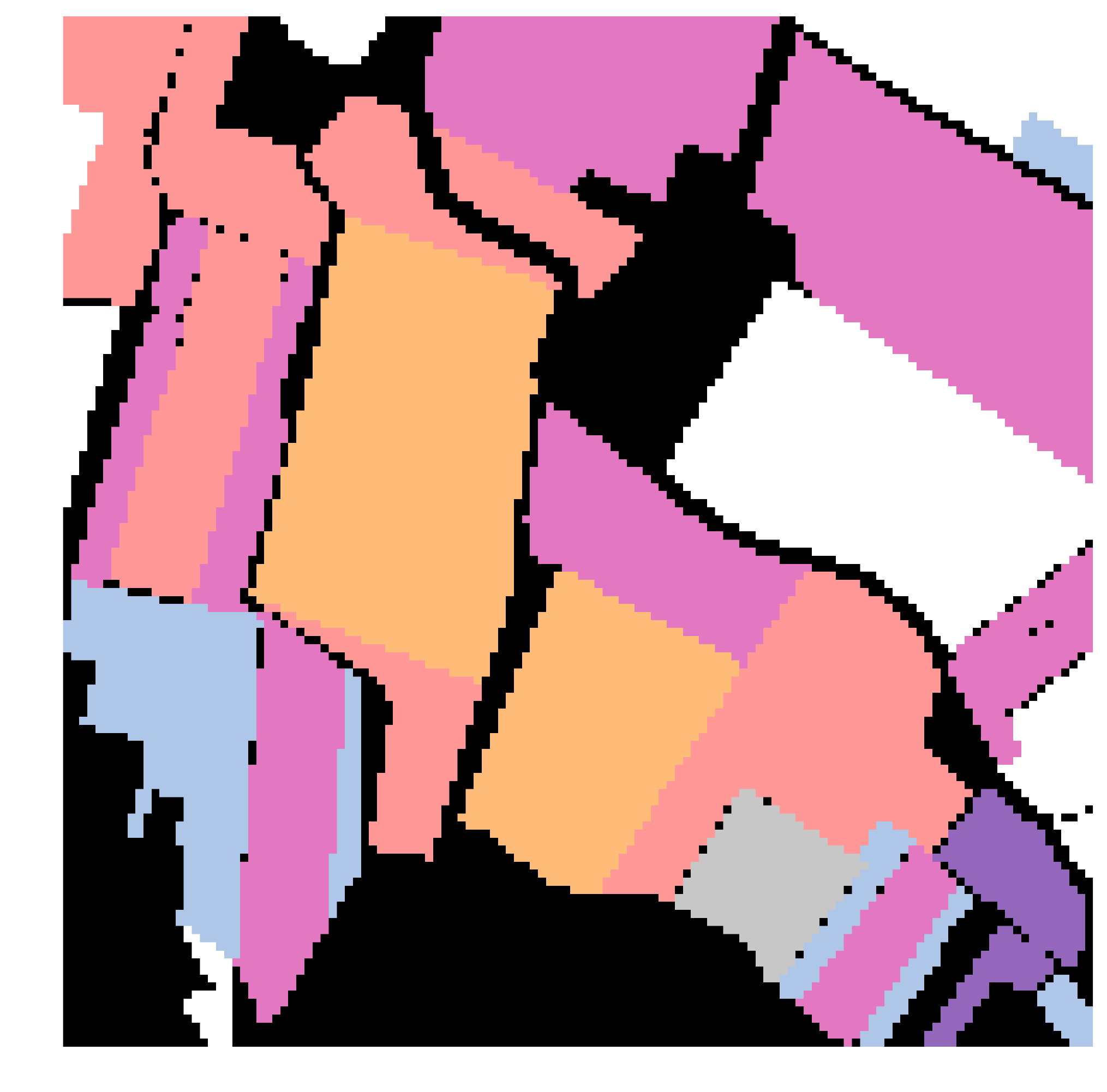}
        \end{subfigure}
    \hfill
        \begin{subfigure}{0.16\textwidth}
        \includegraphics[width=\textwidth, trim=0cm 4cm 0cm 4cm, clip]{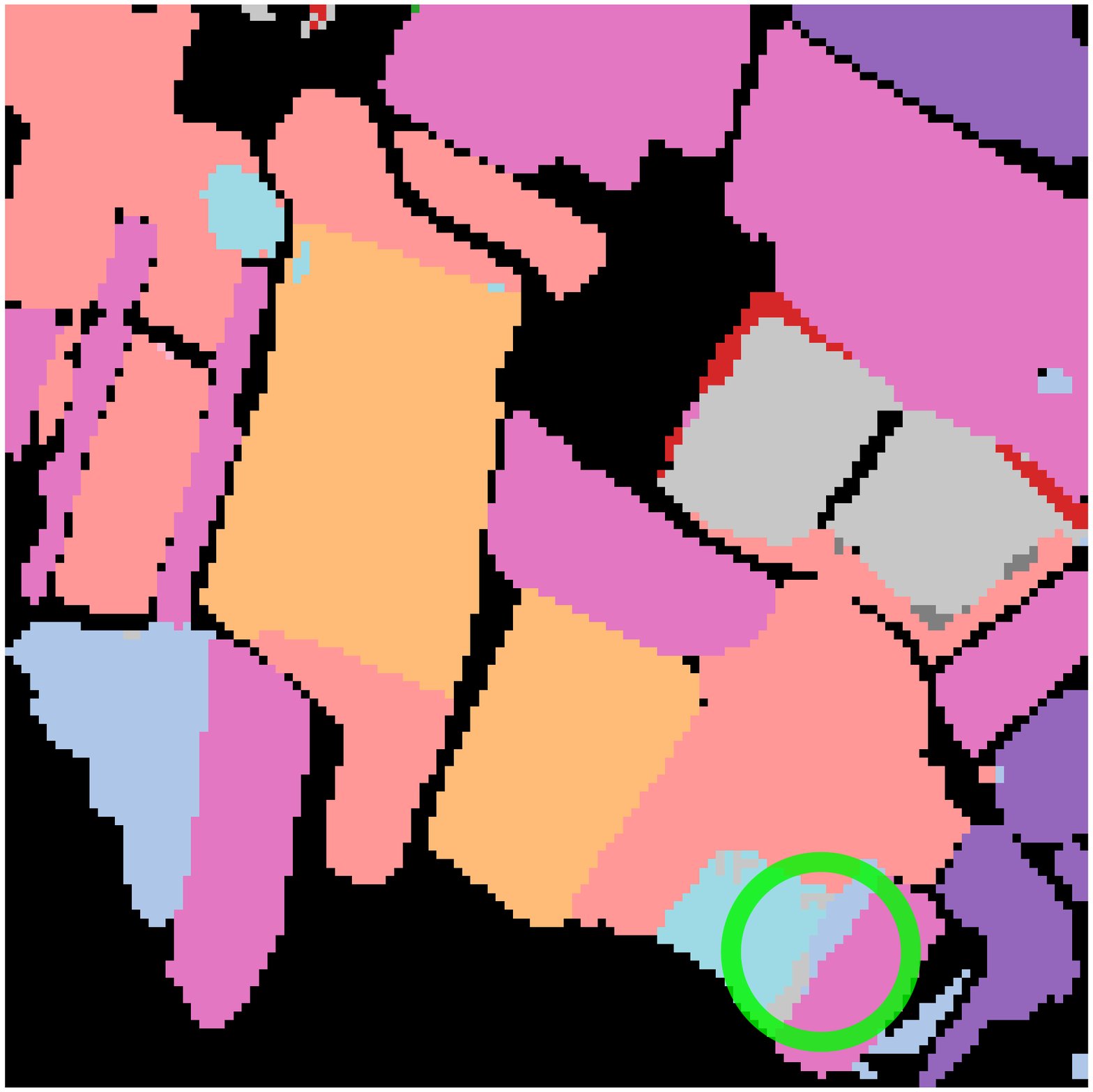}
        \end{subfigure}
    \hfill
        \begin{subfigure}{0.16\textwidth}
        \includegraphics[width=\textwidth]{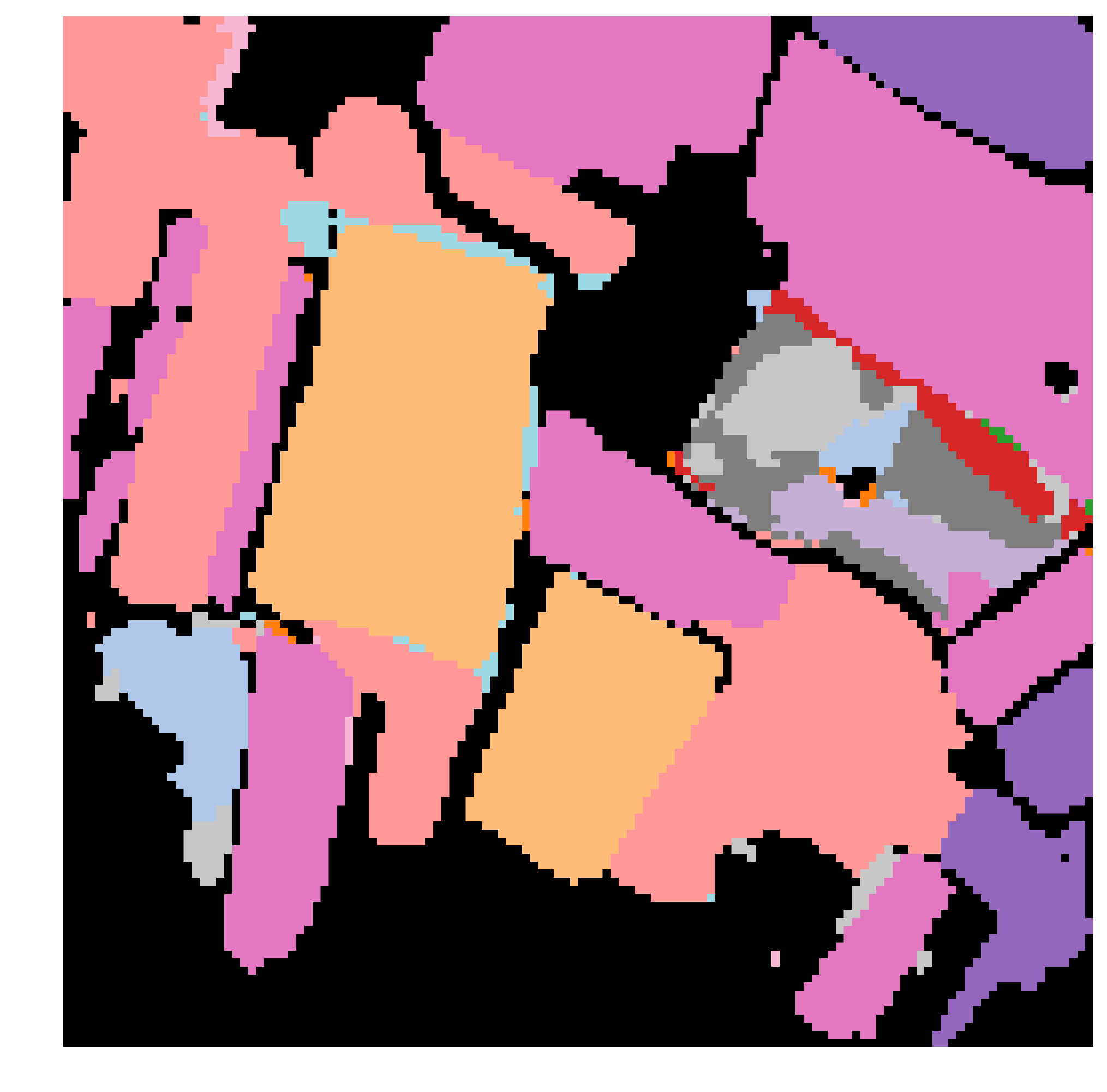}
        \end{subfigure}
    \hfill
        \begin{subfigure}{0.16\textwidth}
        \includegraphics[width=\textwidth]{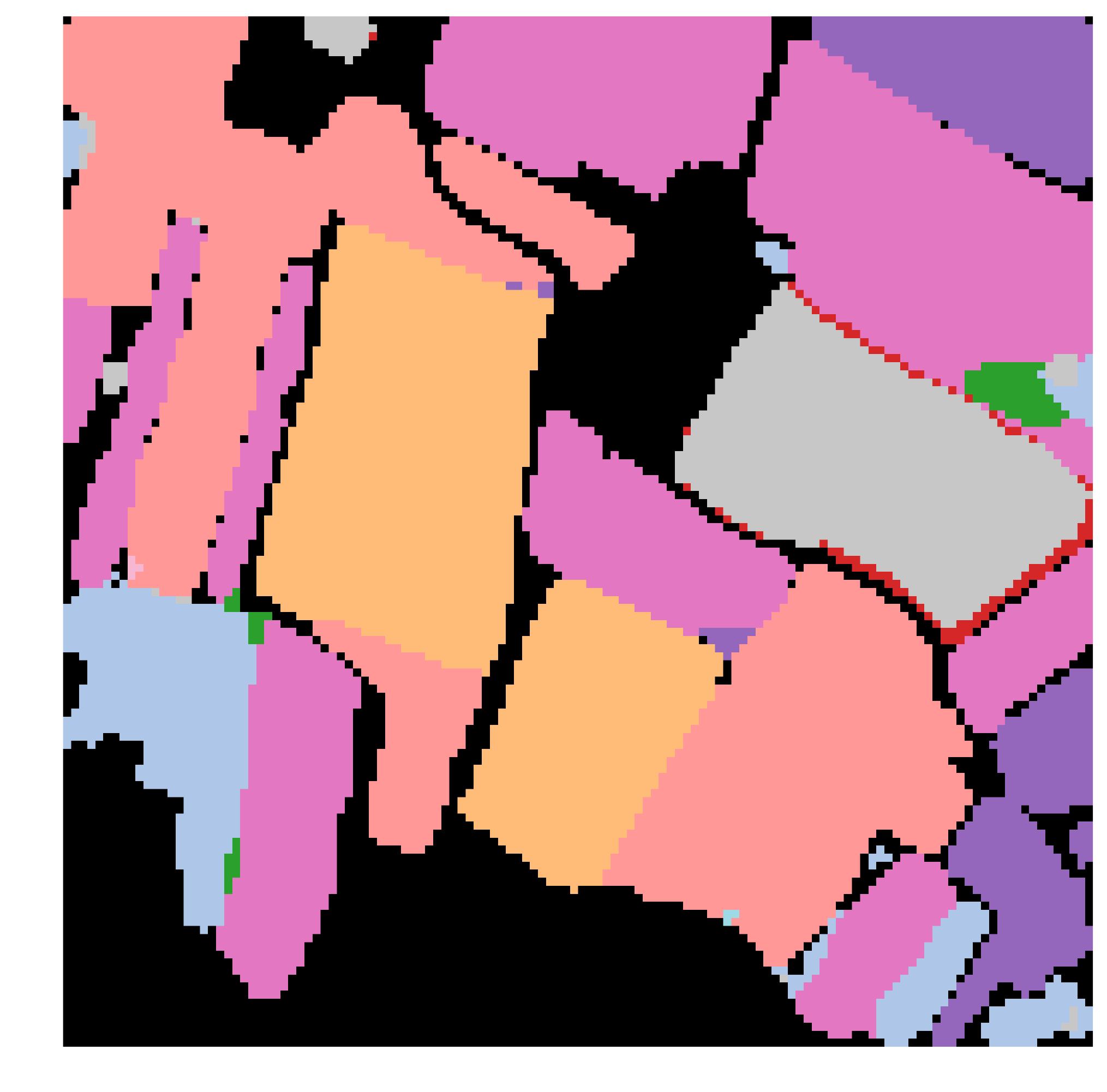}
        \end{subfigure}
\hfill
        \begin{subfigure}{0.16\textwidth}
        \includegraphics[width=\textwidth, trim=0cm 4cm 0cm 4cm, clip]{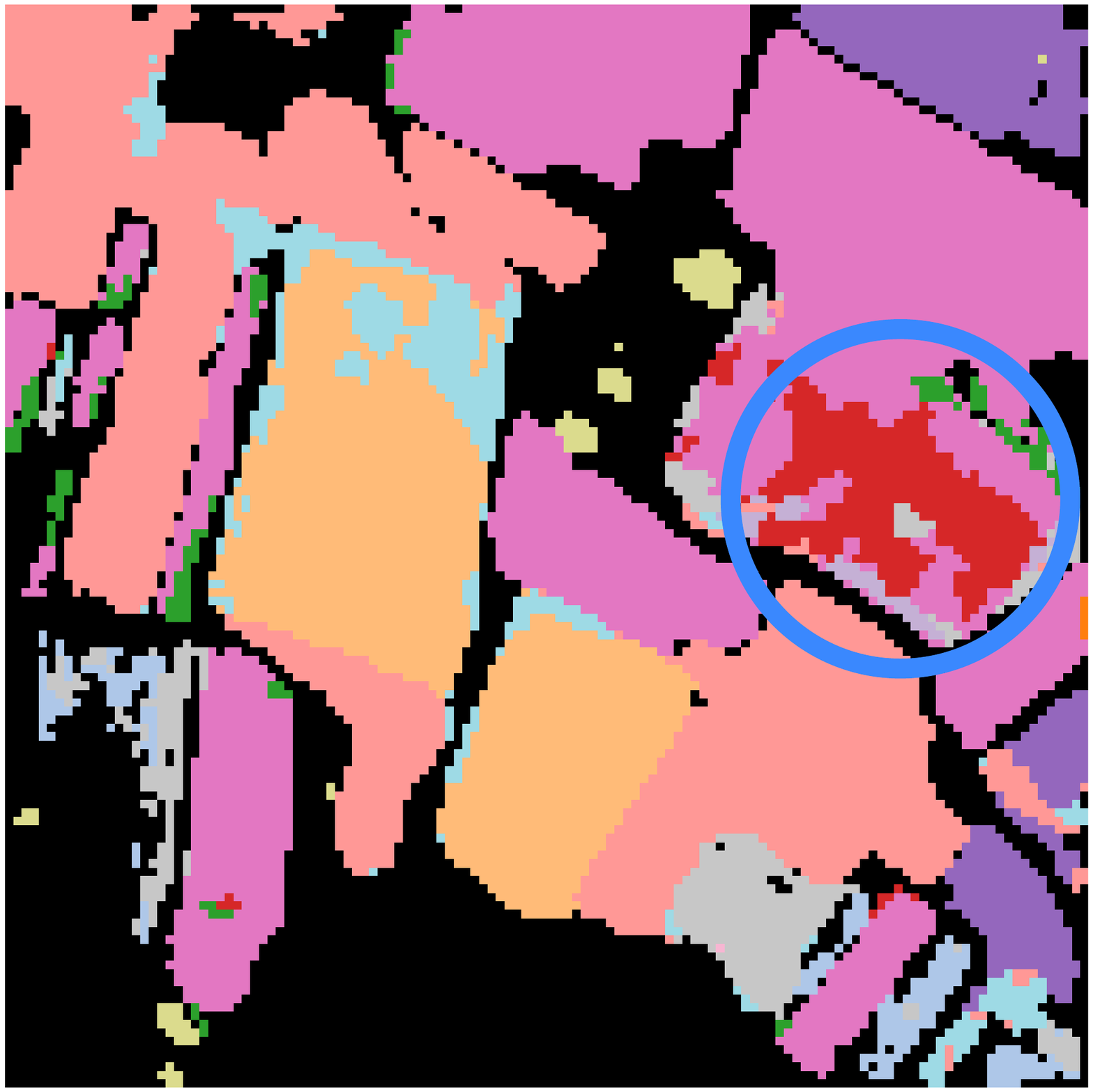}
        \end{subfigure}
\vfill
\begin{subfigure}{0.16\textwidth}
        \centering
        \includegraphics[width=\textwidth]{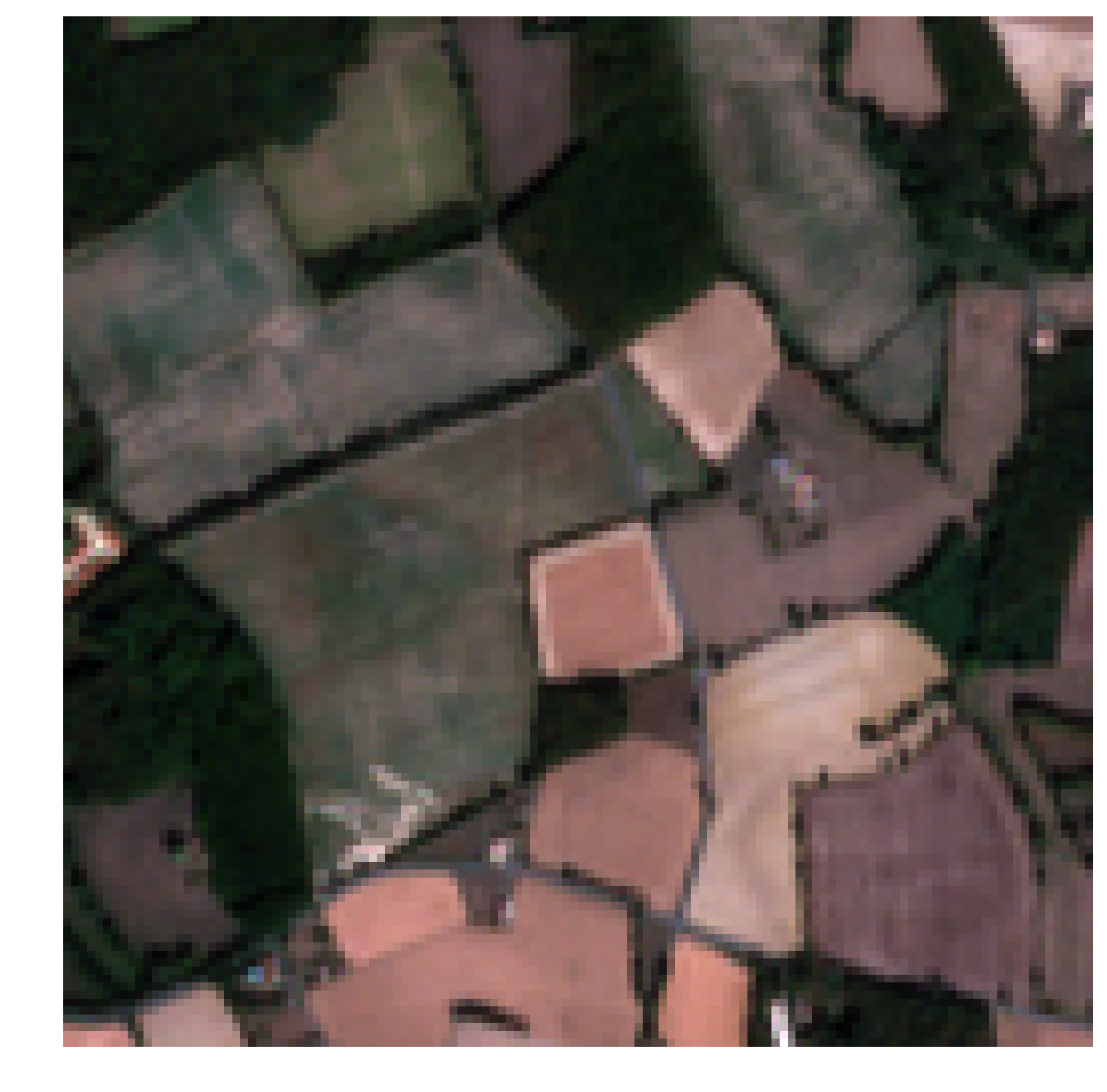}
        \caption{Single image.}
        \label{fig:sem:rgb}
        \end{subfigure}
    \hfill
        \begin{subfigure}{0.16\textwidth}
        \includegraphics[width=\textwidth]{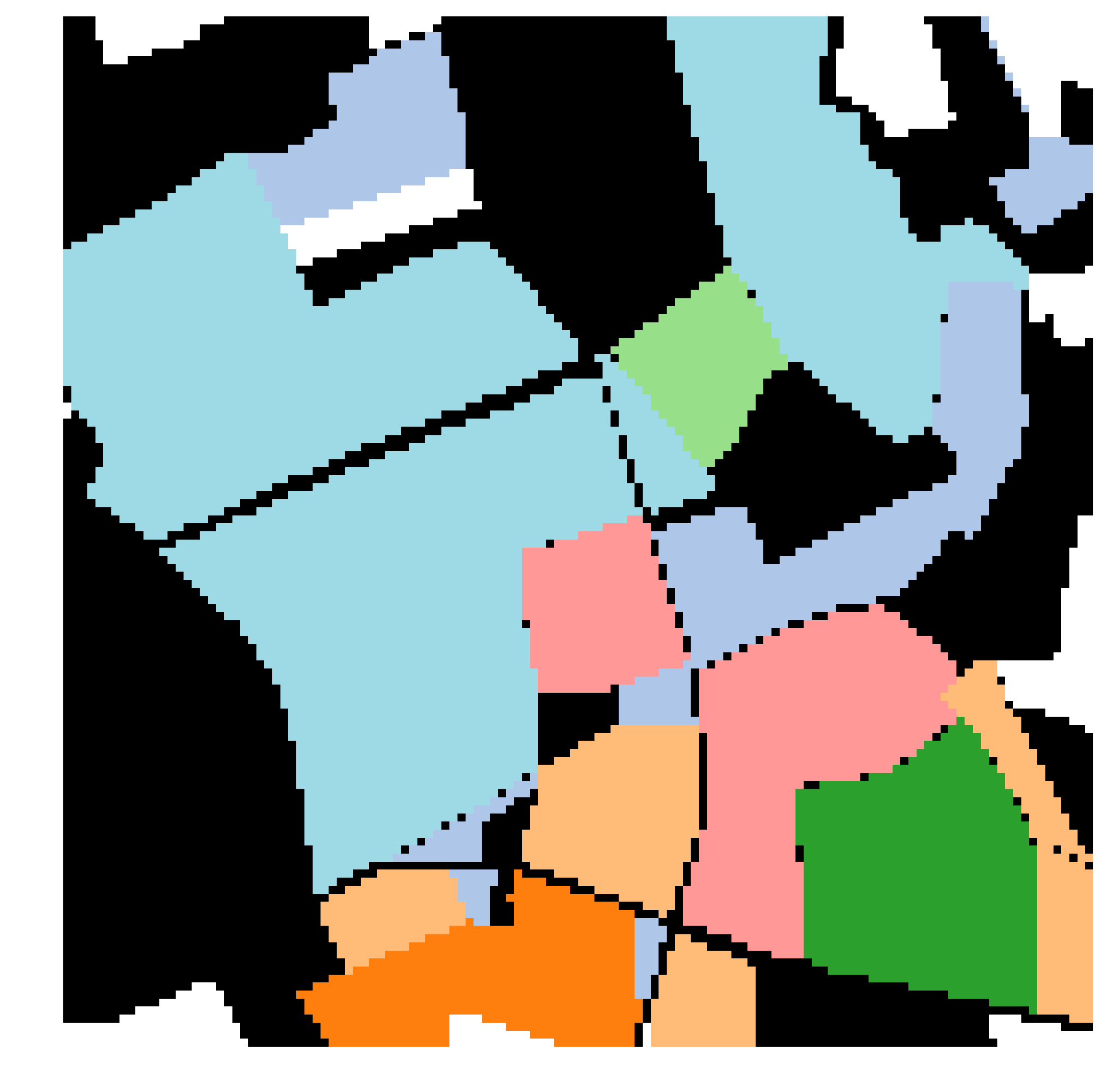}
        \caption{Annotation.}
        \label{fig:sem:gt}
        \end{subfigure}
    \hfill
        \begin{subfigure}{0.16\textwidth}
        \includegraphics[width=\textwidth, trim=0cm 4cm 0cm 4cm, clip]{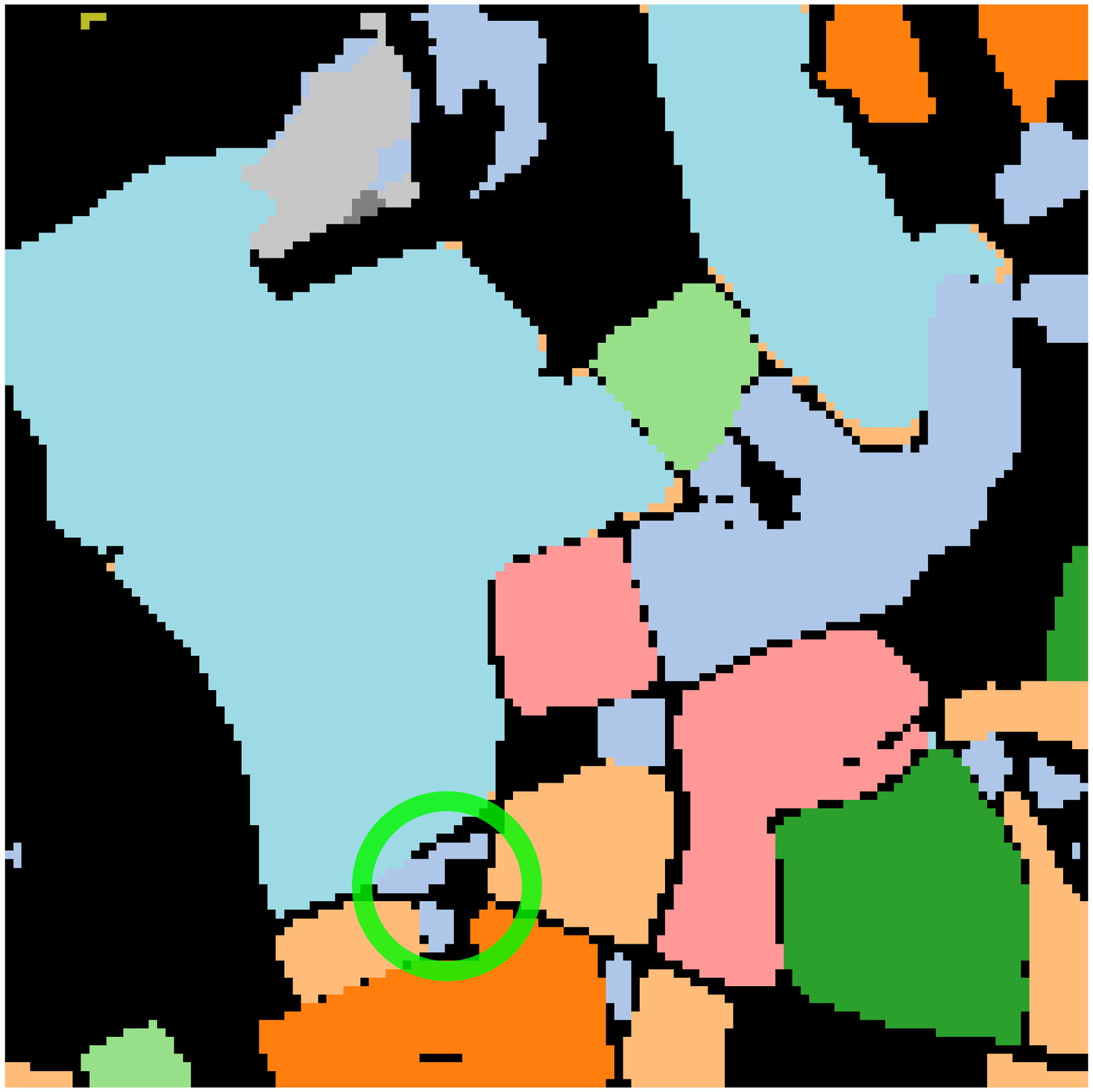}
        \caption{U-TAE.}
        \label{fig:sem:ours}
        \end{subfigure}
    \hfill
        \begin{subfigure}{0.16\textwidth}
        \includegraphics[width=\textwidth]{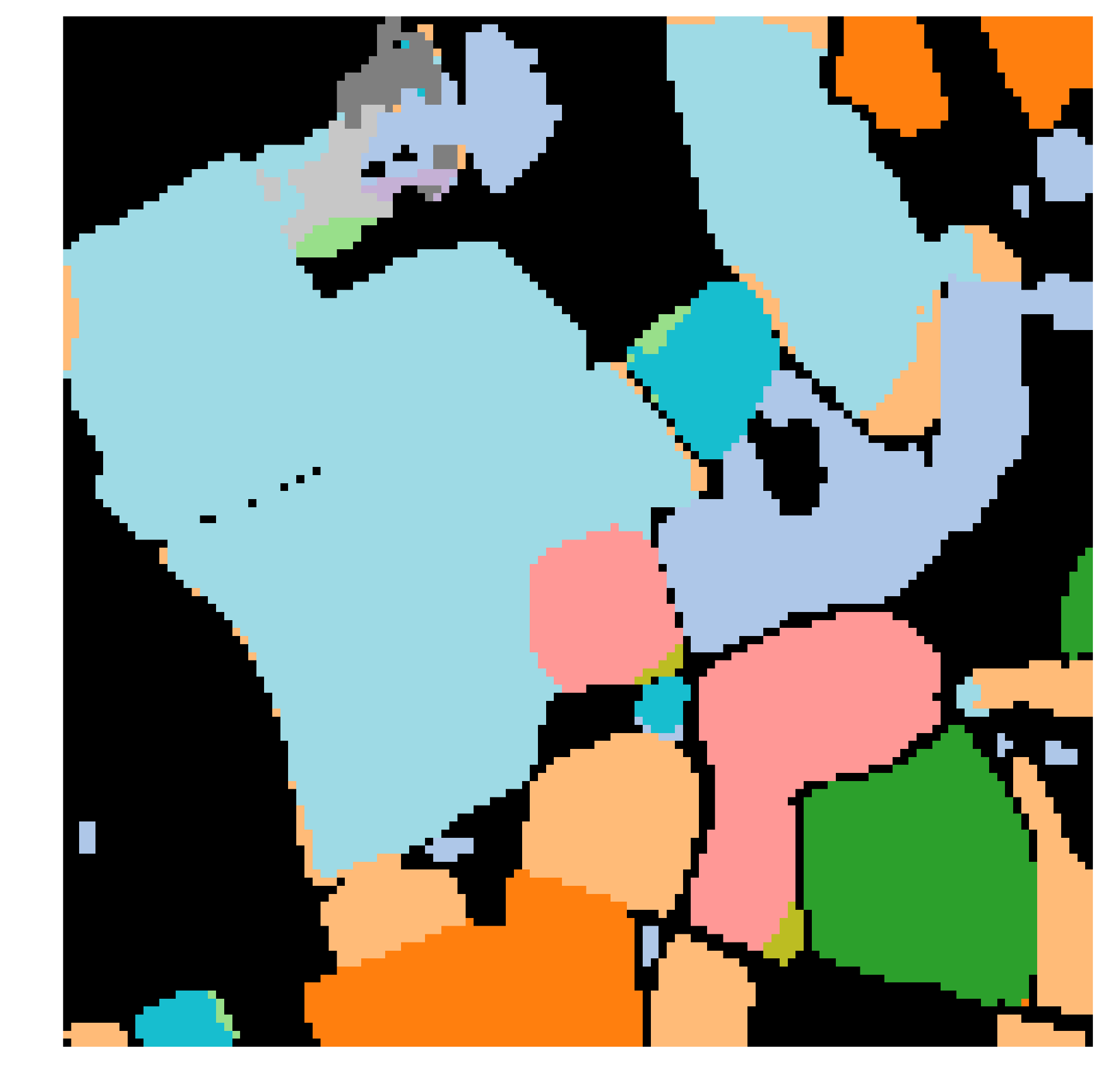}
        \caption{3D-Unet.}
        \label{fig:sem:3d}
        \end{subfigure}
    \hfill
        \begin{subfigure}{0.16\textwidth}
        \includegraphics[width=\textwidth]{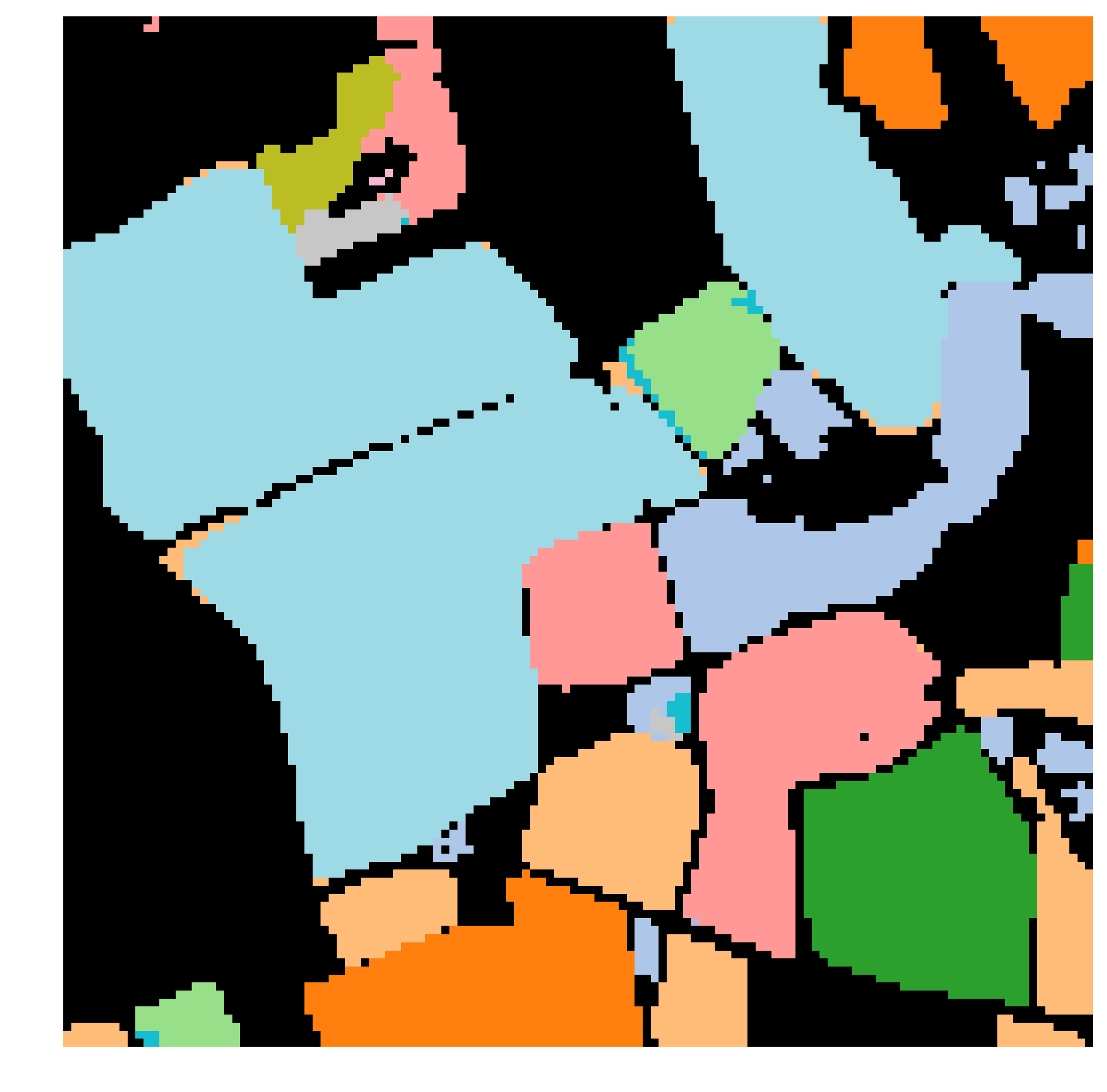}
        \caption{U-BiConvLSTM.}
        \label{fig:sem:ubc}

        \end{subfigure}
\hfill
        \begin{subfigure}{0.16\textwidth}
        \includegraphics[width=\textwidth, trim=0cm 4cm 0cm 4cm, clip]{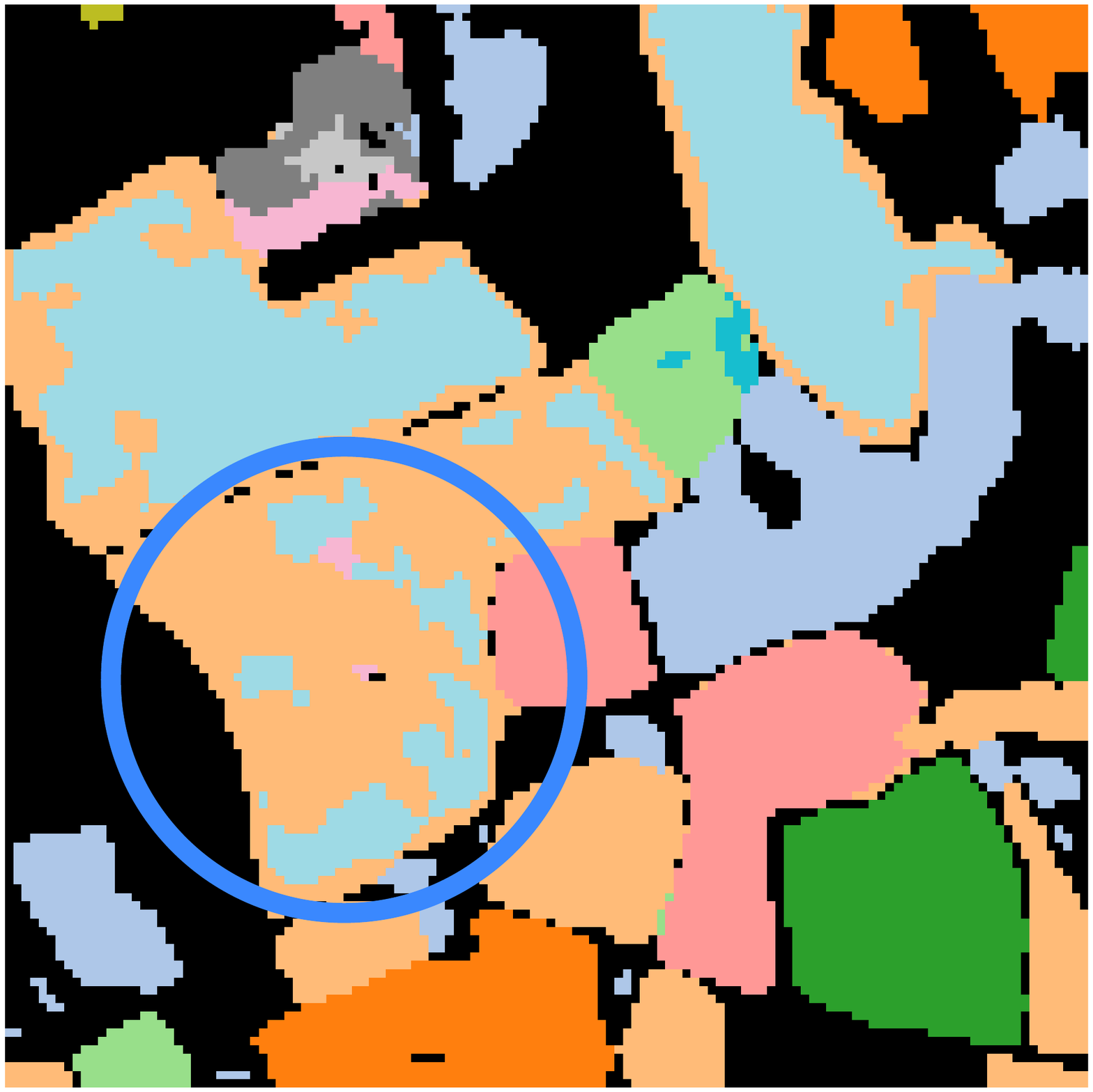}
        \caption{ConvGRU.}
        \label{fig:sem:cgru}
        \end{subfigure}

\caption{\textbf{Qualitative Semantic Segmentation Results.}
We represent a single image from the sequence using the RGB channels (\subref{fig:sem:rgb}), and whose ground truth parcel's limit and crop type are known (\subref{fig:sem:gt}). We then represent the pixelwise prediction from our approach (\subref{fig:sem:ours}), and for three other competing algorithms (\subref{fig:sem:3d}-\subref{fig:sem:cgru}). 
The different predictions shown on this figure illustrate the importance of the resolution at which temporal encoding is performed. ConvGRU applies a recurrent-convolutional network at the highest resolution, which results in predictions with high spatial variability. As a consequence, the prediction over large parcels are inconsistent (blue circles \protect\tikz \protect\node[circle, thick, draw = blue, fill = none, scale = 0.7] {};). Conversely, U-BiConvLSTM applies temporal encoding to feature maps with a larger receptive field, resulting in more spatially consistent predictions. Yet, this architecture often fails to retrieve small or thin parcels. In contrast, our U-TAE produces spatially consistent predictions on large parcels, while being able to retrieve such small parcels (green circles \protect\tikz \protect\node[circle, thick, draw = green!90!black, fill = none, scale = 0.7] {};). 3D-Unet also uses temporal encoding at different resolution levels, yet fails to recover these small parcels.
}
\label{fig:sem}
\end{figure*}

\begin{figure*}[ht!]
\begin{subfigure}{.24\textwidth}
\includegraphics[width=\textwidth]{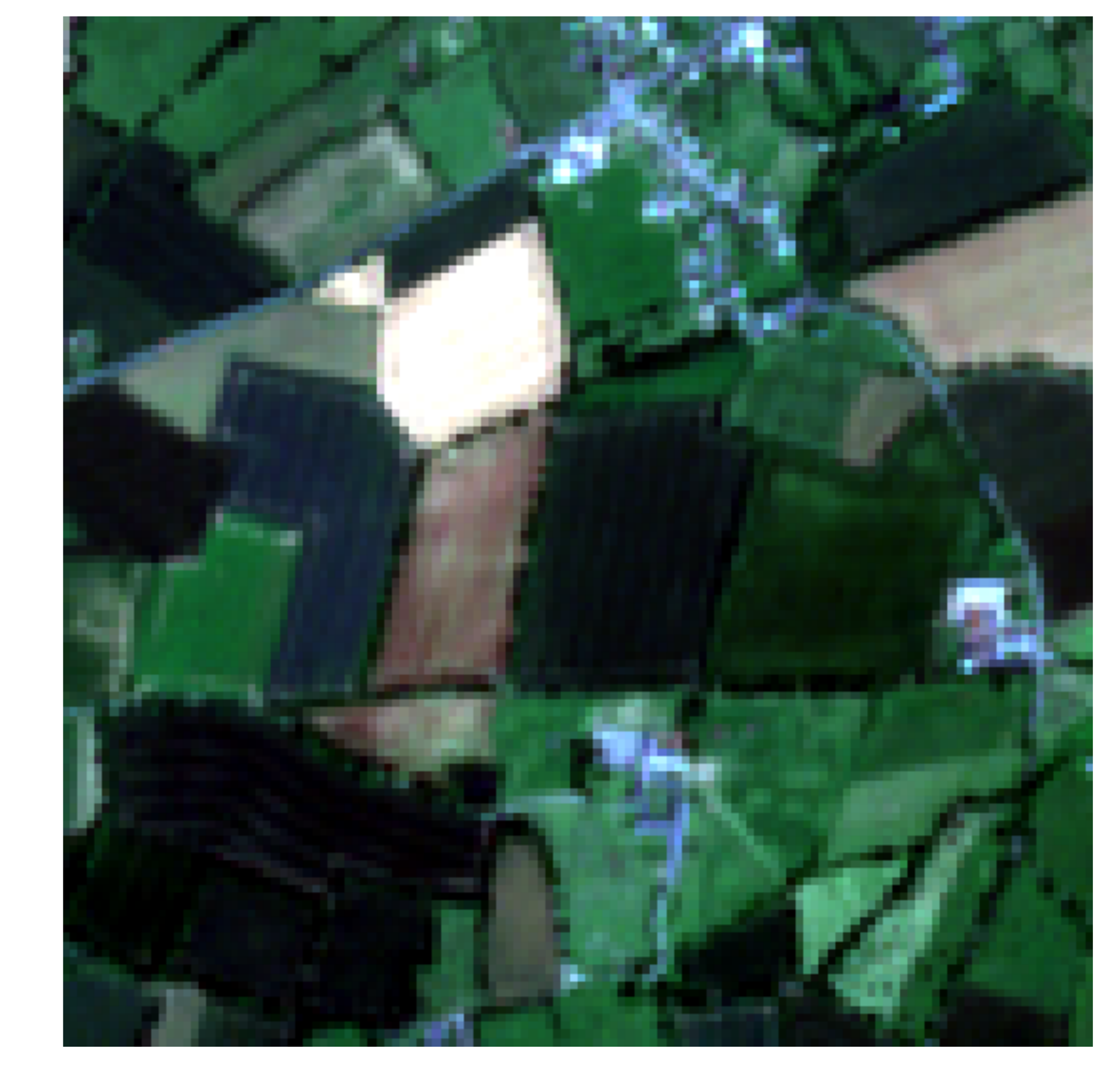}
\caption{Single observation.}
\label{fig:mono:rgb}
\end{subfigure}
\hfill
\begin{subfigure}{.24\textwidth}
\includegraphics[width=\textwidth]{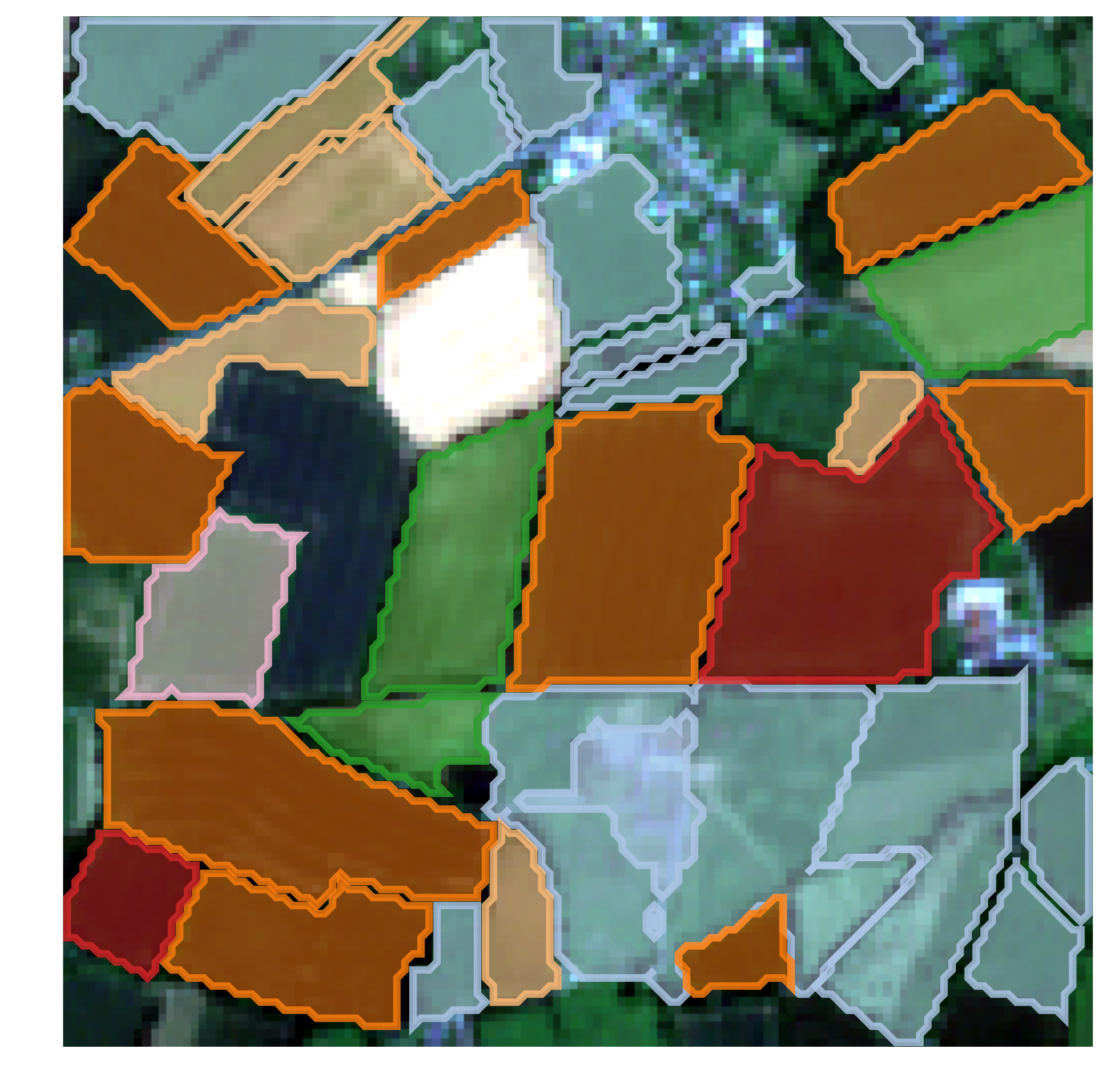}
\caption{Panoptic annotation.}
\label{fig:mono:gt}

\end{subfigure}
\hfill
\begin{subfigure}{.24\textwidth}
\includegraphics[width=\textwidth]{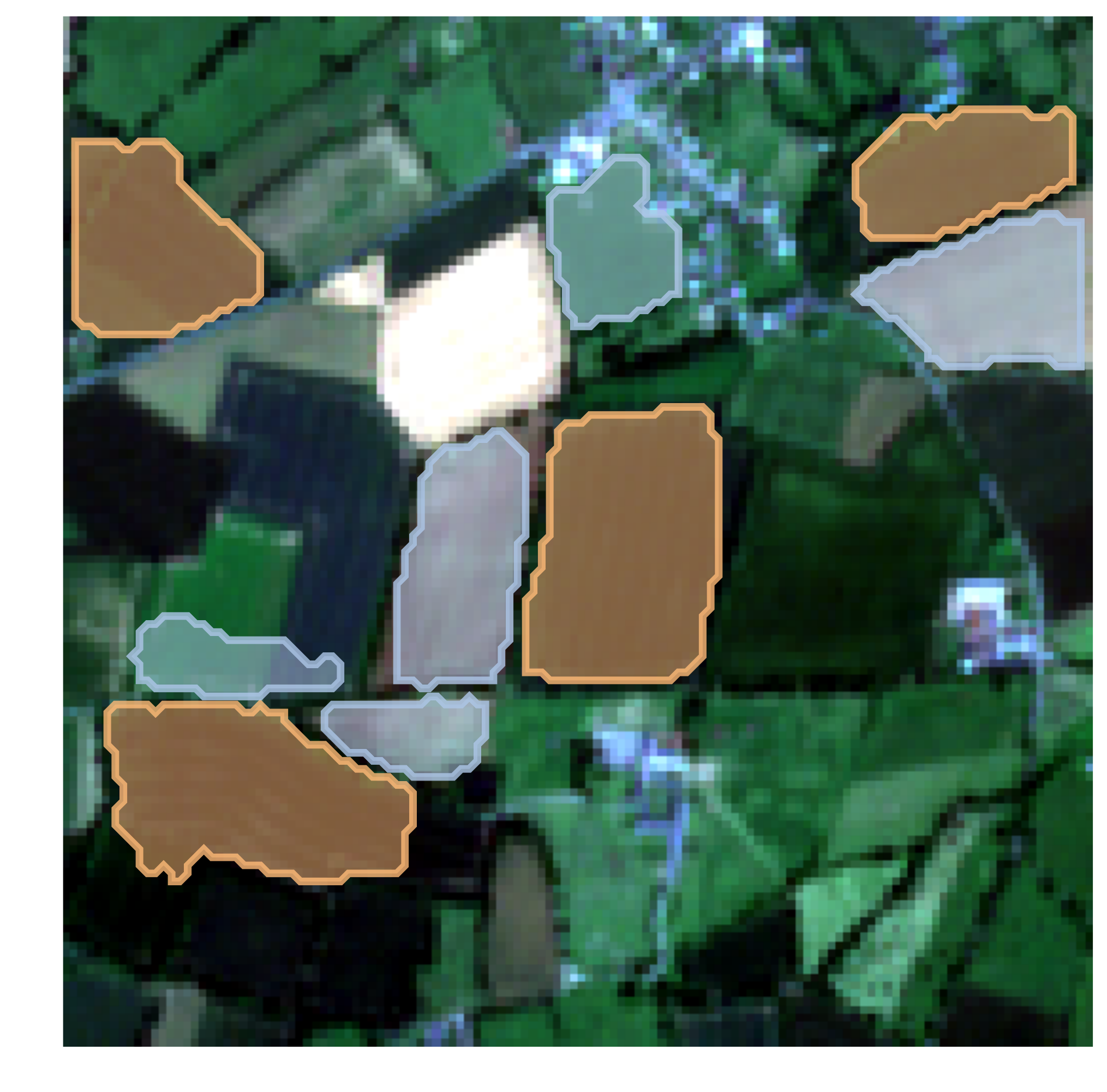}
\caption{Mono-temporal prediction.}
\label{fig:mono:pred}
\end{subfigure}
\hfill
\begin{subfigure}{.24\textwidth}

\begin{tikzpicture}
 \node[anchor=south west,inner sep=0] (image) at (0,0) {\includegraphics[width=\textwidth]{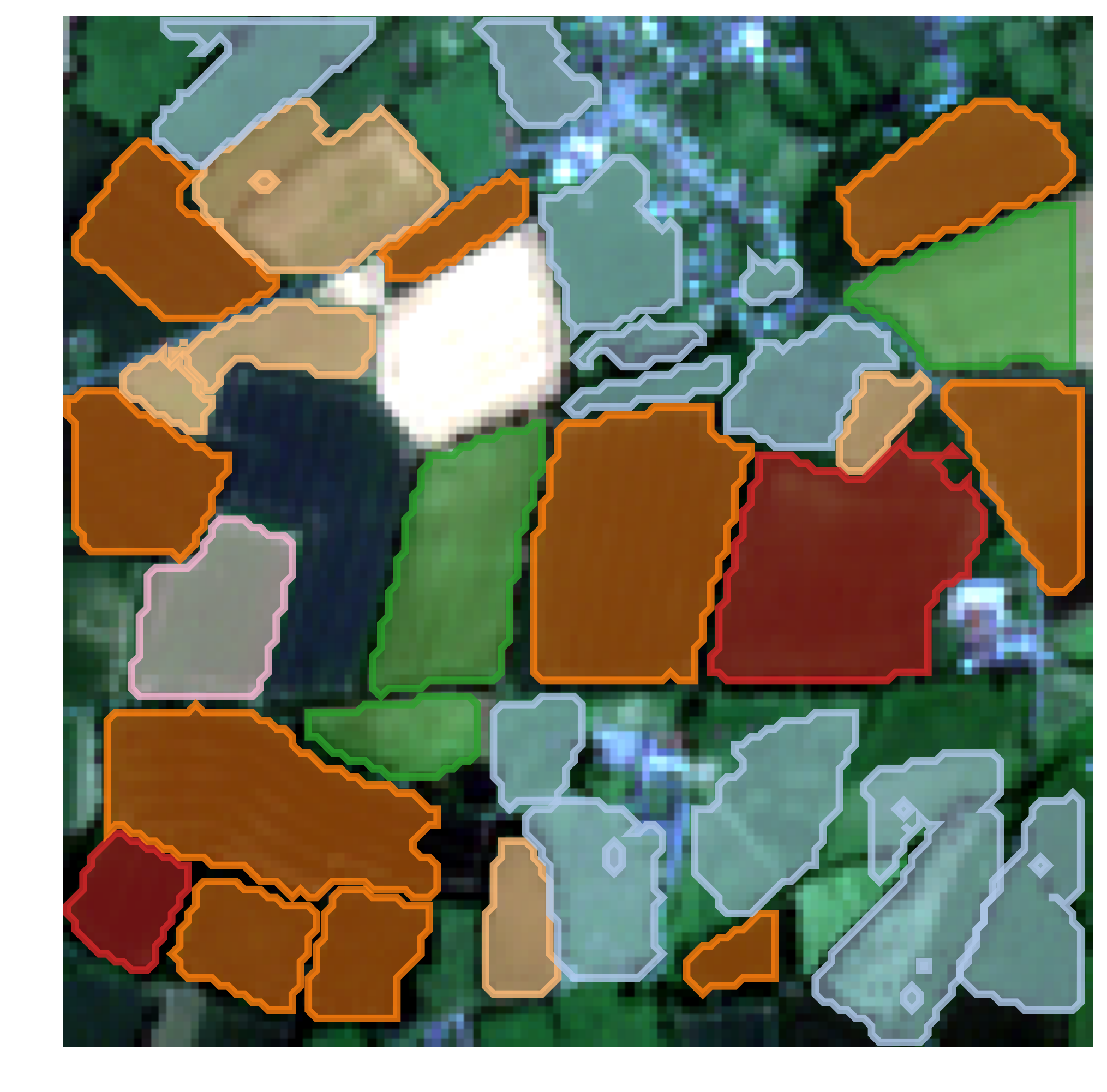}};
\begin{scope}[x={(image.south east)},y={(image.north west)}]
\draw[cyan ,ultra thick] (0.15,0.75) circle (0.1);
\end{scope}
\end{tikzpicture}

\caption{Multi-temporal prediction.}
\label{fig:mono:multi}
\end{subfigure}
\caption{\textbf{Mono-temporal Panoptic Segmentation.} 
We train our mono-temporal model on a single image (\subref{fig:mono:rgb}), with panoptic annotation (\subref{fig:mono:gt}).
We then compare the results of the mono-temporal model in (\subref{fig:mono:pred}) with the results our full model when performing inference on the full length sequence (\subref{fig:mono:multi}) from which the single patch (\subref{fig:mono:rgb}) is drawn.
First, we observe that many parcels are not detected by the mono-temporal model, indicating an overall low predicted {quality}.
Second, we can see that most detected parcels are misclassified by the mono-temporal model. This is in accordance with the low semantic segmentation score of the mono-temporal model: crop types are hard to distinguish from a single observation.
Last, adjacent parcels with no clear borders are predicted as a single parcel, when the multi-temporal model is able to differentiate between the two parcels (cyan circle \protect\tikz \protect\node[circle, thick, draw = cyan, fill = none, scale = 0.7] {};). 
This illustrates how using SITS instead of single images can help resolve ambiguous parcels delineation.
%We represent the image from the sequence used to train the mono-temporal model (\subref{fig:mono:rgb}), and the panoptic target (\subref{fig:mono:gt}). We show on (\subref{fig:mono:pred}) the predictions made by the mono-temporal model and on (\subref{fig:mono:multi}) the prediction made by the multi-temporal model on the same patch. First, we observe that many parcels are not detected by the mono-temporal model, indicating a low confidence in the predictions. Second, we can see that many parcels, when detected, are not attributed the correct semantic label. Last, neighboring parcels with no clear contrast are hardly distinguished by the mono-temporal model while successfully separated by the multi-temporal model (green circle), making a case for leveraging the temporal dimension of SITS.
}
\label{fig:mono}
\end{figure*}

}{}

\end{document}